\documentclass[10pt,journal,compsoc]{IEEEtran}
\usepackage{amsmath}
\usepackage{color}
\usepackage{amssymb}
\usepackage{algorithm}
\usepackage{algorithmic}
\usepackage[pagebackref=true,colorlinks,bookmarks=false]{hyperref}
\usepackage{array,multirow,graphicx}
\usepackage[table, dvipsname]{xcolor}
\usepackage{epstopdf}

\ifCLASSOPTIONcompsoc
  \usepackage[nocompress]{cite}
\else
  \usepackage{cite}
\fi
\ifCLASSINFOpdf
\else
\fi
\ifCLASSOPTIONcompsoc
 \usepackage[caption=false,font=footnotesize,labelfont=sf,textfont=sf]{subfig}
\else
 \usepackage[caption=false,font=footnotesize]{subfig}
\fi
\begin{document}
\title{A Survey on Learning to Hash}

\author{Jingdong~Wang,
Ting Zhang,
Jingkuan Song,
Nicu Sebe,
and Heng Tao Shen\IEEEcompsocitemizethanks{\IEEEcompsocthanksitem
J. Wang is with Microsoft Research,
Beijing, P.R. China. \protect\\
E-mail: jingdw@microsoft.com
\IEEEcompsocthanksitem
T. Zhang is with
University of Science and Technology of China.\protect\\
Email: zting@mail.ustc.edu.cn
\IEEEcompsocthanksitem
J. Song and N. Sebe is with Department of Information Engineering and Computer Science, University of Trento, Italy.\protect\\
Email: \{jingkuan.song, niculae.sebe\}@unitn.it
\IEEEcompsocthanksitem
H.T. Shen is with School of Computer Science and Engineering,
University of Electronic Science and Technology of China.\protect\\
Email: shenhengtao@hotmail.com \protect\\
Corresponding authors: Heng Tao Shen.
}
}

\markboth{IEEE TRANSACTIONS ON PATTERN ANALYSIS AND MACHINE INTELLIGENCE,~Vol.~13, No.~9, April~2017}%
{Shell \MakeLowercase{\textit{et al.}}: A Survey on Learning to Hash}

%\IEEEpubid{\makebox[\columnwidth]{\hfill 0000--0000/00/\$00.00~\copyright~2014 IEEE}%


\IEEEtitleabstractindextext{%
\begin{abstract}
Nearest neighbor search
is a problem
of finding the data points from the database
such that the distances from them to the query point
are the smallest.
Learning to hash is one of the major solutions to this problem
and has been widely studied recently.
In this paper, we present a comprehensive survey
of the learning to hash algorithms,
categorize them
according to the manners
of preserving the similarities into:
pairwise similarity preserving,
multiwise similarity preserving,
implicit similarity preserving,
as well as quantization,
and discuss their relations.
We separate quantization from pairwise similarity preserving
as the objective function is very different
though quantization, as we show,
can be derived
from preserving the pairwise similarities.
In addition,
we present the evaluation protocols,
and the general performance analysis,
and point out that
the quantization algorithms
perform superiorly
in terms of search accuracy,
search time cost,
and space cost.
Finally, we introduce a few emerging topics.

\end{abstract}

}

\maketitle

\IEEEdisplaynontitleabstractindextext

% \fi
\IEEEpeerreviewmaketitle

\IEEEraisesectionheading{\section{Introduction}\label{sec:introduction}}
\IEEEPARstart{T}{he}
problem of nearest neighbor search,
also known as
similarity search,
proximity search,
or close item search,
is aimed at finding an item,
called nearest neighbor,
which is the nearest
to a query item
under a certain distance measure
from a search (reference) database.
The cost of finding the exact nearest neighbor
is prohibitively high
in the case that the reference database
is very large
or that computing the distance
between the query item and the database item
is costly.
The alternative approach,
approximate nearest neighbor search,
is more efficient
and is shown to
be enough and useful
for many practical problems,
thus attracting an enormous number of research efforts.

Hashing, a widely-studied solution to the approximate nearest neighbor search,
aims to
transform a data item
to a low-dimensional representation,
or equivalently a short code consisting of a sequence of bits,
called hash code.
There are two main categories of hashing algorithms:
locality sensitive hashing~\cite{IndykM98, Charikar02} and learning to hash.
Locality sensitive hashing (LSH) is data-independent.
Following the pioneering works~\cite{IndykM98, Charikar02},
there are a lot of efforts,
such as proposing random hash functions
satisfying the locality sensitivity property
for various distance measures~\cite{Broder97, BroderGMZ97, DatarIIM04, Charikar02, DasguptaKS11, MotwaniNP07, ODonnellWZ11},
proving better search efficiency and accuracy~\cite{Panigrahy06, DasguptaKS11},
developing better search schemes~\cite{LvJWCL07, GanFFN12, GanFFN12},
providing a similarity estimator
with smaller variance~\cite{LiCH06},~\cite{JiLYZT12},~\cite{LiOZ12},~\cite{JiLYTZ13},
smaller storage~\cite{LiK10b},~\cite{LiKG10a},
or faster computation of hash functions \cite{LiHC06},~\cite{LiOZ12},~\cite{JiLYTZ13},~\cite{ShrivastavaL14}.
LSH has been adopted in many applications,
e.g., fast object detection~\cite{DeanRSSVY13},
image matching~\cite{MoonNPLSHP16, ChumM10}
The detailed review on LSH can be found in~\cite{WangSSJ14}.

Learning to hash, the interest of this survey,
is a data-dependent hashing approach
which aims to learn hash functions
from a specific dataset
so that the nearest neighbor search result
in the hash coding space
is as close as possible to the search result
in the original space,
and the search cost as well as the space cost are also small.
The development of learning to hash
has been inspired by the connection between the Hamming distance
and the distance provided from the original space, e.g., the cosine distance
shown in SimHash~\cite{Charikar02}.
Since the two early algorithms,
semantic hashing~\cite{SalakhutdinovH07, SalakhutdinovH09} and spectral hashing~\cite{WeissTF08}
that learns projection vectors instead of the random projections as done in~\cite{Charikar02},
learning to hash has been attracting a large amount of research interest
in computer vision and machine learning
and has been applied to a wide-range of applications
such as large scale object retrieval~\cite{JegouDSP10},
image classification~\cite{SanchezP11} and detection~\cite{VedaldiZ12},
and so on.

The main methodology of learning to hash
is similarity preserving,
i.e.,
minimizing the gap between the similarities
computed/given in the original space
and the similarities in the hash coding space
in various forms.
The similarity in the original space might
be from the semantic (class) information,
or from the distance (e.g., Euclidean distance)
computed in the original space,
which is of broad interest and widely studied in real applications,
e.g., large scale image search and image classification.
Hence the later is the main focus in this paper.

This survey
categorizes the algorithms according to
the similarity preserving manner into:
pairwise similarity preserving,
multiwise similarity preserving,
implicit similarity preserving,
quantization
which we will show is also a form of pairwise similarity preserving,
as well as
an end-to-end hash learning strategy
learning the hash codes directly from the object, e.g., image,
under the deep learning framework
instead of first learning the representations
and then learning the hash codes from the representations.
In addition,
we discuss other problems
including evaluation datasets and evaluation schemes,
and so on.
Meanwhile,
we present the empirical observation
that
the quantization approach
outperforms other approaches
and give some analysis about this observation.

In comparison to other surveys on learning to hash~\cite{WangSSJ14,WangLKC16},
this survey focuses more on learning to hash,
discusses more on quantization-based solutions.
Our categorization methodology is helpful
for readers to understand connections and differences
between existing algorithms.
In particular, we point out
the empirical observation
that quantization is superior
in terms of search accuracy, search efficiency
and space cost

%
%The organization of the remaining part is given as the following.
%Section~\ref{sec:overview} introduces
%the exact and approximate nearest neighbor search problems,
%and the search algorithms with hashing.
%Section~\ref{sec:conceptOfLTH} provides the basic concepts
%in the learning-to-hashing approach
%and categorizes the existing algorithms
%from the perspective of loss function
%into four main classes:
%pairwise alignment,
%multiple-wise alignment,
%implicit alignment
%and quantization,
%which are discussed in
%Sections~\ref{sec:pairwise},~\ref{sec:multiplewise},~\ref{sec:implicit},
%and~\ref{sec:quantization},
%respectively.
%Section~\ref{sec:others} presents other works in learning to hash.
%Sections~\ref{sec:evaluationprotocols} and~\ref{sec:discussion}
%give some evaluation protocols
%and performance analysis.
%Finally,
%Sections~\ref{sec:trends} and~\ref{sec:con}
%point out the emerging research trends and
%conclude this survey, respectively.

\section{Background}
\label{sec:overview}
\subsection{Nearest Neighbor Search}
Exact nearest neighbor search is defined as
searching an item $\operatorname{NN}(\mathbf{q})$
(called nearest neighbor)
for a query item $\mathbf{q}$
from a set of $N$ items $\mathcal{X} = \{\mathbf{x}_1, \mathbf{x}_2, \cdots, \mathbf{x}_N\}$
so that
$\operatorname{NN}(\mathbf{q}) = \arg\min_{\mathbf{x} \in \mathcal{X}} \operatorname{dist}(\mathbf{q}, \mathbf{x})$,
where $\operatorname{dist}(\mathbf{q}, \mathbf{x})$ is a distance
computed between $\mathbf{q}$ and $\mathbf{x}$.
A straightforward generalization
is $K$-nearest neighbor search,
where $K$ nearest neighbors are needed to be found.% ($\operatorname{KNN}(\mathbf{q})$)

The distance
between a pair of items $\mathbf{x}$ and $\mathbf{q}$
depends on the specific nearest search problem.
A typical example is
that
the search (reference) database $\mathcal{X}$ lies in
a $d$-dimensional space $\mathbb{R}^d$
and the distance is introduced by an $\ell_s$ norm,
$\|\mathbf{x} - \mathbf{q}\|_s = (\sum_{i=1}^d |x_i - q_i|^s)^{1/s}$.
The search problem under the Euclidean distance, i.e., the $\ell_2$ norm,
is widely studied.
Other forms of the data item,
for example, the data item is formed by a set,
and other forms of distance measures,
such as $\ell_1$ distance,
cosine similarity and so on,
are also possible.

There exist efficient algorithms (e.g., $k$-d trees)
for exact nearest neighbor search
in low-dimensional cases.
In large scale high-dimensional cases,
it turns out that the problem becomes hard
and most algorithms even take higher computational cost
than the naive solution, i.e., the linear scan.
Therefore,
a lot of recent efforts moved
to searching approximate nearest neighbors:
error-constrained nearest (near) neighbor search,
and
time-constrained approximate nearest neighbor search~\cite{MujaL09, MujaL14}.
The error-constrained search includes
(randomized) $(1+\epsilon)$-approximate nearest neighbor search~\cite{IndykM98, Charikar02, AndoniI06},
(approximate) fixed-radius near neighbor ($R$-near neighbor) search~\cite{BentleySW77},
and so on.

Time-constrained approximate nearest neighbor search
limits the time spent during the search
and is studied mostly for real applications,
though it usually lacks an elegant theory behind.
The goal is to make the search as accurate as possible
by comparing the returned $K$ approximate nearest neighbors
and the $K$ exact nearest neighbors,
and to make the query cost as small as possible.
For example,
when comparing the learning to hash approaches
that use linear scan based on the Hamming distance for search,
it is typically assumed that the search time is the same for the same code length
by ignoring other small cost.
When comparing the indexing structure algorithms,
e.g., tree-based~\cite{MujaL09, WangWJLZZH13, MujaL14} or neighborhood graph-based~\cite{WangWZGLG13},
the time-constrained search is usually transformed
to another approximate way:
terminate the search after examining a fixed number of data points.

\iffalse
\begin{figure*}
\centering
\subfloat[(a)][]{\includegraphics[width=.45\linewidth, clip]{figs/multiTableLookup}}~~~~
\subfloat[(b)][]{\includegraphics[width=.45\linewidth, clip]{figs/singleTableLookup}}\\
\subfloat[(c)][]{\includegraphics[width=.45\linewidth, clip]{figs/hashCodeRanking}}~~~~
\subfloat[(c)][]{\includegraphics[width=.45\linewidth, clip]{figs/nonExhaustiveSearch}}
\caption{Illustrating the search strategies.
(a) Multi table lookup:
the list
corresponding to the hash code of the query
in each table
is retrieved.
(b) Single table lookup:
the lists corresponding to and near to the hash code
of the query are retrieved.
(c) Hash code ranking:
compare the query with each reference item
in the coding space.
(d) Non-exhaustive search:
hash table lookup (or other inverted index struture) retrieves the candidates,
followed by hash code ranking over the candidates.
The hash codes are different in two stages.}
\label{fig:searchschemes}
\end{figure*}
\fi

\begin{figure*}
\centering
(a)~{\includegraphics[width=.4\linewidth, clip]{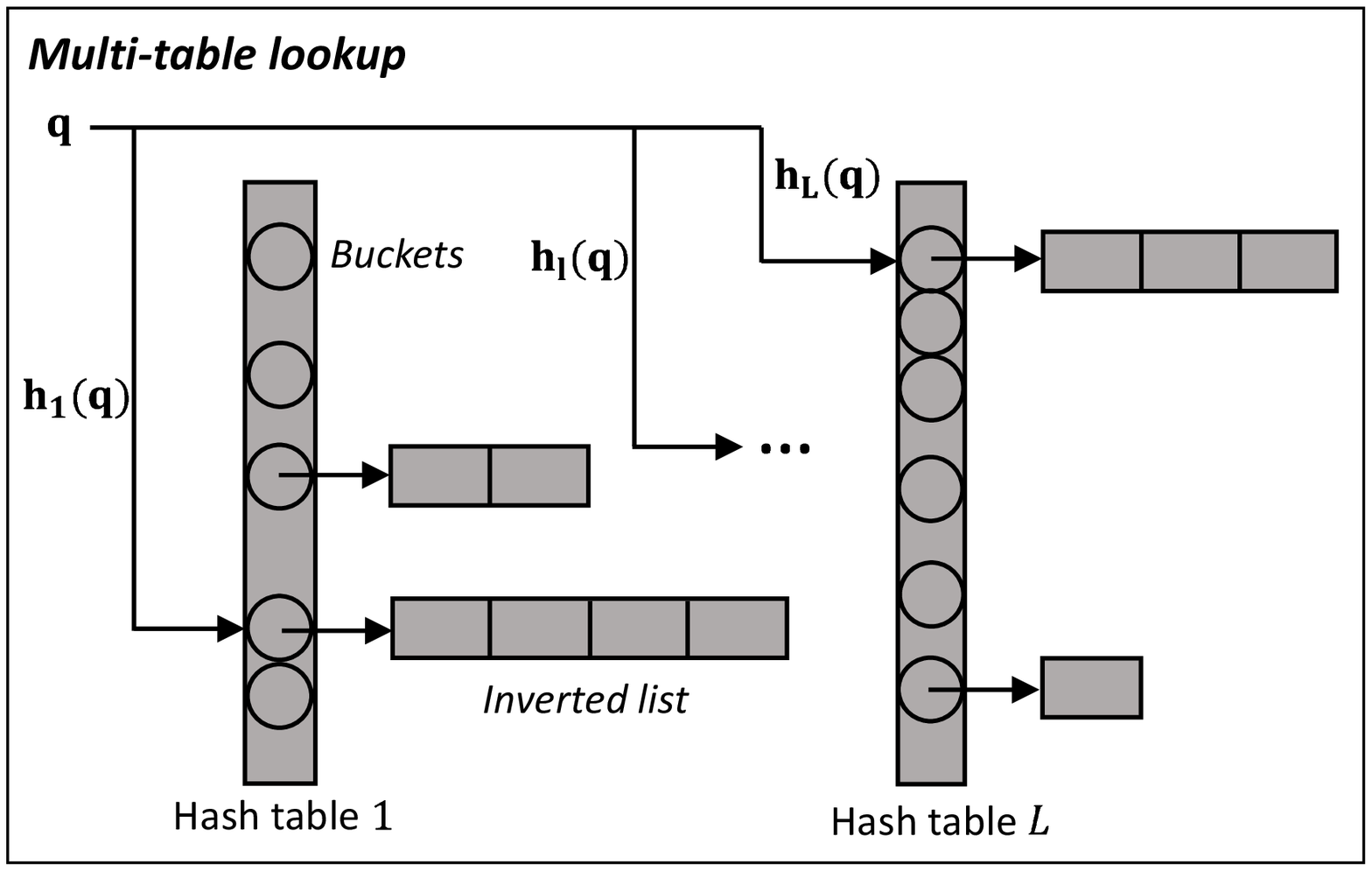}}~~~~~~
(b)~{\includegraphics[width=.4\linewidth, clip]{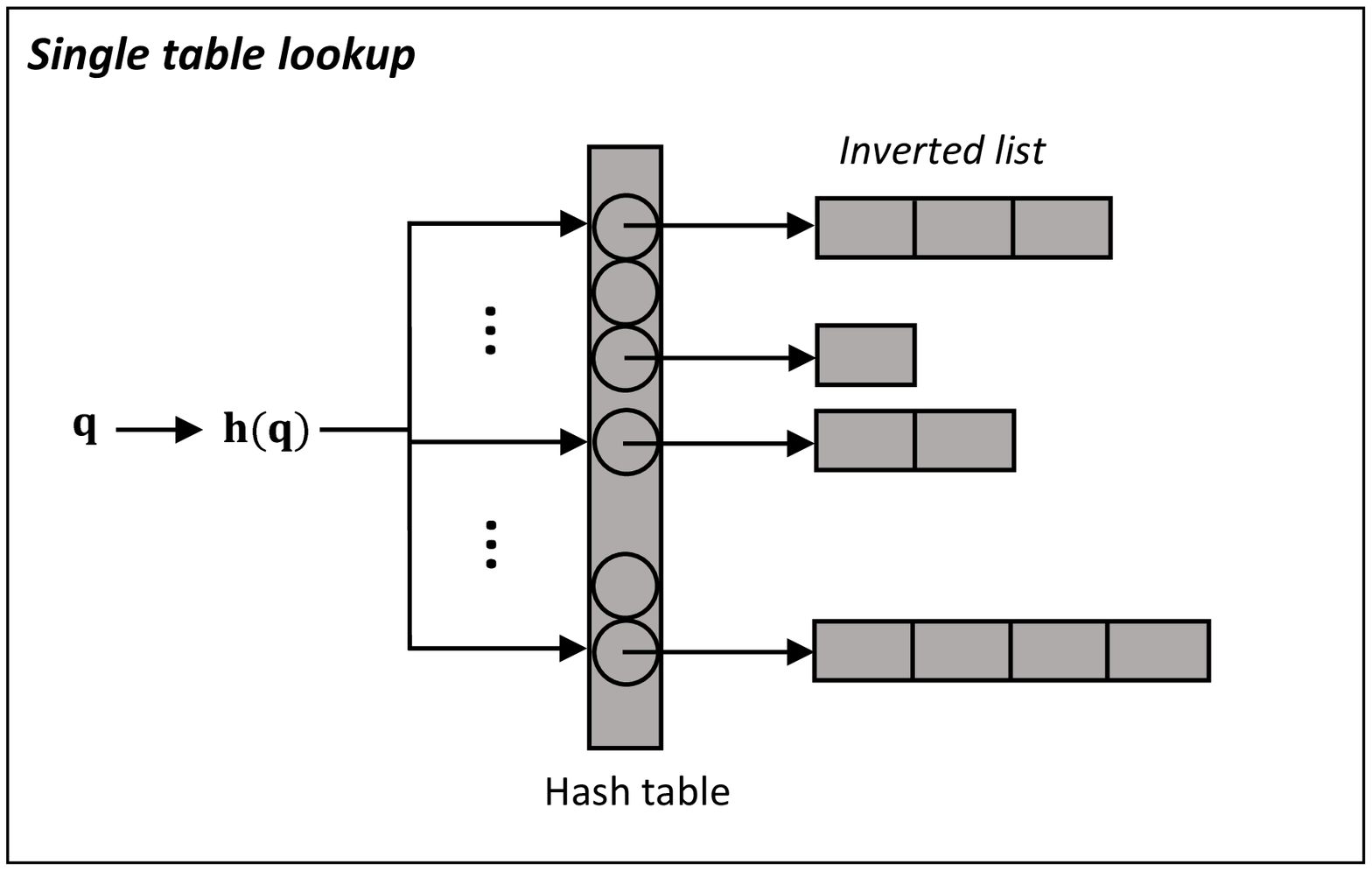}}\\
(c)~{\includegraphics[width=.4\linewidth, clip]{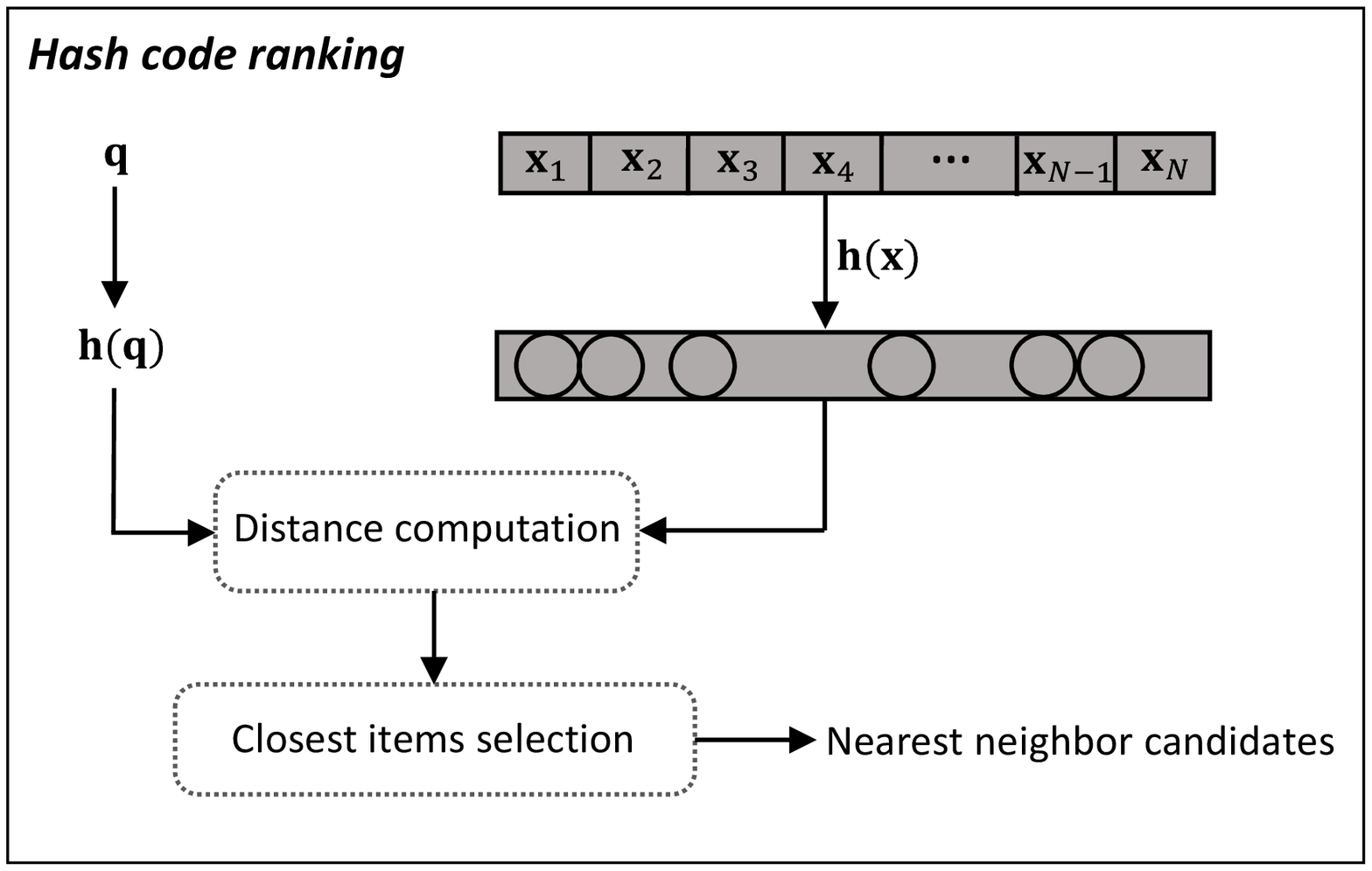}}~~~~~~
(d)~{\includegraphics[width=.4\linewidth, clip]{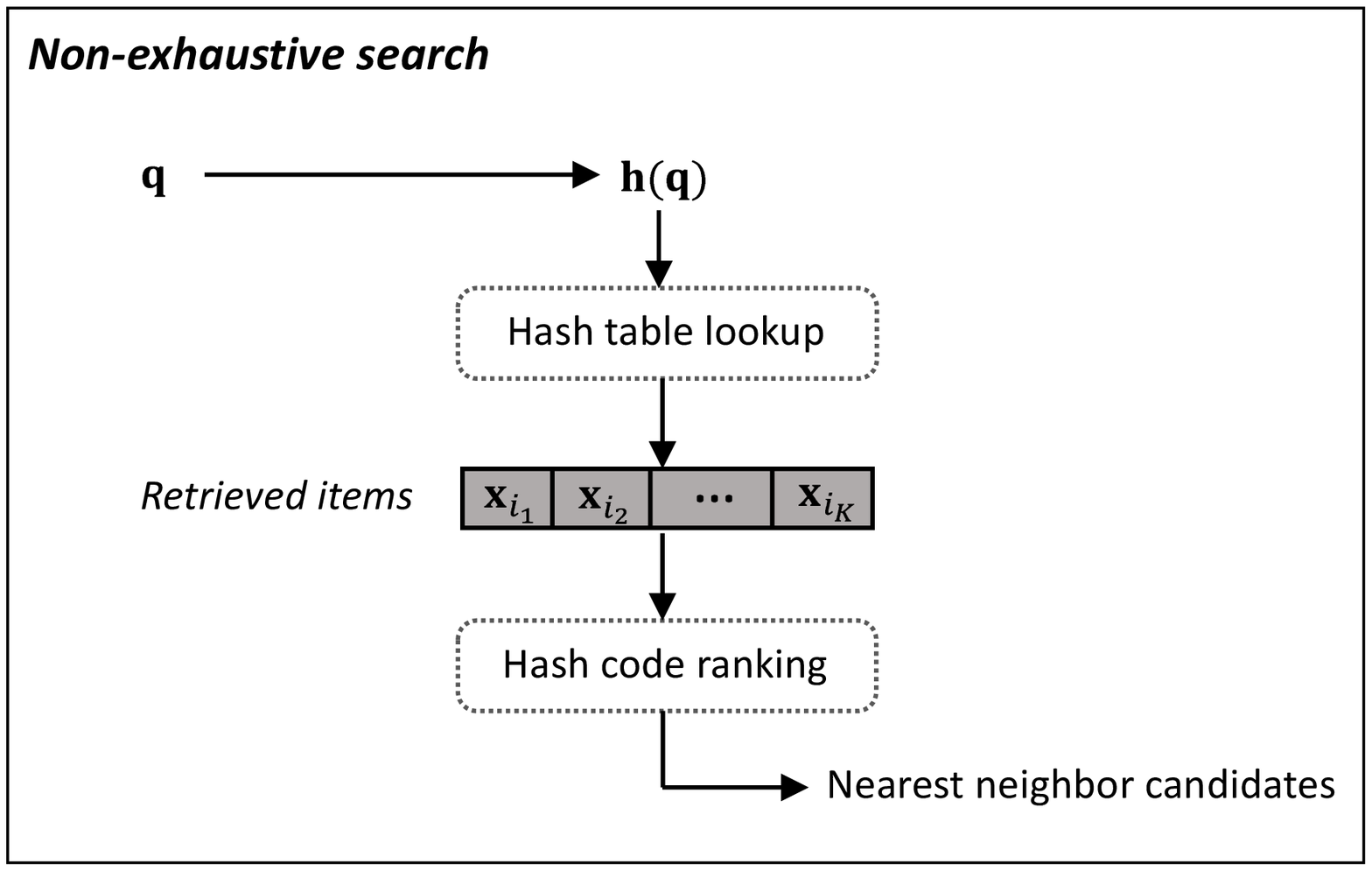}}
\caption{Illustrating the search strategies.
(a) Multi table lookup:
the list
corresponding to the hash code of the query
in each table
is retrieved.
(b) Single table lookup:
the lists corresponding to and near to the hash code
of the query are retrieved.
(c) Hash code ranking:
compare the query with each reference item
in the coding space.
(d) Non-exhaustive search:
hash table lookup (or other inverted index structure) retrieves the candidates,
followed by hash code ranking over the candidates.}
\label{fig:searchschemes}
\vspace{-.5cm}
\end{figure*}

\subsection{Search with Hashing}
The hashing approach aims to
map the reference (and query) items
to the target items
so that approximate nearest neighbor search
is efficiently and accurately performed
by resorting to the target items and possibly
a small subset of the raw reference items.
The target items are
called hash codes
(a.k.a., hash values, or
simply hashes).
In this paper, we may also call it
short/compact codes interchangeably.

The hash function is formally defined as:
$y = h(\mathbf{x})$,
where $y$ is the hash code,
may be an integer, or a binary value:
$1$ and $0$ (or $-1$),
and $h(\cdot)$ is the hash function.
In the application
to approximate nearest neighbor search,
usually several hash functions are used
together
to compute the compound hash code:
$\mathbf{y} = \mathbf{h}(\mathbf{x})$,
where $\mathbf{y} = [y_1~y_2~\cdots~y_M]^\top$
and $\mathbf{h}(\mathbf{x})= [h_1(\mathbf{x})~h_2(\mathbf{x})~\cdots~h_M(\mathbf{x})]^\top$.
Here we use a vector $\mathbf{y}$
to represent the compound hash code
for convenience.

There are two basic strategies for
using hash codes to perform
nearest (near) neighbor search:
\emph{hash table lookup}
and \emph{hash code ranking}.
The search strategies are illustrated
in Figure~\ref{fig:searchschemes}.

The main idea of \emph{hash table lookup} for accelerating the search
is reducing the number of the distance computations.
The data structure,
called hash table
(a form of inverted index),
is composed of buckets
with each bucket indexed by a hash code.
Each reference item $\mathbf{x}$ is placed
into a bucket $\mathbf{h}(\mathbf{x})$.
Different from the conventional hashing algorithm
in computer science
that avoids collisions
(i.e., avoids mapping two items
into some same bucket),
the hashing approach using a hash table
essentially aims to maximize the probability
of collision of near items
and at the same time minimize
the probability of collision
of the items that are far away.
Given the query $\mathbf{q}$,
the items lying in the bucket $\mathbf{h}(\mathbf{q})$
are retrieved as the candidates of the nearest items of $\mathbf{q}$.
Usually this is followed by
a reranking step:
rerank the retrieved nearest neighbor candidates
according to the true distances
computed using the original features
and attain the nearest neighbors.

To improve the recall,
two ways are often adopted.
The first way is
to visit a few more buckets (but with a single hash table),
whose corresponding hash codes
are the nearest to (the hash code $\mathbf{h}(\mathbf{q})$ of) the query
according to the distances in the coding space.
The second way is to construct several (e.g., $L$) hash tables.
The items lying in the $L$ hash buckets
$\mathbf{h}_1(\mathbf{q}), \cdots, \mathbf{h}_L(\mathbf{q})$
are retrieved
as the candidates of near items of $\mathbf{q}$
which are possibly ordered according to the number of hits
of each item in the $L$ buckets.
To guarantee the high precision,
each of the $L$ hash codes, $\mathbf{y}_l$,
needs to be a long code.
This means that
the total number of the buckets is too large
to index directly,
and thus
the buckets that are non-empty are retained
by using the conventional hashing over the hash codes
$\mathbf{h}_l(\mathbf{x})$.

The second way essentially stores multiple copies
of the id for each reference item.
Consequently, the space cost is larger.
In contrast,
the space cost for the first way
is smaller
as it only uses a single table
and stores one copy of the id for each reference item,
but it needs to access more buckets to guarantee
the same recall
with the second way.
The multiple assignment scheme is also studied:
construct a single table,
but assign a reference item to multiple hash buckets.
In essence,
it is shown that
the second way, multiple hash tables,
can be regarded as
a form of multiple assignment.

\emph{Hash code ranking}
performs an exhaustive search:
compare the query with each reference item
by fast evaluating their distance
(e.g., using distance table lookup or using the CPU instruction $\operatorname{\_\_popcnt}$ for Hamming distance)
according to (the hash code of) the query and the hash code of the reference item,
and retrieve the reference items with the smallest distances
as the candidates of nearest neighbors.
Usually this is followed by
a reranking step:
rerank the retrieved nearest neighbor candidates
according to the true distances
computed using the original features
and attain the nearest neighbors.

This strategy exploits one main advantage
of hash codes: the distance using hash codes
is efficiently computed
and the cost is much smaller
than that of the distance computation
in the original input space.

\textbf{Comments:}
Hash table lookup is mainly used in locality sensitive hashing,
and has been used for evaluating learning to hash in a few publications.
It has been pointed in~\cite{WeissTF08}
and also observed from empirical results
that LSH-based hash table lookup,
except min-hash,
is rarely adopted in reality,
while hash table lookup with quantization-based hash codes
is widely used in the non-exhaustive strategy
to retrieve coarse candidates~\cite{JegouDS11}.
Hash code ranking goes through all the candidates
and thus is inferior in search efficiency
compared with hash table lookup
which only checks a small subset of candidates,
which are determined by a lookup radius.

A practical way
is to do a non-exhaustive search
which is suggested in~\cite{JegouDS11, BabenkoL12}:
first retrieve a small set of candidates
using the inverted index that can be viewed as a hash table,
and then compute the distances of the query
to the candidates using the hash codes which are longer,
providing the top candidates subsequently reranked using the original features.
Other research efforts include organizing the hash codes to avoid the exhaustive search
with a data structure,
such as a tree or a graph structure~\cite{MujaL12}.

\section{Learning to Hash}
\label{sec:conceptOfLTH}
Learning to hash
is the task of
learning a (compound) hash function,
$\mathbf{y} = \mathbf{h}(\mathbf{x})$,
mapping an input item $\mathbf{x}$
to a compact code $\mathbf{y}$,
aiming that
the nearest neighbor search result for a query $\mathbf{q}$
is as close as possible to the true nearest search result
and the search in the coding space is also efficient.
A learning-to-hash approach
needs to consider five problems:
what hash function $\mathbf{h}(\mathbf{x})$ is adopted,
what similarity in the coding space is used,
what similarity is provided in the input space,
what loss function is chosen for the optimization objective,
and what optimization technique is adopted.

\subsection{Hash Function}
The hash function can be based on linear projection,
kernels,
spherical function,
(deep) neural networks,
a non-parametric function,
and so on.
One popular hash function is the linear hash function,
e.g.,~\cite{WangKC10a, StrechaBBF12}:
\begin{align}
y = h(\mathbf{x}) = \operatorname{sgn}(\mathbf{w}^\top\mathbf{x} + b),
\end{align}
where $\operatorname{sgn}(z) = 1$
if $z \geqslant 0$
and $\operatorname{sgn}(z) = 0$
(or equivalently $-1$)
otherwise,
$\mathbf{w}$ is the projection vector,
and $b$ is the bias variable.
The kernel function,
\begin{align}
y = h(\mathbf{x}) = \operatorname{sgn}\left(\sum\nolimits_{t=1}^Tw_t K(\mathbf{s}_t,\mathbf{x}) + b\right),
\end{align}
is also adopted in some approaches,
e.g.,~\cite{KulisD09,HeLC10},
where $\{\mathbf{s}_t\}$ is a set of representative samples
that are randomly drawn from the dataset
or cluster centers of the dataset
and $\{w_t\}$ are the weights.
The non-parametric function based on nearest vector assignment is widely
used for quantization-based solutions:
\begin{align}
y = \arg\min\nolimits_{k \in \{1, \cdots, K\}} \|\mathbf{x} - \mathbf{c}_k\|_2,
\end{align}
where $\{\mathbf{c}_1, \cdots, \mathbf{c}_K\}$ is a set of centers
computed by some algorithms, e.g., $K$-means,
and $y \in \mathbb{Z}^{+}$ is an integer.
In contrast to other hashing algorithms
in which the distance, e.g., Hamming distance,
is often directly computed from hash codes,
the hash codes generated
from the nearest vector assignment-based hash function
are the indices of the nearest vectors,
and the distance is computed using the centers
corresponding to the hash codes.

The form of hash function is an important factor
influencing the search accuracy using the hash codes,
as well as the time cost of computing hash codes.
A linear function is efficiently evaluated,
while the kernel function and the nearest vector assignment based function
lead to better search accuracy
as they are more flexible.
Almost all the methods using a linear hash function
can be extended to nonlinear hash functions,
such as kernelized hash functions, or
neural networks.
Thus we do not use the hash function to categorize the hash algorithms.

\subsection{Similarity}
In the input space
the distance $d^o_{ij}$
between any pair of items $(\mathbf{x}_i, \mathbf{x}_j)$
could be the Euclidean distance, $\| \mathbf{x}_i - \mathbf{x}_j \|_2$
or others.
The similarity $s^o_{ij}$ is often
defined as a function about the distance $d^o_{ij}$,
and a typical function is the Gaussian function:
$s^o_{ij} = g(d^o_{ij}) = \exp{(-\frac{(d^o_{ij})^2}{2\sigma^2})}$.
There exist other similarity forms,
such as
cosine similarity $\frac{\mathbf{x}_i^\top\mathbf{x}_j}{\|\mathbf{x}_i\|_2 \|\mathbf{x}_j\|_2}$
and so on.
Besides,
the semantic similarity is often used
for semantic similarity search.
In this case, the similarity $s^o_{ij}$
is usually binary,
valued $1$ if the two items $\mathbf{x}_i$ and $\mathbf{x}_j$
belong to the same semantic class,
$0$ (or $-1$) otherwise.
The hashing algorithms for semantic similarity
usually can be applied to other distances, such as Euclidean distance,
by defining a pseudo-semantic similarity:
$s^o_{ij} = 1$ for nearby points $(i,j)$
and $s^o_{ij} = 0$ (or $-1$)
for farther points $(i,j)$.

In the hash coding space,
the typical distance $d^h_{ij}$
between $\mathbf{y}_i$ and $\mathbf{y}_j$
is the Hamming distance.
It is
defined as
the number of bits
where the values are different
and is mathematically formulated
as
$$d^h_{ij} = \sum\nolimits_{m=1}^M \delta[y_{im} \neq y_{jm}],$$
which is equivalent to $d^h_{ij} = \|\mathbf{y}_i - \mathbf{y}_j\|_1$
if the code is valued by $1$ and $0$.
The distance for the codes valued by $1$ and $-1$
is similarly defined.
The similarity based on the Hamming distance
is defined as
$s^h_{ij} = M - d^h_{ij}$
for the codes valued by $1$ and $0$,
computing the number of bits
where the values are the same.
The inner product $s^h_{ij} = \mathbf{y}_i^\top \mathbf{y}_j$ is used as the similarity
for the codes valued by $1$ and $-1$.
These measures are also extended to the weighted case:
e.g.,
$d^h_{ij} = \sum_{m=1}^M \lambda_m \delta[y_{im} \neq y_{jm}]$
and $s^h_{ij} = \mathbf{y}_i^\top \boldsymbol{\Lambda} \mathbf{y}_j$,
where $\boldsymbol{\Lambda} = \operatorname{Diag}(\lambda_1, \lambda_2, \cdots, \lambda_M)$ is a diagonal matrix
and each diagonal entry is the weight
of the corresponding hash code.

Besides the Hamming distance/similarity and its variants,
the Euclidean distance is
typically used
in quantization approaches,
and is evaluated between the vectors
corresponding to the hash codes,
$d^h_{ij} = \|\mathbf{c}_{y_i} - \mathbf{c}_{y_j}\|_2$
(symmetric distance)
or between the query $\mathbf{q}$
and the center that is the approximation to $\mathbf{x}_j$,
$d^h_{qj} = \|\mathbf{q} - \mathbf{c}_{y_j}\|_2$
(asymmetric distance,
which is preferred
because the accuracy is higher
and the time cost is almost the same).
The distance is usually evaluated in the search stage efficiently
by using a distance lookup table.
There are also some works
learning/optimizing the distances between hash codes~\cite{GordoPGL14, WangSYYLW14}
after the hash codes are already computed.

\subsection{Loss Function}
The basic rule of designing the loss function
is to preserve the similarity order,
i.e.,
minimize the gap between
the approximate nearest neighbor search result
computed from the hash codes
and the true search result
obtained from the input space.

The widely-used solution
is pairwise similarity preserving,
making the distances or similarities
between a pair of items
from the input and coding spaces
as consistent as possible.
The multiwise similarity preserving solution,
making the order among multiple items
computed from the input and coding spaces
as consistent as possible,
is also studied.
One class of solutions,
e.g., spatial partitioning,
implicitly preserve the similarities.
The quantization-based solution
and other reconstruction-based solutions
aim to find the optimal approximation
of the item in terms of the reconstruction error
through a reconstruction function
(e.g., in the form of a lookup table in quantization
or an auto-encoder in~\cite{SalakhutdinovH07}).
Besides similarity preserving items,
some approaches introduce bucket balance or its approximate variants
as extra constraints,
which is also important for
obtaining better results
or avoiding trivial solutions.

\subsection{Optimization}
The challenges for optimizing the hash function parameters
lie in two main factors.
One is that the problem contains the $\operatorname{sgn}$ function,
which leads to a challenging mixed-binary-integer optimization problem.
The other is that
the time complexity is high when processing a large number of data points,
which is usually handled by sampling a subset of points
or a subset of constraints (or equivalent basic terms in the objective functions).
%and will be discussed in Section~\ref{sec:discussion:queryperformance:HCR}.

The ways
to handle
the $\operatorname{sgn}$ function
are summarized below.
The first way is the most widely-adopted continuous relaxation,
including sigmoid relaxation, tanh relaxation,
%$\operatorname{sgn}(z) \approx \phi_{\alpha}(z) = \frac{1}{1 + e^{-\alpha z}}$
and directly dropping the sign function
$\operatorname{sgn}(z) \approx z$.
The relaxed problem is then solved
using various standard optimization techniques.
The second one is a two-step scheme~\cite{LinSSH13, LinSSHS14}
with its extension to alternative optimization~\cite{GeH014}:
optimizing the binary codes without considering the hash function,
followed by estimating the function parameters
from the optimized hash codes.
The third one is discretization:
drop the sign function
$\operatorname{sgn}(z) \approx z$
and regard the hash code as an approximation
of the hash function,
which is formulated as a loss $(y-z)^2$.
There also exist other ways only adopted in a few algorithms,
e.g., transforming the problem into a latent structure-SVM formulation in~\cite{NorouziF11, NorouziFS12} ,
the coordinate-descent approach in~\cite{KulisD09}
(fixing all but one weight,
optimize the original objective with respect to a single weight in each iteration),
both of which do not conduct continuous relaxation.

\subsection{Categorization}
Our survey categorizes the existing algorithms
to various classes:
the pairwise similarity preserving class,
the multiwise similarity preserving class,
the implicit similarity preserving class,
as well as the quantization class,
according to what similarity preserving manner is adopted
to formulate the objective function.
We separate the quantization class
from the pairwise similarity preserving class
as they are very different in formulation
though
the quantization class can be explained
from the perspective of
pairwise similarity preserving.
In the following description,
we may call quantization as quantization-based hashing
and other algorithms in which a hash function
generates a binary value
as binary code hashing.
In addition,
we will also discuss other studies on learning to hash.
The summary of the representative algorithms
is given in Table~\ref{table:summary}.

The main reason we choose the similarity preserving manner
to do the categorization
is that similarity preservation is the essential goal
of hashing.
It should be noted that
as pointed in~\cite{WangLKC16, WangSSJ14},
other factors, such as the hash function,
or the optimization algorithm,
is also important
for the search performance.

\begin{table*}[t]
\scriptsize
%  \parbox[t]{2mm}{\multirow{5}{*}{\rotatebox[origin=c]{90}{pair-wise}}}
\caption{A summary of representative hashing algorithms
with respect to similarity preserving functions,
code balance,
hash function
similarity in the coding space,
and the manner to handle the $\operatorname{sgn}$ function.
pres. = preserving,
sim. = similarity.
BB = bit balance, BU = bit uncorrelation, BMIM = bit mutual information minimization,
BKB = bucket balance.
H = Hamming distance,
WH = weighted Hamming distance,
SH = spherical Hamming distance,
C = Cosine,
E = Euclidean distance,
DNN = deep neural networks;
Drop = drop the $\operatorname{sgn}$ operator in the hash function,
Sigmoid = Sigmoid relaxation,
$[a,b]$ = $[a,b]$ bounded relaxation,
Tanh = Tanh relaxation,
Discretize = drop the $\operatorname{sgn}$ operator in the hash function
and regard the hash code as a discrete approximation
of the hash value,
Keep = optimization without relaxation for $\operatorname{sgn}$,
Two-step = two-step optimization.
}
\vspace{-.3cm}
\label{table:summary}
\centering
\begin{tabular}{|@{~}c@{~}|@{~}l@{~}||@{~}c@{~}|@{~}c@{~}|@{~}c@{~}||@{~}c@{~}||@{~}c@{~}|}
\hline
\multicolumn{2}{|c||}{Approach} & Similarity pres. & Code balance  & Hash function & Code sim. & $\operatorname{sgn}$ \\
 \hline
 {\multirow{20}{*}{\rotatebox[origin=c]{0}{Pairwise}}} & Spectral hashing~\cite{WeissTF08} $(2008)$&
 {\multirow{9}{*}{\rotatebox[origin=c]{0}{ $s_{ij}^od_{ij}^h$}}}
 & BB + BU & Eigenfunction &  {\multirow{9}{*}{\rotatebox[origin=c]{0}{H}}}  & Drop\\
& ICA hashing~\cite{HeCRB11} $(2011)$ &   & BB + BMIM   & Linear & & Drop \\
& Kernelized spectral hashing~\cite{HeLC10} $(2010)$&
  & BB + BU   & Kernel & & Drop \\
& Hashing with graphs~\cite{LiuWKC11} $(2011)$& & BB + BU  & Eigenfunction &  & Drop \\
& Discrete graph hashing~\cite{LiuMKC14} $(2014)$&  & BB + BU   & Kernel & & Discretize \\
& Kernelized Discrete graph hashing~\cite{ShiXCZXY16} $(2016)$&  & BB + BU   & Kernel & & Discretize \& Two-step\\
& Self-taught hashing~\cite{ZhangWCL10b} $(2010)$&
  & BB + BU & Linear &  & Two-step\\
& LDA hashing~\cite{StrechaBBF12} $(2012)$ &
  & BU   & Linear & & Drop\\
& Minimal loss hashing~\cite{NorouziF11} $(2011)$ &
  & -   & Linear & & Keep \\
  &Deep supervised hashing~\cite{Liu0SC16} $(2016)$&
    & -   & DNN & & Drop \\
\cline{2-7}
& Semi-supervised hashing~\cite{WangKC10a,WangKC10b,WangKC12} $(2010)$&
$s_{ij}^os_{ij}^h$ & BB+BU & Linear & H  & Drop\\
\cline{2-7}
& Topology preserving hashing~\cite{ZhangZTGLT13} $(2013)$ &
$d_{ij}^od_{ij}^h + s_{ij}^od_{ij}^h$ & BB+BU    & Linear & H & Drop\\
\cline{2-7}
& Binary reconstructive embedding~\cite{KulisD09} $(2009)$ &
\multirow{3}{*}{$(d_{ij}^o-d_{ij}^h)^2$} & {-}  & Kernel & \multirow{3}{*}{H} & Keep \\
& {ALM or NMF based hashing~\cite{MukherjeeRIHS15} $(2015)$ } &
 &-   & Linear &  & $[0,1]$ + two-step\\
 & Compressed hashing~\cite{LinJCYL13}$(2013)$ &
  & BB   & Kernel & & Drop \\
\cline{2-7}
& Supervised hashing with kernels~\cite{LiuWJJC12} $(2012)$ &
{\multirow{7}{*}{\rotatebox[origin=c]{0}{$(s_{ij}^o-s_{ij}^h)^2$}}}  & -   & Kernel & {\multirow{7}{*}{\rotatebox[origin=c]{0}{H}}} & Sigmoid\\
& Bilinear hyperplane hashing~\cite{LiuWMKC12} $(2012)$&
 & -  & BiLinear & & Sigmoid  \\
& Label-regularized maximum margin hashing~\cite{MuSY10} $(2010)$ &
& BB  & Kernel && Drop \\
& {Scalable graph hashing~\cite{JiangL15} $(2015)$ }&
& BU   & Kernel & & Drop\\
& Binary hashing~\cite{DoDNC16} $(2016)$ &
& -   & Kernel & & Two-step\\
& {CNN hashing~\cite{XiaPLLY14} $(2014)$ }&
& -   &  {DNN} & & $[-1,1]$\\
& Multi-dimensional spectral hashing~\cite{WeissFT12} $(2012)$& & BI + BU & Eigenfunction & WH & Drop \\
\cline{2-7}
& Spec hashing~\cite{LinRY10} $(2010)$&
$\operatorname{KL}(\{\bar{s}_{ij}^o\}, \{\bar{s}_{ij}^h\})$ & -  & Decision stump & H  & Two-step\\
 \hline
 \hline
 {\multirow{5}{*}{\rotatebox[origin=c]{0}{Multiwise}}} &
 Order preserving hashing~\cite{WangWYL13} $(2013)$ & Rank order &  BKB   & Linear  & H & Sigmoid\\
  \cline{2-7}
 &
   {Top rank supervised binary coding~\cite{SongLJMS15} $(2015)$}& \multirow{5}{*}{Triplet loss} &  -   & Linear  & \multirow{5}{*}{H} & Tanh\\
& Triplet loss hashing~\cite{NorouziFS12}  $(2012)$&  & -  &  Linear + NN &  & Keep \\
&  {Deep semantic ranking based hashing~\cite{ZhaoHWT15} $(2015)$}  &  & -    &  {DNN} & & Sigmoid \\
&  {Simultaneous Feature Learning and Hash Coding~\cite{LaiPLY15} $(2015)$}  &  & -   &   {DNN} &   & Drop \\
& Listwise supervision hashing~\cite{WangLSJ13} $(2013)$ &  & BU   & Linear & & Drop\\
 \hline
 \hline
 {\multirow{3}{*}{\rotatebox[origin=c]{0}{Implicit}}} &
 Picodes~\cite{BergamoTF11} $(2011)$ & \multirow{5}{*}{-} &  -   &  Linear  & H & Keep \\
   &
    Random maximum margin hashing~\cite{JolyB11} $(2011)$ &  &  BB   & Kernel   & H & Keep \\
   & Complementary projection hashing~\cite{JinHLZLCL13} $(2013)$ &    &   BB+BU  &  Kernel & H & Drop \\
  &  Spherical hashing~\cite{HeoLHCY12} $(2012)$ &    &  BB   & Spherical  & SH & Keep \\
 \hline
 \hline
 {\multirow{13}{*}{\rotatebox[origin=c]{0}{Quantization}}} &
Isotropic hashing~\cite{KongL12a} $(2012)$ & $ \approx ||\mathbf{x}-\mathbf{y}||_2$ & BU   & Linear  & H & Drop\\
   \cline{2-7}
& Iterative quantization~\cite{GongL11,GongLGP13} $(2011)$ &  {\multirow{6}{*}{\rotatebox[origin=c]{0}{$||\mathbf{x}-\mathbf{y}||_2$}}}  & -   &
{\multirow{4}{*}{\rotatebox[origin=c]{0}{Linear }}} &  {\multirow{3}{*}{\rotatebox[origin=c]{0}{H}}}  & Keep \\
& Harmonious hashing~\cite{XuBLCHC13} $(2013)$&   & BB+BU   &  && Drop \\
& Matrix hashing~\cite{GongKRL13} $(2013)$ &  &  - & &  &  Keep \\
& Angular quantization~\cite{GongKVL12} $(2012)$&  & -  &  & C & Keep \\
& {Deep hashing~\cite{LiongLWMZ15} $(2015)$} &  &  BB+BU & {DNN} &  H & Discretize\\
& {Hashing with binary deep
neural network~\cite{DoDC16} $(2016)$} &  &  BB+BU & {DNN} &  H & Discretize\\
  \cline{2-7}
& Product quantization (PQ)~\cite{JegouDS11} $(2011)$ & {\multirow{6}{*}{\rotatebox[origin=c]{0}{$||\mathbf{x}-\mathbf{y}||_2$}}}   & -  & {\multirow{6}{*}{\rotatebox[origin=c]{0}{Nearest vector}}}  & {\multirow{6}{*}{\rotatebox[origin=c]{0}{E}}}& -\\
& Cartesian $k$-means~\cite{NorouziF13} (Optimized PQ~\cite{GeHK013}) $(2013)$ &   & -  &  & & -\\
& Composite quantization~\cite{ZhangDW14} $(2014)$ &   &  -  &  &  &-\\
& {Additive quantization~\cite{BabenkoK14}} $(2014)$ &   &  -  &  &  &-\\
& {Revisiting additive quantization~\cite{MartinezCHL16}} $(2016)$ &   &  -  &  &  &-\\
& {Quantized sparse representations~\cite{JainPGZJ16}} $(2016)$ &   &  -  &  &  &-\\
\cline{2-7}
 &
  Supervised discrete hashing~\cite{ShenSLS15} $(2015)$ & {\multirow{2}{*}{\rotatebox[origin=c]{0}{$||\mathbf{x}-\mathbf{y}||_2$}}} &  -   & Kernel  & H & Discretize \\
  &
  Supervised quantization~\cite{WangZQTW16} $(2016)$ & &  -   & Nearest vector & E & - \\
\hline
\end{tabular}\vspace{-.3cm}
\end{table*}

\section{Pairwise Similarity Preserving}
\label{sec:pairwise}
The algorithms aligning the distances or similarities
of a pair of items
computed from the input space and the Hamming coding space
are roughly divided in the following groups:

\begin{itemize}
\item Similarity-distance product minimization (SDPM):
$\operatorname{min} \sum_{(i,j) \in \mathcal{E}} s_{ij}^o d_{ij}^h$.
The distance in the coding space is expected to be smaller if the similarity in the original space is larger.
Here $\mathcal{E}$ is a set of pairs of items that are considered.

\item Similarity-similarity product maximization (SSPM):
$\operatorname{max} \sum_{(i,j) \in \mathcal{E}} s_{ij}^os_{ij}^h$.
The similarity in the coding space is expected to be larger if the similarity in the original space is larger.

\item Distance-distance product maximization (DDPM):
$\operatorname{max} \sum_{(i,j) \in \mathcal{E}} d_{ij}^od_{ij}^h$.
The distance in the coding space is expected to be larger if the distance in the original space is larger.

\item Distance-similarity product minimization (DSPM):
$\operatorname{min} \sum_{(i,j) \in \mathcal{E}} d_{ij}^os_{ij}^h$.
The similarity in the coding space is expected to be smaller if the distance in the original space is larger.

\item Similarity-similarity difference minimization (SSDM):
$\operatorname{min} \sum_{(i,j) \in \mathcal{E}} (s_{ij}^o-s_{ij}^h)^2$.
The difference between the similarities is expected to be as small as possible.

\item Distance-distance difference minimization (DDDM):
$\operatorname{min} \sum_{(i,j) \in \mathcal{E}} (d_{ij}^o - d_{ij}^h)^2$.
The difference between the distances is expected to be as small as possible.

\item Normalized similarity-similarity divergence minimization (NSSDM):\\
$\operatorname{min~KL}(\{\bar{s}_{ij}^o\}, \{\bar{s}_{ij}^h\}) = \operatorname{min} (-\sum_{(i,j) \in \mathcal{E}}\bar{s}_{ij}^o \operatorname{log} \bar{s}_{ij}^h)$.
Here $\bar{s}_{ij}^o$ and $\bar{s}_{ij}^h$
are normalized similarities in the input space and the coding space:
$\sum_{ij}\bar{s}_{ij}^o = 1$ and
$\sum_{ij}\bar{s}_{ij}^h = 1$.

\end{itemize}

The following reviews these groups of algorithms
except the distance-similarity product minimization group for
which we are not aware of any algorithm belonging to.
It should be noted that merely optimizing the above similarity preserving function,
e.g., SDPM and SSPM,
is not enough and may lead to trivial solutions,
and it is necessary to combine other constraints,
which is detailed in the following discussion.
We also point out the relation
between similarity-distance product minimization
and similarity-similarity product maximization,
the relation between similarity-similarity product maximization
and similarity-similarity difference minimization,
as well as
the relation between distance-distance product maximization
and distance-distance difference minimization.

\subsection{Similarity-Distance Product Minimization}
We first introduce spectral hashing and its extensions,
and then review other forms.
\subsubsection{Spectral Hashing}
The goal
of spectral hashing~\cite{WeissTF08} is to minimize $ \sum_{(i,j) \in \mathcal{E}} s_{ij}^o d_{ij}^h$,
where the Euclidean distance in the hashing space,
$d_{ij}^h = \|\mathbf{y}_i - \mathbf{y}_j\|_2^2$,
is used
for formulation simplicity and optimization convenience,
and the similarity in the input space is defined as:
$s_{ij}^o = \exp{(-\frac{\|\mathbf{x}_i - \mathbf{x}_j\|_2^2}{2\sigma^2})}$.
Note that the Hamming distance in the search stage
can be still used for higher efficiency
as the Euclidean distance and the Hamming distance
in the coding space are consistent:
the larger the Euclidean distance, the larger the Hamming distance.
The objective function can be written in a matrix form,
\begin{align}
\operatorname{min} \sum\nolimits_{(i,j) \in \mathcal{E}} s_{ij}^o d_{ij}^h
= \operatorname{trace}(\mathbf{Y}(\mathbf{D}-\mathbf{S})\mathbf{Y}^\top),
\label{eqn:spectralhashingobjectivefunction}
\end{align}
where $\mathbf{Y} = [\mathbf{y}_1~\mathbf{y}_2\cdots\mathbf{y}_N]$
is a matrix of $M \times N$,
$\mathbf{S} = [s^o_{ij}]_{N \times N}$ is the similarity matrix,
and $\mathbf{D} = \operatorname{diag}(d_{11}, \cdots, d_{NN})$
is a diagonal matrix,
$d_{nn} = \sum_{i=1}^N s^o_{ni}$.

There is a trivial solution to the problem~(\ref{eqn:spectralhashingobjectivefunction}):
$\mathbf{y}_1 = \mathbf{y}_2 = \cdots = \mathbf{y}_N$.
To avoid it,
the code balance condition is introduced:
the number of data items mapped to each hash code is the same.
Bit balance and bit uncorrelation are used
to approximate the code balance condition.
Bit balance means that
each bit has about $50\%$ chance of being $1$ or $-1$.
Bit uncorrelation means that different bits are uncorrelated.
The two conditions are formulated as,
\begin{align}
\mathbf{Y}\mathbf{1} = 0,~~~~
\mathbf{Y} \mathbf{Y}^\top = \mathbf{I},
\end{align}
where $\mathbf{1}$ is an $N$-dimensional all-$1$ vector,
and $\mathbf{I}$ is an identity matrix of size $N$.

Under the assumption of separate multi-dimensional uniform data distribution,
the hashing algorithm is given as follows,
\begin{enumerate}
	\item Find the principal components of the $N$ $d$-dimensional reference data items using principal component analysis (PCA).
	\item Compute the $M$ one-dimensional Laplacian eigenfunctions with the $M$ smallest eigenvalues
	along each PCA direction ($d$ directions in total).
	\item Pick the $M$ eigenfunctions with the smallest eigenvalues among $Md$ eigenfunctions.
	\item Threshold the eigenfunction at zero, obtaining the binary codes.
\end{enumerate}

The one-dimensional Laplacian eigenfunction for
the case of uniform distribution on $[r_l, r_r]$
is $\phi_m(x) = \sin(\frac{\pi}{2} + \frac{m\pi}{r_r - r_l}x)$,
and the corresponding eigenvalue is
$\lambda_m = 1 - \exp{(-\frac{\epsilon^2}{2} |\frac{m\pi}{r_r - r_l}|^2)}$,
where $m$ ($= 1, 2, \cdots$) is the frequency
and $\epsilon$ is a fixed small value.
The hash function is formally written as
$h(\mathbf{x}) = \operatorname{sgn}(\sin(\frac{\pi}{2} + \gamma \mathbf{w}^\top\mathbf{x}))$,
where $\gamma$ depends on the frequency $m$
and the range of the projection along the direction $\mathbf{w}$.

\textbf{Analysis:}
In the case the spreads along the top $M$ PCA directions are the same,
the hashing algorithm
partitions each direction into two parts
using the median (due to the bit balance requirement) as the threshold,
which is equivalent to thresholding at the mean value
under the assumption of
uniform data distributions.
In the case that the true data distribution is a multi-dimensional isotropic Gaussian distribution,
the algorithm is equivalent to
two quantization algorithms:
iterative quantization~\cite{GongLGP13},~\cite{GongL11} and isotropic hashing~\cite{KongL12a}.

Regarding the performance,
this method performs
well for a short hash code
but poor for a long hash code.
The reason includes three aspects.
First,
the assumption
that
the data follow a uniform distribution
does not hold in real cases.
Second,
the eigenvalue monotonously decreases
with respect to $|\frac{m}{r_r - r_l}|^2$,
which means that
the PCA direction with a large spread ($|r_r - r_l|$)
and a lower frequency ($m$) is preferred.
Hence there might be more than one eigenfunction picked along a single PCA direction,
which breaks the uncorrelation requirement.
Last,
thresholding the eigenfunction
$\phi_m(x) = \sin(\frac{\pi}{2} + \frac{m\pi}{r_r - r_l}x)$
at zero
leads to
that near points may be mapped to different hash values
and farther points may be mapped to the same hash value.
As a result,
the Hamming distance is not well consistent to the distance in the input space.

\textbf{Extensions:}
There are some extensions using PCA.
(1) \emph{Principal component hashing}~\cite{MatsushitaW09}
uses the principal direction
to formulate the hash function;
(2) \emph{Searching with expectations}~\cite{SandhawaliaJ10}
and~\emph{transform coding}~\cite{Brandt10}
that transforms the data using PCA
and then adopts
the rate distortion optimization (bits allocation) approach
to determine which principal direction is used
and how many bits are assigned to such a direction;
(3) \emph{Double-bit quantization}
that handles the third drawback in spectral hashing
by distributing two bits into each projection direction,
conducting only $3$-cluster quantization,
and assigning $01$, $00$, and $11$
to each cluster.
Instead of PCA, \emph{ICA hashing}~\cite{HeCRB11}
adopts independent component analysis for hashing
and uses bit balance and bit mutual information minimization
for code balance.

There are many other extensions
in a wide range,
including similarity graph extensions~\cite{LiWCXL13},
\cite{ZhuangLWZS11},\cite{LiuSXWZ13},\cite{LiuWKC11},\cite{LiuMKC14},\cite{LinJCYL13},
\cite{ShiXCZXY16},\cite{ZhangLGZ16},
hash function extensions~\cite{HeLC10},\cite{ShaoWOZ12},
weighted Hamming distance~\cite{WangZS13},
self-taught hashing~\cite{ZhangWCL10b},
sparse hash codes~\cite{ZhuHCCS13},
discrete hashing~\cite{YangSSLL15},
and so on.

\subsubsection{Variants}
\emph{Linear discriminant analysis (LDA) hashing}~\cite{StrechaBBF12}
minimizes a form of the loss function:
$\operatorname{min} \sum_{(i,j) \in \mathcal{E}} s_{ij}^o d_{ij}^h$,
where $d_{ij}^h = \|\mathbf{y}_i - \mathbf{y}_j\|_2^2$.
Different from spectral hashing,
(1) $s_{ij}^o = 1$ if data items $\mathbf{x}_i$ and $\mathbf{x}_j$ are a similar pair,
$(i,j) \in \mathcal{E}^+$,
and $s_{ij}^o = -1$ if data items $\mathbf{x}_i$ and $\mathbf{x}_j$ are a dissimilar pair,
$(i,j) \in \mathcal{E}^-$
(2) a linear hash function is used:
$\mathbf{y} = \operatorname{sgn}(\mathbf{W}^\top\mathbf{x} + \mathbf{b})$,
and (3) a weight $\alpha$ is imposed to
$s_{ij}^o d_{ij}^h$ for the similar pair.
As a result,
the objective function is written as:
\begin{align}
\alpha\sum\nolimits_{(i,j)\in \mathcal{E}^+} \|\mathbf{y}_i - \mathbf{y}_j\|_2^2
- \sum\nolimits_{(i,j)\in \mathcal{E}^-} \|\mathbf{y}_i - \mathbf{y}_j\|_2^2.
\label{eqn:ldahashingobjectivefunction}
\end{align}

The projection matrix $\mathbf{W}$ and the threshold $\mathbf{b}$
are separately optimized:
(1) to estimate the orthogonal matrix $\mathbf{W}$, drop the $\operatorname{sgn}$ function in Equation~(\ref{eqn:ldahashingobjectivefunction}),
leading to an eigenvalue decomposition problem;
(2) estimate $\mathbf{b}$ by minimizing Equation~(\ref{eqn:ldahashingobjectivefunction})
with fixed $\mathbf{W}$
through a simple $1$D search scheme.
A similar loss function,
contrastive loss, is adopted in~\cite{DaiLWJ16}
with a different optimization technique.

The loss function in \emph{minimal loss hashing}~\cite{NorouziF11}
is in the form of $\operatorname{min} \sum_{(i,j) \in \mathcal{E}} s_{ij}^o d_{ij}^h$.
Similar to LDA hashing,
$s_{ij}^o = 1$ if $(i,j) \in \mathcal{E}^+$
and $s_{ij}^o = -1$ if $(i,j) \in \mathcal{E}^-$.
Differently,
the distance is hinge-like:
$d^h_{ij} = \max(\|\mathbf{y}_i - \mathbf{y}_j\|_1 + 1, \rho)$ for $(i,j) \in \mathcal{E}^+$
and
$d^h_{ij} = \min(\|\mathbf{y}_i - \mathbf{y}_j\|_1 - 1, \rho)$ for $(i,j) \in \mathcal{E}^-$.
The intuition is
that there is no penalty if the Hamming distance
for similar pairs is small enough
and if the Hamming distance for dissimilar pairs is large enough.
The formulation, if $\rho$ is fixed,
is equivalent to,
\begin{align}
\min &~\sum\nolimits_{(i,j) \in \mathcal{E}^+}  \max(\|\mathbf{y}_i - \mathbf{y}_j\|_1 - \rho + 1, 0) \nonumber \\
&~ + \sum\nolimits_{(i,j) \in \mathcal{E}^-}  \lambda \max(\rho - \|\mathbf{y}_i - \mathbf{y}_j\|_1 + 1, 0),
\end{align}
where $\rho$ is a hyper-parameter
used as a threshold
in the Hamming space
to differentiate similar pairs from dissimilar pairs,
$\lambda$ is another hyper-parameter
that controls the ratio
of the slopes
for the penalties
incurred for similar (or dissimilar) points.
The hash function is in the linear form:
$\mathbf{y} = \operatorname{sgn}(\mathbf{W}^\top\mathbf{x})$.
The projection matrix $\mathbf{W}$
is estimated
by transforming $\mathbf{y} = \operatorname{sgn}(\mathbf{W}^\top\mathbf{x}) = \arg\max_{\mathbf{y}' \in \mathcal{H}} \mathbf{h}'^\top\mathbf{W}^\top\mathbf{x}$
and optimizing using structured prediction with latent variables.
The hyper-parameters $\rho$ and $\lambda$ are chosen
via cross-validation.

\textbf{Comments:}
Besides the optimization techniques,
the main differences
of the three representative algorithms,
i.e., spectral hashing, LDA hashing, and minimal loss hashing,
are twofold.
First,
the similarity in the input space in spectral hashing
is defined as a continuous positive number
computed from the Euclidean distance,
while in LDA hashing and minimal loss hashing
the similarity is set to $1$ for a similar pair and $-1$ for a dissimilar pair.
Second,
the distance in the hashing space for minimal loss hashing
is different from spectral hashing and LDA hashing.

\subsection{Similarity-Similarity Product Maximization}
\emph{Semi-supervised hashing}~\cite{WangKC10a},~\cite{WangKC10b},~\cite{WangKC12}
is the representative algorithm
in this group.
The objective function is $\operatorname{max} \sum_{(i,j)\in \mathcal{E}}s_{ij}^os_{ij}^h$.
The similarity $s_{ij}^o$ in the input space
is $1$
if the pair of items $\mathbf{x}_i$ and $\mathbf{x}_j$ belong to the same class or
are nearby points,
and $-1$ otherwise.
The similarity in the coding space is defined as
$s_{ij}^h = \mathbf{y}_i^\top\mathbf{y}_j$.
Thus, the objective function is rewritten as maximizing:
\begin{align}
\sum\nolimits_{(i,j) \in \mathcal{E}}s^o_{ij} \mathbf{y}_i^\top\mathbf{y}_j.
\end{align}
The hash function is in a linear form
$\mathbf{y} = \mathbf{h}(\mathbf{x}) = \operatorname{sgn}(\mathbf{W}^\top\mathbf{x})$.
Besides,
the bit balance is also considered,
and is formulated
as maximizing the variance, $\operatorname{trace}(\mathbf{Y}\mathbf{Y}^\top)$,
rather than letting the mean be $0$,
$\mathbf{Y}\mathbf{1} = 0$.
The overall objective is to maximize
\begin{align}
\operatorname{trace}(\mathbf{Y}\mathbf{S}\mathbf{Y}^\top) + \eta \operatorname{trace}(\mathbf{Y}\mathbf{Y}^\top),
\label{eqn:objectiveFunctionForSSH}
\end{align}
subject to
$\mathbf{W}^\top\mathbf{W} = \mathbf{I}$,
which is a relaxation of the bit uncorrelation condition.
The estimation of $\mathbf{W}$ is done
by directly dropping the $\operatorname{sgn}$ operator.

An unsupervised extension is given in~\cite{WangKC12}:
sequentially compute the projection vector $\{\mathbf{w}_m\}_{m=1}^M$
from $\mathbf{w}_1$ to $\mathbf{w}_M$
by optimizing the problem \ref{eqn:objectiveFunctionForSSH}.
In particular,
the first iteration computes the PCA direction as the first $\mathbf{w}$,
and at each of the later iterations,
$s_{ij}^o=1$ if nearby points are mapped to different hash values in the previous iterations,
and $s_{ij}^o=-1$ if far points are mapped to same hash values in the previous iterations.
An extension of the semi-supervised hashing to nonlinear hash functions
is presented in~\cite{WuZCCB13}
using the kernel hash function.
An iterative two-step optimization using graph cuts is given in~\cite{GeH014}.

\textbf{Comments:}
It is interesting to note that
$\sum_{(i,j) \in \mathcal{E}}s^o_{ij} \mathbf{y}_i^\top\mathbf{y}_j =
\texttt{const} - \frac{1}{2}\sum_{(i,j) \in \mathcal{E}}s^o_{ij} \|\mathbf{y}_i - \mathbf{y}_j\|_2^2
= \texttt{const} - \frac{1}{2}\sum_{(i,j) \in \mathcal{E}}s^o_{ij} d_{ij}^h$
if $\mathbf{y} \in \{1, -1\}^M$,
where $\texttt{const}$ is a constant variable
(and thus $\operatorname{trace}(\mathbf{Y}\mathbf{S}\mathbf{Y}^\top)
= \texttt{const} - \operatorname{trace}(\mathbf{Y}(\mathbf{D} - \mathbf{S})\mathbf{Y}^\top)$).
In this case,
similarity-similarity product maximization
is equivalent to
similarity-distance product minimization.

\subsection{Distance-Distance Product Maximization}
The mathematical formulation
of distance-distance product maximization
is $\operatorname{max}\sum_{(i,j)\in \mathcal{E}}d_{ij}^od_{ij}^h$.
Topology preserving hashing~\cite{ZhangZTGLT13} formulates the objective function
by starting with this rule:
\begin{align}
\sum\nolimits_{i,j}d^o_{ij}d^h_{ij}
= \sum\nolimits_{i,j}d^o_{ij}\|\mathbf{y}_i - \mathbf{y}_j\|_2^2
= \operatorname{trace}(\mathbf{Y}\mathbf{L}_d\mathbf{Y}^\top),
\end{align}
where $\mathbf{L}_d = \operatorname{Diag}\{\mathbf{D}^o\mathbf{1}\} - \mathbf{D}^o$
and $\mathbf{D}^o = [d^o_{ij}]_{N \times N }$.

In addition,
similarity-distance product minimization is also considered:
\begin{align}
\sum\nolimits_{(i,j) \in \mathcal{E} }s_{ij} \|\mathbf{y}_i - \mathbf{y}_j\|_2^2
=\operatorname{trace}(\mathbf{Y}\mathbf{L}\mathbf{Y}^\top).
\end{align}

The overall formulation is given as follows,
\begin{align}
\max \frac{\operatorname{trace}(\mathbf{Y}(\mathbf{L}_d + \alpha \mathbf{I})\mathbf{Y}^\top)}
{\operatorname{trace}(\mathbf{Y}\mathbf{L}\mathbf{Y}^\top)},
\end{align}
where $\alpha\mathbf{I}$ is introduced as a regularization term,
$\operatorname{trace}(\mathbf{Y}\mathbf{Y}^\top)$,
maximizing the variances,
which is the same for semi-supervised hashing~\cite{WangKC10a}
for bit balance.
The problem is optimized by dropping the $\operatorname{sgn}$ operator
in the hash function $\mathbf{y} = \operatorname{sgn}(\mathbf{W}^\top\mathbf{x})$
and
letting $\mathbf{W}^\top\mathbf{X}\mathbf{L}\mathbf{X}^\top\mathbf{W}$ be an identity matrix.

\subsection{Distance-Distance Difference Minimization}
\emph{Binary reconstructive embedding}~\cite{KulisD09}
belongs to this group:
$\operatorname{min}\sum_{(i,j) \in \mathcal{E}}(d_{ij}^o - d_{ij}^h)^2$.
The Euclidean distance is used
in both the input and coding spaces.
The objective function is formulated as follows,
\begin{align}
\min\sum\nolimits_{(i,j) \in \mathcal{E}} (\frac{1}{2} \|\mathbf{x}_i - \mathbf{x}_j\|_2^2 - \frac{1}{M} \|\mathbf{y}_i - \mathbf{y}_j\|_2^2)^2.\label{eqn:objectiveFunctionForBRE}
\end{align}
The kernel hash function is used:
\begin{align}
y_{nm} = h_m(\mathbf{x})
= \operatorname{sgn}(\sum\nolimits_{t=1}^{T_m} w_{mt} K(\mathbf{s}_{mt}, \mathbf{x})),
\end{align}
where $\{\mathbf{s}_{mt}\}_{t=1}^{T_m}$ are sampled data items,
$K(\cdot,\cdot)$ is a kernel function,
and $\{w_{mt}\}$ are the weights
to be learnt.

Instead of relaxing or dropping the $\operatorname{sgn}$ function,
a coordinate descent optimization scheme is presented in~\cite{KulisD09}:
fix all but one weight $w_{mt}$
and optimize the problem~\ref{eqn:objectiveFunctionForBRE}
with respect to $w_{mt}$.
There is
an exact, optimal update to this weight $w_{mt}$
(fixing all the other weights),
which is achieved with the time complexity $O(N\log N + |\mathcal{E}|)$.
Alternatively, a two-step solution is presented in~\cite{MukherjeeRIHS15}:
first relax the binary variables to $(0,1)$ and
optimize the problem via an augmented Lagrangian formulation
and a nonnegative matrix factorization formulation.

\textbf{Comments:}
We have the following equation,
\begin{align}
&~\operatorname{min}\sum\nolimits_{(i,j) \in \mathcal{E}}(d_{ij}^o - d_{ij}^h)^2 \\
=&~ \operatorname{min}\sum\nolimits_{(i,j) \in \mathcal{E}}((d_{ij}^o)^2 + (d_{ij}^h)^2 - 2 d_{ij}^o d_{ij}^h) \\
=&~ \operatorname{min}\sum\nolimits_{(i,j) \in \mathcal{E}}((d_{ij}^h)^2 - 2 d_{ij}^o d_{ij}^h).
\end{align}
This shows that the difference between distance-distance difference minimization
and distance-distance product maximization
lies on $\operatorname{min} \sum_{(i,j) \in \mathcal{E}}(d_{ij}^h)^2$,
minimizing the distances between the data items in the hash space.
This could be regarded as a regularizer,
complementary to distance-distance product maximization $\operatorname{max}\sum_{(i,j) \in \mathcal{E}} d_{ij}^o d_{ij}^h$
which tends to maximize the distances between the data items in the hash space.

\subsection{Similarity-Similarity Difference Minimization}
Similarity-similarity difference minimization
is mathematically
formulated as
$\operatorname{min}\sum_{(i,j) \in \mathcal{E}}(s_{ij}^o-s_{ij}^h)^2$.
\emph{Supervised hashing with kernels}~\cite{LiuWJJC12},
one representative approach in this group,
aims to minimize an objective function,
\begin{align}
\min\sum\nolimits_{(i,j) \in \mathcal{E}} (s^o_{ij}  - \frac{1}{M}\mathbf{y}_i^\top\mathbf{y}_j)^2,\label{eqn:objectiveFunctionForHashwithKernelsWeiLiu}
\end{align}
where $s^o_{ij} = 1$ if $(i,j)$ is similar,
and $s^o_{ij} = -1$ if it is dissimilar.
$\mathbf{y} = \mathbf{h}(\mathbf{x})$
is a kernel hash function.
Kernel reconstructive hashing~\cite{YangBZRZC14}
extends this technique
using a normalized Gaussian kernel similarity.
Scalable graph hashing~\cite{JiangL15}
uses the feature transformation to
approximate the similarity matrix (graph)
without explicitly computing the similarity matrix.
Binary hashing~\cite{DoDNC16} solves the problem
using a two-step approach,
in which the first step adopts semi-definite relaxation
and augmented lagrangian to estimate the discrete labels.

\textbf{Comments:}
We have the following equation,
\begin{align}
&~ \operatorname{min}\sum\nolimits_{(i,j) \in \mathcal{E}}(s_{ij}^o-s_{ij}^h)^2 \\
=&~ \operatorname{min}\sum\nolimits_{(i,j) \in \mathcal{E}}((s_{ij}^o)^2 + (s_{ij}^h)^2 - 2 s_{ij}^o s_{ij}^h) \\
=&~ \operatorname{min}\sum\nolimits_{(i,j) \in \mathcal{E}}((s_{ij}^h)^2 - 2 s_{ij}^o s_{ij}^h).
\end{align}
This shows that the difference between similarity-similarity difference minimization
and similarity-similarity product maximization
lies in $\operatorname{min} \sum_{(i,j) \in \mathcal{E}}(s_{ij}^h)^2$,
minimizing the similarities between the data items in the hash space,
intuitively
letting the hash codes be as different as possible.
This could be regarded as a regularizer
complementary to similarity-similarity product maximization $\operatorname{max}\sum_{(i,j) \in \mathcal{E}} s_{ij}^o s_{ij}^h$,
which has a trivial solution: the hash codes are the same for all data points.

\textbf{Extensions and variants:}
\emph{Multi-dimensional spectral hashing}~\cite{WeissFT12}
uses a similar objective function,
but with a weighted Hamming distance,
\begin{align}
\min ~&~ \sum\nolimits_{(i,j) \in \mathcal{E}} (s^o_{ij} - \mathbf{y}_i^\top\boldsymbol{\Lambda}\mathbf{y}_j)^2,
\label{eqn:objectiveFunctionForMultidimensionalSpectralHashing}
\end{align}
where $\boldsymbol{\Lambda}$ is a diagonal matrix.
Both $\boldsymbol{\Lambda}$ and hash codes $\{\mathbf{y}_i\}$ are needed to be optimized.
The algorithm for solving the problem~\ref{eqn:objectiveFunctionForMultidimensionalSpectralHashing}
to compute the hash codes
is similar to that given in spectral hashing~\cite{WeissTF08}.
\emph{Bilinear hyperplane hashing}~\cite{LiuWMKC12} extends
the formulation of supervised hashing with kernels
by introducing a bilinear hyperplane hashing function.
\emph{Label-regularized maximum margin hashing}~\cite{MuSY10}
formulates the objective function
from three components:
the similarity-similarity difference,
a hinge loss from the hash function,
and the maximum margin part.

\subsection{Normalized Similarity-Similarity Divergence Minimization}

Spec hashing~\cite{LinRY10}, belonging to this group,
views each pair of data items as a sample
and their (normalized) similarity
as the probability,
and finds the hash functions
so that
the probability distributions
from the input space and the coding space
are well aligned.
The objective function is written as follows,
\begin{align}
\operatorname{KL}(\{\bar{s}_{ij}^o\}, \{\bar{s}_{ij}^h\}) = \texttt{const} -\sum\nolimits_{(i,j)\in\mathcal{E}}\bar{s}_{ij}^o \operatorname{log} \bar{s}_{ij}^h.
\end{align}
Here,
$\bar{s}_{ij}^o$ is the normalized similarity in the input space,
$\sum_{ij}\bar{s}_{ij}^o = 1$.
$\bar{s}^h_{ij}$
is the normalized similarity
in the Hamming space,
$\bar{s}^h_{ij} = \frac{1}{Z} \exp{(-\lambda d^h_{ij})}$,
where $Z$ is a normalization variable
$Z = \sum_{ij} \exp{(-\lambda d^h_{ij})}$.

Supervised binary hash code learning~\cite{Fan13}
presents a supervised learning algorithm
based on the Jensen-Shannon divergence
which is derived from
minimizing an upper bound
of the probability of Bayes decision errors.

\section{Multiwise Similarity Preserving}
\label{sec:multiplewise}
This section reviews the category of hashing algorithms
that formulate the loss function
by maximizing the agreement
of the similarity orders over more than two items
computed from the input space and the coding space.

\emph{Order preserving hashing}~\cite{WangWYL13}
aims to learn hash functions through aligning the orders
computed from the original space
and the ones in the coding space.
Given a data point $\mathbf{x}_n$,
the database points $\mathcal{X}$
are divided into $(M+1)$ categories,
$(\mathcal{C}^h_{n0}, \mathcal{C}^h_{n1},\cdots,\mathcal{C}^h_{nM})$,
where $\mathcal{C}^h_{nm}$
corresponds to the items whose distance to the given point is $m$,
and
$(\mathcal{C}^o_{n0}, \mathcal{C}^o_{n1},\cdots,\mathcal{C}^o_{nM})$,
using the distances in the hashing space
and the distances in the input (original) space, respectively.
$(\mathcal{C}^o_{n0}, \mathcal{C}^o_{n1},\cdots,\mathcal{C}^o_{nM})$ is constructed
such that
in the ideal case
the probability of assigning an item to any hash code
is the same.
The basic objective function maximizing the alignment between the two categories
is given as follows,
\begin{align}
L(\mathbf{h}(\cdot); \mathcal{X})
=\sum\nolimits_{n\in\{1,\cdots,N\}} \sum_{m=0}^{M} (|\mathcal{C}^o_{nm} - \mathcal{C}^h_{nm}| + |\mathcal{C}^h_{nm} - \mathcal{C}^o_{nm}|), \nonumber
\end{align}
where $|\mathcal{C}^o_{nm} - \mathcal{C}^h_{nm}|$
is the cardinality of the difference of the two sets.
The linear hash function $\mathbf{h}(\mathbf{x})$
is used and
dropping the $\operatorname{sgn}$ function is adopted
for optimization.

Instead of preserving the order,
KNN hashing~\cite{DingHFP15} directly maximizes the kNN accuracy of the search result,
which is solved by using the factorized neighborhood representation to parsimoniously model
the neighborhood relationships inherent in the training data.

\emph{Triplet loss hashing}~\cite{NorouziFS12} formulates the hashing problem
by maximizing the similarity order agreement
defined over
triplets of items,
$\{(\mathbf{x}, \mathbf{x}^+, \mathbf{x}^-)\}$,
where the pair $(\mathbf{x}, \mathbf{x}^-)$
is less similar than
the pair $(\mathbf{x}, \mathbf{x}^+)$.
The triplet loss is defined as
\begin{align}
\ell_{\text{triplet}} (\mathbf{y}, \mathbf{y}^+, \mathbf{y}^-)
= \max (1 - \|\mathbf{y} - \mathbf{y}^-\|_1 + \|\mathbf{y} - \mathbf{y}^+\|_1, 0).
\label{eqn:tripletlosshashing}
\end{align}
The objective function is given as follows,
\begin{align}
\sum_{(\mathbf{x}, \mathbf{x}^+, \mathbf{x}^-) \in \mathcal{D}}
\ell_{\text{triplet}} (\mathbf{h}(\mathbf{x}), \mathbf{h}(\mathbf{x}^+), \mathbf{h}(\mathbf{x}^-))
+ \frac{\lambda}{2} \operatorname{trace}{(\mathbf{W}^\top\mathbf{W})},\nonumber
\end{align}
where $\mathbf{h}(\mathbf{x}) = \mathbf{h}(\mathbf{x};\mathbf{W})$ is the compound hash function.
The problem is optimized
using the algorithm similar to minimal loss hashing \cite{NorouziF11}.
The extension to asymmetric Hamming distance is also discussed in~\cite{NorouziFS12}.
Binary optimized hashing~\cite{DaiLWJ16}
also uses a triplet loss function, with a slight different distance measure in the Hamming space
and a different optimization technique.

Top rank supervised binary coding~\cite{SongLJMS15} presents another form of triplet losses
in order to penalize the samples that are incorrectly
ranked at the top of a Hamming-distance ranking list more
than those at the bottom.

\emph{Listwise supervision hashing}~\cite{WangLSJ13} also uses triplets of items.
The formulation is based on a triplet tensor $\mathbf{S}^o$ defined as follows,
\begin{equation}
s^o_{ijk} = s(\mathbf{q}_i; \mathbf{x}_j, \mathbf{x}_k)
= \left\{ \begin{array}{l l}
     1 & \quad \text{if $\operatorname{s}^o(\mathbf{q}_i, \mathbf{x}_j) <  \operatorname{s}^o(\mathbf{q}_i, \mathbf{x}_k)$}\\
     -1 & \quad \text{if $\operatorname{s}^o(\mathbf{q}_i, \mathbf{x}_j) >  \operatorname{s}^o(\mathbf{q}_i, \mathbf{x}_k)$}\\
     0 & \quad \text{if $\operatorname{s}^o(\mathbf{q}_i, \mathbf{x}_j) =  \operatorname{s}^o(\mathbf{q}_i, \mathbf{x}_k)$}\\
   \end{array}. \right. \nonumber
\end{equation}
The objective is to maximize triple-similarity-triple-similarity product:
\begin{align}
\sum\nolimits_{i,j,k}s^h_{ijk}s^o_{ijk},
\end{align}
where $s^h_{ijk}$ is a ranking triplet computed by the binary code using the cosine similarity,
$s^h_{ijk} = \operatorname{sgn}(\mathbf{h}(\mathbf{q}_i)^\top\mathbf{h}(\mathbf{x}_j) - \mathbf{h}(\mathbf{q}_i)^\top\mathbf{h}(\mathbf{x}_k))$.
Through dropping the $\operatorname{sgn}$ function,
the objective function is transformed to
\begin{align}
-\sum\nolimits_{i, j, k} \mathbf{h}(\mathbf{q}_i)^\top(\mathbf{h}(\mathbf{x}_j) - \mathbf{h}(\mathbf{x}_k))s^o_{ijk},
\end{align}
which is solved by dropping the $\operatorname{sgn}$ operator in the hash function
$\mathbf{h}(\mathbf{x}) = \operatorname{sgn}(\mathbf{W}^\top\mathbf{x})$.

\textbf{Comments:}
Order preserving hashing considers the relation between the search lists
while triplet loss hashing and listwise supervision hashing consider triplewise relation.
The central ideas of triplet loss hashing and listwise supervision hashing
are very similar,
and their difference lies in how to formulate the loss function
besides the different optimization techniques they adopted.

\section{Implicit Similarity Preserving}
\label{sec:implicit}
We review the category of hashing algorithms
that focus on pursuing effective space partitioning
without explicitly
evaluating the relation between the distances/similarities
in the input and coding spaces.
The common idea is
to partition the space,
formulated as a classification problem,
with the maximum margin criterion
or the code balance condition.

\emph{Random maximum margin hashing}~\cite{JolyB11} learns
a hash function with the maximum margin criterion.
The point is that
the positive and negative labels
are randomly generated
by randomly sampling $N$ data items
and randomly labeling half of the items
with $-1$
and the other half with $1$.
The formulation is a standard SVM formulation
that is equivalent to the following form,
\begin{align}
\max \frac{1}{\|\mathbf{w}\|_2}
\min\{\min_{i=1,\cdots,\frac{N}{2}}(\mathbf{w}^\top\mathbf{x}_i^+ + b), \min_{i=1, \cdots, \frac{N}{2}}(-\mathbf{w}^\top\mathbf{x}_i^- - b)\},\nonumber
\end{align}
where $\{\mathbf{x}_i^+\}$ are the positive samples
and $\{\mathbf{x}_i^-\}$ are the negative samples.
Note that this is different from PICODES~\cite{BergamoTF11}
as random maximum margin hashing adopts
the hyperplanes learnt from SVM to form the hash functions
while PICODES~\cite{BergamoTF11}
exploits the hyperplanes to check
whether the hash codes are semantically separable
rather than forming hash functions.

\emph{Complementary projection hashing}~\cite{JinHLZLCL13},
similar to complementary hashing~\cite{XuWLZLY11},
finds the hash function
such that
the items are as far away as possible
from the partition plane corresponding to the hash function.
It is formulated
as
$\mathcal{H}(\epsilon - |\mathbf{w}^\top\mathbf{x} + b|)$,
where $\mathcal{H}(\cdot) = \frac{1}{2}(1+ \operatorname{sgn}(\cdot))$ is the unit step function.
Moreover, the bit balance condition,
$\mathbf{Y}\mathbf{1} = 0$,
and the bit uncorrelation condition,
the non-diagonal entries in $\mathbf{Y}\mathbf{Y}\top$ are $0$,
are considered.
An extension is also given
by using the kernel hash function.
In addition,
when learning the $m$th hash function,
the data item is weighted
by a variable,
which is computed
according to the previously computed $(m-1)$ hash functions.

\emph{Spherical hashing}~\cite{HeoLHCY12}
uses a hypersphere
to partition the space.
The spherical hash function is defined as
$h(\mathbf{x}) = 1$ if $d(\mathbf{p}, \mathbf{x}) \leqslant t$
and $h(\mathbf{x}) = 0$ otherwise.
The compound hash function
consists of $M$ spherical functions,
depending on
$M$ pivots $\{\mathbf{p}_1, \cdots, \mathbf{p}_M\}$
and $M$ thresholds $\{t_1, \cdots, t_M\}$.
The distance in the coding space is defined based on
the distance:
$\frac{\|\mathbf{y}_1 - \mathbf{y}_2\|_1}{ \mathbf{y}_1^T  \mathbf{y}_2}$.
Unlike the pairwise and multiwise similarity preserving algorithms,
there is no explicit function
penalizing the disagreement of the similarities
computed in the input and coding spaces.
The $M$ pivots and thresholds
are learnt
such that
it satisfies a pairwise bit balance condition:
$|\{\mathbf{x}~|~h_m(\mathbf{x}) = 1\}|
= |\{\mathbf{x}~|~h_m(\mathbf{x}) = 0\}|$,
and
$|\{\mathbf{x}~|~h_i(\mathbf{x}) = b_1, h_j(\mathbf{x}) = b_2\}|
= \frac{1}{4} |\mathcal{X}|, b_1, b_2 \in \{0, 1\}, i \neq j$.

\section{Quantization}
\label{sec:quantization}

The following provides a simple derivation
showing that
the quantization approach can be derived
from the distance-distance difference minimization criterion.
There is a similar statement in~\cite{JegouDS11}
obtained from the statistical perspective:
the distance reconstruction error is statistically bounded
by the quantization error.
Considering two points $\mathbf{x}_i$ and $\mathbf{x}_j$
and their approximations $\mathbf{z}_i$ and $\mathbf{z}_j$,
we have
\begin{align}
&~ |d^o_{ij} - d^h_{ij}|  \\
=&~ | |\mathbf{x}_i - \mathbf{x}_j|_2 - |\mathbf{z}_i - \mathbf{z}_j|_2 | \\
=&~ | |\mathbf{x}_i - \mathbf{x}_j|_2 - |\mathbf{x}_i - \mathbf{z}_j|_2 + |\mathbf{x}_i - \mathbf{z}_j|_2 - |\mathbf{z}_i - \mathbf{z}_j|_2 | \\
\leqslant &~ | |\mathbf{x}_i - \mathbf{x}_j|_2 - |\mathbf{x}_i - \mathbf{z}_j|_2 | + | |\mathbf{x}_i - \mathbf{z}_j|_2 - |\mathbf{z}_i - \mathbf{z}_j|_2 | \\
\leqslant &~ |\mathbf{x}_j - \mathbf{z}_j|_2 + |\mathbf{x}_i - \mathbf{z}_i|_2.
\end{align}
Thus, $|d^o_{ij} - d^h_{ij}|^2 \leqslant 2 (|\mathbf{x}_j - \mathbf{z}_j|^2_2 + |\mathbf{x}_i - \mathbf{z}_i|^2_2)$,
and
\begin{align}
&\min \sum\nolimits_{i, j \in \{1, 2, \cdots, N \} } |d^o_{ij} - d^h_{ij}|^2 \\
\leqslant ~& \min 2\sum\nolimits_{i, j \in \{1, 2, \cdots, N\} } (|\mathbf{x}_j - \mathbf{z}_j|^2_2 + |\mathbf{x}_i - \mathbf{z}_i|^2_2) \\
=~& \min 4\sum\nolimits_{i \in \{1, 2, \cdots, N\} }|\mathbf{x}_i - \mathbf{z}_i|^2_2.
\end{align}
This means that
the distance-distance difference minimization rule is transformed
to minimizing its upper-bound,
the quantization error,
which is described as a theorem below.

\newtheorem{Theorem}{Theorem}
\newcommand{\TheoremRef}[1]{Theorem.~#1}

\begin{Theorem}
\label{theorem1}
The distortion error
in the quantization approach
is an upper bound (with a scale)
of the differences
between the pairwise distances
computed from the input features
and from the approximate representation.
\end{Theorem}

The quantization approach for hashing
is roughly divided into two main groups:
hypercubic quantization,
in which the approximation $\mathbf{z}$ is equal to the hash code $\mathbf{y}$,
and
Cartesian quantization,
in which
the approximation $\mathbf{z}$ corresponds to a vector formed
by the hash code $\mathbf{y}$,
e.g., $\mathbf{y}$ represents the index of a set of candidate approximations.
In addition,
we will review the related reconstruction-based hashing algorithms.

\subsection{Hypercubic Quantization}
\label{sec:LTH2:HD}
Hypercubic quantization refers to
a category of algorithms
that quantize a data item
to a vertex in a hypercubic,
i.e., a vector belonging to
a set $\{[y_1~y_2~\cdots~ y_M]^\top~|~y_m \in \{-1, 1\}\}$
or the rotated hypercubic vertices.
It is in some sense related to $1$-bit compressive sensing~\cite{BoufounosB08}:
Its goal is to design a measurement matrix $\mathbf{A}$
and a recovery algorithm
such that a k-sparse unit vector $\mathbf{x}$
can be efficiently recovered from the sign of its linear measurements,
i.e., $\mathbf{b} = \operatorname{sgn}(\mathbf{A}\mathbf{x})$,
while
hypercubic quantization
aims to find the matrix $\mathbf{A}$ which is usually a rotation matrix,
and the codes $\mathbf{b}$, from the input $\mathbf{x}$.

The widely-used scalar quantization approach
with only one bit assigned to each dimension
can be viewed as a hypercubic quantization approach,
and can be derived
by minimizing
\begin{align}
||\mathbf{x}_i - \mathbf{y}_i||^2_2
\label{eqn:scalarquantizationloss}
\end{align}
subject to $\mathbf{y}_i \in \{1,-1\}$.
The local digit coding approach~\cite{KoudasOST04}
also belongs to this category.

\subsubsection{Iterative quantization}
Iterative quantization~\cite{GongL11},~\cite{GongLGP13}
preprocesses the data,
by reducing the dimension using PCA to $M$ dimensions,
$\mathbf{v} = \mathbf{P}^\top\mathbf{x}$,
where $\mathbf{P}$ is a matrix of size $d\times M$ ($M \leqslant d$)
computed using PCA,
and then finds an optimal rotation $\mathbf{R}$ followed by a scalar quantization.
The formulation is given as,
\begin{align}
\min\|\mathbf{Y} - \mathbf{R}^\top\mathbf{V}\|_F^2, \label{eqn:ObjectiveFunctionForITQ}
\end{align}
where $\mathbf{R}$ is a matrix of $M \times M$,
$\mathbf{V} = [\mathbf{v}_1 \mathbf{v}_2\cdots\mathbf{v}_N]$
and $\mathbf{Y} = [\mathbf{y}_1 \mathbf{y}_2\cdots \mathbf{y}_N]$.

The problem is solved via alternative optimization.
There are two alternative steps.
Fixing $\mathbf{R}$,
$\mathbf{Y} = \operatorname{sgn}(\mathbf{R}^\top\mathbf{V})$.
Fixing $\mathbf{B}$,
the problem becomes the classic orthogonal Procrustes problem,
and the solution is $\mathbf{R} = \hat{\mathbf{S}} \mathbf{S}^\top$,
where $\mathbf{S}$ and $\hat{\mathbf{S}}$
is obtained
from the SVD of $\mathbf{Y}\mathbf{V}^\top$,
$\mathbf{Y}\mathbf{V}^\top = \mathbf{S}\boldsymbol{\Lambda}\hat{\mathbf{S}}^\top$.

\textbf{Comments:}
We present an integrated objective function
that is able to explain
the necessity of PCA dimension reduction.
Let $\bar{\mathbf{y}}$ be a $d$-dimensional vector,
which is a concatenated vector from $\mathbf{y}$ and an all-zero subvector:
$\bar{\mathbf{y}} = [\mathbf{y}^\top 0... 0]^\top$.
The integrated objective function is written as follows:
\begin{align}
\min\|\bar{\mathbf{Y}} - \bar{\mathbf{R}}^\top\mathbf{X}\|_F^2,\label{eqn:IntegratedObjectiveFunctionForITQ}
\end{align}
where $\bar{\mathbf{Y}} = [\bar{\mathbf{y}}_1 \bar{\mathbf{y}}_2 \cdots \bar{\mathbf{y}}_N]$,
$\mathbf{X} = [\mathbf{x}_1 \mathbf{x}_2 \cdots \mathbf{x}_N]$,
and $\bar{\mathbf{R}}$ is a rotation matrix of $d \times d$.
Let $\bar{\mathbf{P}}$ be the projection matrix of $d \times d$,
computed using PCA,
$\bar{\mathbf{P}} = [\mathbf{P} \mathbf{P}_{\perp}]$,
and $\mathbf{P}_{\perp}$ is a matrix of $d \times (d-M)$.
It can be derived that,
the solutions for $\mathbf{y}$ of the two problems
in~\ref{eqn:IntegratedObjectiveFunctionForITQ} and~\ref{eqn:ObjectiveFunctionForITQ}
are the same,
and $\bar{\mathbf{R}} = \bar{\mathbf{P}}\operatorname{diag}(\mathbf{R}, \mathbf{I}_{(d-M) \times (d-M)})$.

%{\color{red}The question: are they equivalent?}
%
%
%{\color{red}connect to scalar quantization, BRE,....}
%
%{\color{green} show it is a special case of Cartesian $k$-means}
%
%{\color{red} Cubic quantization}

\subsubsection{Extensions and Variants}
\emph{Harmonious hashing}~\cite{XuBLCHC13}
modifies iterative quantization
by adding an extra constraint:
$\mathbf{Y}\mathbf{Y}^\top=\sigma \mathbf{I}$.
The problem is solved
by relaxing $\mathbf{Y}$ to continuous values:
fixing $\mathbf{R}$,
let $\mathbf{R}^\top\mathbf{V} = \mathbf{\mathbf{U} \boldsymbol{\Lambda}\mathbf{V}^\top}$,
then $\mathbf{Y} = \sigma^{1/2} \mathbf{U} \mathbf{V}^\top$;
fixing $\mathbf{\mathbf{Y}}$,
$\mathbf{R} = \hat{\mathbf{S}} \mathbf{S}^\top$,
where $\mathbf{S}$ and $\hat{\mathbf{S}}$
is obtained
from the SVD of $\mathbf{Y}\mathbf{V}^\top$,
$\mathbf{Y}\mathbf{V}^\top = \mathbf{S}\boldsymbol{\Lambda}\hat{\mathbf{S}}^\top$.
The hash function is finally computed
as $\mathbf{y} = \operatorname{sgn}(\mathbf{R}^\top\mathbf{v})$.

\emph{Isotropic hashing}~\cite{KongL12a}
finds a rotation following PCA preprocessing
such that
$\mathbf{R}^\top\mathbf{V}\mathbf{V}^\top\mathbf{R} = \boldsymbol{\Sigma}$
becomes a matrix with equal diagonal values,
i.e.,
$[\boldsymbol{\Sigma}]_{11} = [\boldsymbol{\Sigma}]_{22} = \cdots = [\boldsymbol{\Sigma}]_{MM}$.
The objective function is written as
$\|\mathbf{R}^\top\mathbf{V}\mathbf{V}^\top\mathbf{R} - \mathbf{Z}\|_F = 0$,
where $\mathbf{Z}$ is a matrix with all the diagonal entries equal to an unknown variable $\sigma$.
The problem can be solved by two algorithms: lift and projection, and gradient flow.

\textbf{Comments:}
The goal of making the variances along the $M$ directions being the same
is to make the bits in the hash codes equally contributing to the distance evaluation.
In the case that
the data items satisfy the isotropic Gaussian distribution,
the solution from isotropic hashing is equivalent to iterative quantization.

Similar to iterative quantization,
the PCA preprocessing in isotropic hashing is also interpretable:
finding a global rotation matrix $\bar{\mathbf{R}}$ such that
the first $M$ diagonal entries of $\bar{\boldsymbol{\Sigma}} = \bar{\mathbf{R}}^\top\mathbf{X}\mathbf{X}^\top\bar{\mathbf{R}}$
are equal,
and their sum is as large as possible,
which is formally written as follows,
\begin{align}
\max ~&~ \sum\nolimits_{m=1}^M [\bar{\boldsymbol{\Sigma}}]_{mm} \\
\operatorname{s.t.}
~&~ [\bar{\boldsymbol{\Sigma}}]_{mm} = \sigma, m=1,\cdots,M ,~
\bar{\mathbf{R}}^\top\bar{\mathbf{R}} = \mathbf{I}.
\end{align}

Other extensions include
cosine similarity preserving quantization
(\emph{Angular quantization}~\cite{GongKVL12}),
nonlinear embedding replacing PCA embedding~\cite{IrieLWC14}~\cite{ZhaoLM14},
matrix hashing~\cite{GongKRL13},
and so on.
Quantization is also applied to supervised problems:
Supervised discrete hashing~\cite{ShenSLS15, ShenZ0SST16, ZhangZSLC16, ZhangLGZ16},
present an SVM-like formulation
to minimize the quantization loss
and the classification loss in the hash coding space,
and jointly optimize the hash function parameters
and the SVM weights.
Intuitively, the goal of these methods
is that the hash codes are semantically separable,
which is guaranteed through maximizing the classification performance.

%
%{\color{red} why ITQ and IsoH are better than spectral hashing? equal variance? and
%
%is independence really useful?}
%
%{\color{red} Cubic quantization}

\subsection{Cartesian Quantization}
\label{sec:LTH2:CAQ}
Cartesian quantization refers to
a class of quantization algorithms
in which
the composed dictionary $\mathcal{C}$
is formed from a Cartesian product
of a set of small source dictionaries
$\{\mathcal{C}_1, \mathcal{C}_2, \cdots, \mathcal{C}_P\}$:
$\mathcal{C} = \mathcal{C}_1 \times \mathcal{C}_2 \times \cdots \times \mathcal{C}_P
= \{(\mathbf{c}_{1i_1}, \mathbf{c}_{2i_2}, \cdots, \mathbf{c}_{Pi_P})\}$,
where $\mathcal{C}_p = \{\mathbf{c}_{p0}, \mathbf{c}_{p2}, \cdots, \mathbf{c}_{p(K_p-1)}\}$,
$i_p \in \{0, 1, \cdots, K_p -1\}$.

The benefits include that (1) $P$ small dictionaries,
with totally $\sum_{p=1}^PK_p$ dictionary items,
generate a larger dictionary
with $\prod_{p=1}^PK_p$ dictionary items;
(2) the (asymmetric) distance from a query $\mathbf{q}$
to the composed dictionary item
$(\mathbf{c}_{1i_1}, \mathbf{c}_{2i_2}, \cdots, \mathbf{c}_{Pi_P})$
(an approximation of a data item)
is computed from the distances
$\{\operatorname{dist}(\mathbf{q}, \mathbf{c}_{1i_1}), \cdots,
\operatorname{dist}(\mathbf{q}, \mathbf{c}_{Pi_P})\}$
through a sum operation,
thus the cost of the distance computation
between a query and a data item
is $O(P)$,
if the distances between the query and the source dictionary items
are precomputed;
and
(3) the query cost with a set of $N$ database items
is reduced from
$Nd$ to $NP$
through looking up a distance table
which is efficiently computed
between the query and the $P$ source dictionaries.

\subsubsection{Product Quantization}
Product quantization~\cite{JegouDS11},
which initiates the quantization-based compact coding solution to similarity search,
forms the $P$ source dictionaries
by dividing the feature space
into $P$ disjoint subspaces,
accordingly dividing the database into $P$ sets,
each set consisting of $N$ subvectors
$\{\mathbf{x}_{p1}, \cdots, \mathbf{x}_{pN}\}$,
and then quantizing each subspace separately
into (usually $K_1 = K_2 = \cdots = K_P = K$) clusters.
Let $\{\mathbf{c}_{p1}, \mathbf{c}_{p2},
\cdots, \mathbf{c}_{pK}\}$ be the cluster centers
of the $p$th subspace.
The operation forming an item in the dictionary
from a P-tuple $(\mathbf{c}_{1i_1}, \mathbf{c}_{2i_2}, \cdots, \mathbf{c}_{Pi_P})$
is the concatenation $[\mathbf{c}_{1i_1}^\top \mathbf{c}_{2i_2}^\top \cdots\mathbf{c}_{Pi_P}^\top]^\top$.
A data point assigned to
the nearest dictionary item $(\mathbf{c}_{1i_1}, \mathbf{c}_{2i_2}, \cdots, \mathbf{c}_{Pi_P})$
is represented
by a compact code $(i_1,i_2, \cdots, i_P)$,
whose length is $P\log_2 K$.
The distance $\operatorname{dist}(\mathbf{q}, \mathbf{c}_{pi_p})$
between a query $\mathbf{q}$ and the dictionary element in the $p$th dictionary
is computed as $\|\mathbf{q}_p - \mathbf{c}_{pi_p}\|_2^2$,
where $\mathbf{q}_p$ is the subvector of $\mathbf{q}$
in the $p$th subspace.

Mathematically,
product quantization can be viewed
as minimizing the following objective function,
\begin{align}
\min\nolimits_{\mathbf{C}, \{\mathbf{b}_n\}} ~&~
\sum\nolimits_{n=1}^N \|\mathbf{x}_n - \mathbf{C}\mathbf{b}_n\|_2^2.
\end{align}
Here $\mathbf{C}$ is a matrix of $d \times PK$
in the form of
\begin{align}
\mathbf{C} = \operatorname{diag}(\mathbf{C}_1, \mathbf{C}_2, \cdots, \mathbf{C}_P)
=
\begin{bmatrix}
\mathbf{C}_1 & \mathbf{0} &  \cdots &  \mathbf{0}  \\[0.3em]
\mathbf{0} &  \mathbf{C}_2 &  \cdots &  \mathbf{0}  \\[0.3em]
\vdots  & \vdots  & \ddots & \vdots   \\[0.3em]
\mathbf{0} &  \mathbf{0} &  \cdots &  \mathbf{C}_P   \\
\end{bmatrix}, \nonumber
\end{align}
where $\mathbf{C}_p = [\mathbf{c}_{p1} \mathbf{c}_{p2}\cdots \mathbf{c}_{pK}]$.
$\mathbf{b}_n = [\mathbf{b}_{n1}^\top \mathbf{b}_{n2}^\top \cdots \mathbf{b}_{nP}^\top]^\top$ is the composition vector,
and its subvector
$\mathbf{b}_{np}$
of length $K$
is an indicator vector
with only one entry being $1$
and all others being $0$,
showing which element is selected
from the $p$th source dictionary for quantization.

\textbf{Extensions:}
\emph{Distance-encoded product quantization}~\cite{HeoLY14}
extends product quantization
by encoding both the cluster index
and the distance between the cluster center and the point.
The cluster index is encoded in a way similar to that in product quantization.
The way
of encoding the distance between a point and its cluster center
is as follows:
the points belonging to one cluster
are partitioned (quantized)
according to the distances to the cluster center,
the points in each partition are represented
by the corresponding partition index,
and accordingly the distance of each partition to the cluster center is also recorded with the partition index.

\emph{Cartesian $k$-means}~\cite{NorouziF13}
and \emph{optimized production quantization}~\cite{GeHK013} extend product quantization
and introduce a rotation $\mathbf{R}$
into the objective function,
\begin{align}
\min\nolimits_{\mathbf{R}, \mathbf{C}, \{\mathbf{b}_n\}} ~&~
\sum\nolimits_{n=1}^N \|\mathbf{R}^\top\mathbf{x}_n - \mathbf{C}\mathbf{b}_n\|_2^2.
\end{align}
The introduced rotation does not
affect the Euclidean distance
as the Euclidean distance is invariant to the rotation,
and helps to find an optimized subspace partition
for quantization.
\emph{Locally optimized product quantization}~\cite{KalantidisA14}
applies optimized production quantization to the search algorithm
with the inverted index,
where there is a quantizer for each inverted list.

\subsubsection{Composite Quantization}
In composite quantization~\cite{ZhangDW14},
the operation forming an item in the dictionary
from a P-tuple $(\mathbf{c}_{1i_1}, \mathbf{c}_{2i_2}, \cdots, \mathbf{c}_{Pi_P})$
is the summation $\sum_{p=1}^P\mathbf{c}_{pi_p}$.
In order to
compute the distance from a query $\mathbf{q}$
to the composed dictionary item formed
by $(\mathbf{c}_{1i_1}, \mathbf{c}_{2i_2}, \cdots, \mathbf{c}_{Pi_P})$
from the distances
$\{\operatorname{dist}(\mathbf{q}, \mathbf{c}_{1i_1}), \cdots,
\operatorname{dist}(\mathbf{q}, \mathbf{c}_{1i_1})\}$,
a constraint is introduced:
the summation of the inner products of all
pairs of elements that are used to approximate the vector $\mathbf{x}_n$
but from different dictionaries,
$\sum_{i=1}^P\sum_{j=1, \neq i}^P \mathbf{c}_{ik_{in}} \mathbf{c}_{jk_{jn}}$,
is constant.

The problem is formulated as
\begin{align}\small
\min_{\{\mathbf{C}_p\},\{\mathbf{b}_n\}, \epsilon } ~&~\sum\nolimits_{n=1}^N\|\mathbf{x}_n -  [\mathbf{C}_1 \mathbf{C}_2 \cdots \mathbf{C}_P]\mathbf{b}_{n}\|_2^2 \label{eqn:originalobjectivefunction} \\
\operatorname{s.t.}
~&~ \sum\nolimits_{j=1}^P \sum\nolimits_{i=1, i\neq j}^P  \mathbf{b}_{ni}^\top \mathbf{C}_{i}^\top\mathbf{C}_{j}\mathbf{b}_{nj} = \epsilon ,\nonumber \\
~&~ \mathbf{b}_n  = [\mathbf{b}_{n1}^\top\mathbf{b}_{n2}^\top\cdots\mathbf{b}_{nP}^\top]^\top, \nonumber \\
~&~ \mathbf{b}_{np} \in \{0, 1\}^K,~\|\mathbf{b}_{np}\|_1 = 1 ,\nonumber \\
~&~ n=1,2,\cdots,N; p=1,2, \cdots P.\nonumber
\end{align}
Here, ${\mathbf{C}_p}$ is a matrix of size $d \times K$,
and each column corresponds to an element
of the $p$th dictionary $\mathcal{C}_p$.

Sparse composite quantization~\cite{ZhangQTW15} improves composite quantization
by constructing a sparse dictionary,
$\sum_{p=1}^P\sum_{k=1}^K \|\mathbf{c}_{pk}\|_0 \leqslant S$,
with $S$ being a parameter controlling the sparsity degree,
resulting in a great reduction of the distance table computation cost
which takes almost the same as the most efficient approach: product quantization.

\begin{figure}
\centering
%\small{(a)}~{\includegraphics[width=.3\linewidth, clip]{figs/PQ}}~~
%\small{(b)}~{\includegraphics[width=.3\linewidth, clip]{figs/CKM}}~~
%\small{(c)}~{\includegraphics[width=.3\linewidth, clip]{figs/CQ}}
\subfloat[(a)][]{\includegraphics[width=.3\linewidth, clip]{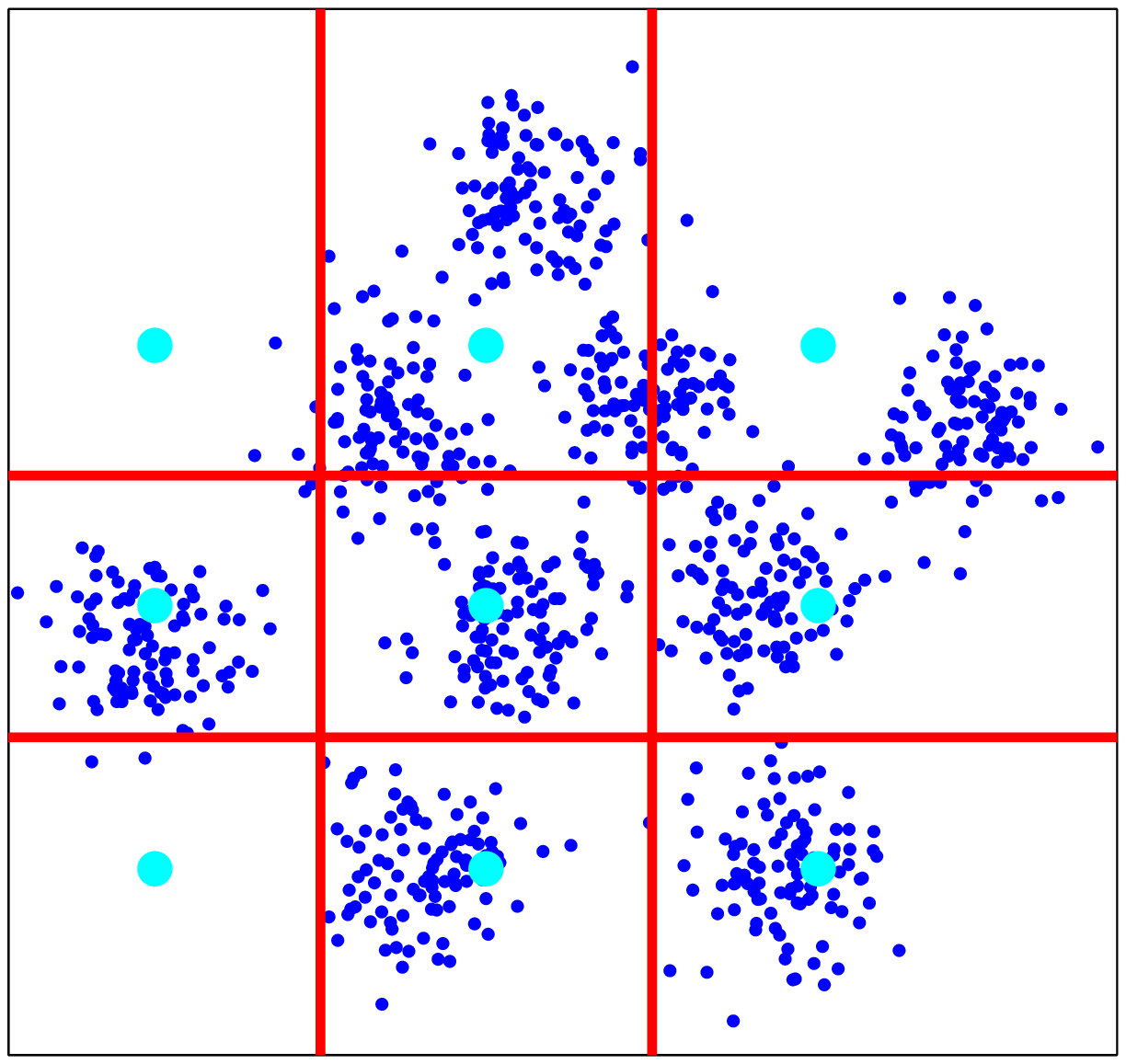}}~~~~
\subfloat[(b)][]{\includegraphics[width=.3\linewidth, clip]{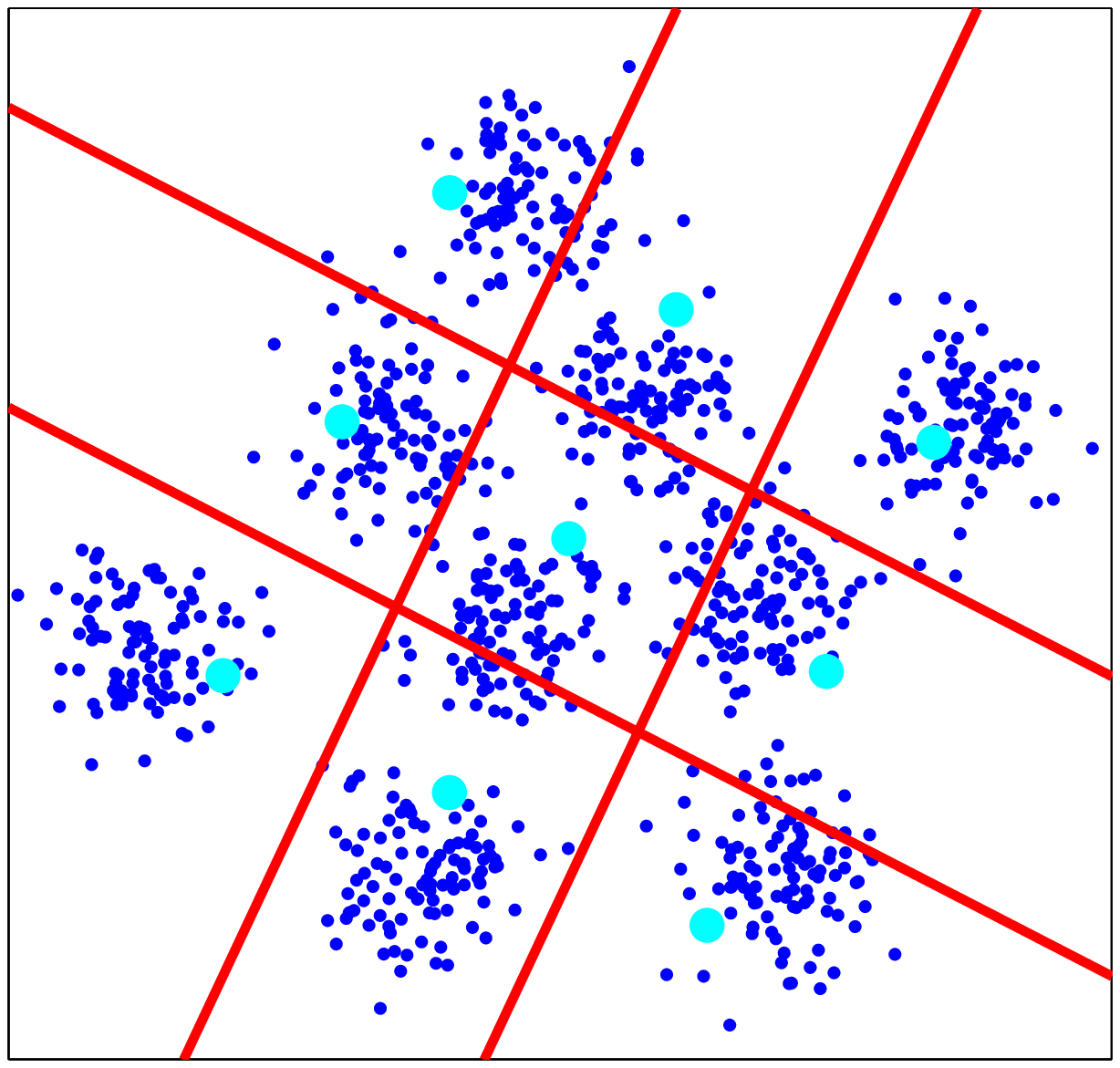}}~~~~
\subfloat[(c)][]{\includegraphics[width=.3\linewidth, clip]{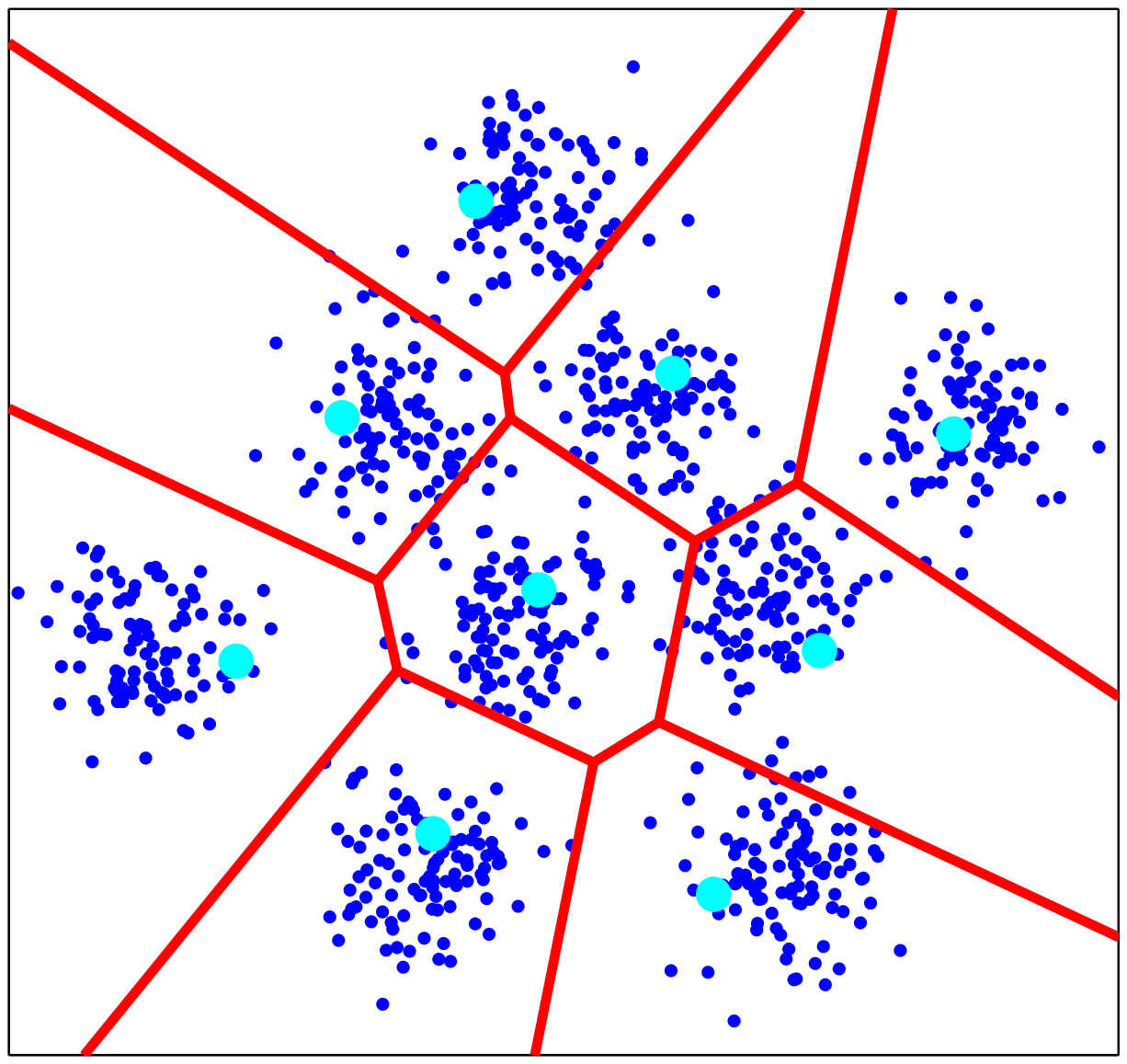}}
\caption{$2$D toy examples illustrating the quantization algorithms.
The space partitioning results are
generated by (a) product quantization,
(b) Cartesian $k$-means,
and (c) composite quantization.
The space partition from composition quantization is more flexible. }
\label{fig:quantization}
\vspace{-.5cm}
\end{figure}

\textbf{Connection with product quantization:}
It is shown in~\cite{ZhangDW14}
that both product quantization and Cartesian $k$-means
can be regarded as constrained versions of composite quantization.
Composite quantization attains smaller quantization errors,
yielding better search accuracy
with similar search efficiency.
A $2$D illustration of the three algorithms is given
in Figure~\ref{fig:quantization},
where $2$D points are grouped
into $9$ groups.
It is observed that composition quantization
is more flexible in partitioning the space
and thus the quantization error is possibly smaller.

Composite quantization, product quantization, and Cartesian $k$-means (optimized product quantization)
can be explained from the view of sparse coding,
as pointed in~\cite{ZhangDW14, BabenkoK14,VedaldiZisserman12}:
the dictionary ($\{\mathbf{C}_p\}$) in
composite quantization (product quantization and Cartesian $k$-means)
satisfies the constant (orthogonality) constraint,
and the sparse codes ($\{\mathbf{b}_n\}$)
are $0$ and $1$ vectors
where there is only one $1$ for each subvector corresponding to a source dictionary.

\textbf{Comments:}
As discussed in product quantization~\cite{JegouDS11},
the idea of using the summation
of several dictionary items
as an approximation of a data item
has already been studied
in the signal processing research area,
known as multi-stage vector quantization,
residual quantization,
or more generally structured vector quantization~\cite{GrayN98},
and recently re-developed
for similarity search under the Euclidean distance
(additive quantization~\cite{BabenkoK14},~\cite{WangWSXSL14},
and tree quantization~\cite{BabenkoK15}
modifying additive quantization
by introducing a tree-structure sparsity)
and inner product~\cite{DuW14}.

\subsubsection{Variants}
The work in~\cite{GordoPGL14}
presents an approach
to compute the source dictionaries
given the $M$ hash functions $\{h_m(\mathbf{x}) = b_m(g_m(\mathbf{x}))\}$,
where $g_m()$ is a real-valued embedding function
and $b_m()$ is a binarization function,
for a better distance measure,
quantization-like distance,
instead of Hamming or weighted Hamming distance.
It computes $M$ dictionaries,
each corresponding to a hash bit
and computed as
\begin{align}
\bar{g}_{kb} = \operatorname{E}(g_k(\mathbf{x})~|~  b_k(g_k(\mathbf{x})) = b),
\end{align}
where $b=0$ and $b=1$.
The distance computation cost is $O(M)$
through looking up a distance table,
which can be accelerated
by
dividing the hash functions into groups
(e.g., each group contains $8$ functions,
and thus the cost is reduced to $O(\frac{M}{8})$),
building a table (e.g., consisting of $256$ entries)
per group instead of per hash function,
and forming a larger distance lookup table.
In contrast,~\emph{optimized code ranking}~\cite{WangSYYLW14}
directly estimates the distance table
rather than computing it from the estimated dictionary.

Composite quantization~\cite{ZhangDW14}
points to relation between Cartesian quantization
and sparse coding.
This indicates the application of sparse coding to similarity search.
\emph{Compact sparse coding}~\cite{Cherian14},
the extension of robust sparse coding~\cite{CherianMP12},
adopts sparse codes to represent
the database items:
the atom indices corresponding to nonzero codes,
which is equivalent to letting the hash bits associated with nonzero codes
be $1$ and $0$ for zero codes,
are
used to build the inverted index,
and the nonzero coefficients
are used to
reconstruct the database items
and calculate the distances
between the database items and
the query.
\emph{Anti-sparse coding}~\cite{JegouFF12}
aims to learn a hash code
so that non-zero elements in the hash code are as many as possible.

\subsection{Reconstruction}
We review a few reconstruction-based hashing approaches.
Essentially, quantization can be viewed as a reconstruction approach
for a data item.
\emph{Semantic hashing}~\cite{SalakhutdinovH07},~\cite{SalakhutdinovH09}
generates the hash codes using the deep generative model,
a restricted Boltzmann machine (RBM),
for reconstructing the data item.
As a result, the binary codes are used for finding similar data.
A variant method proposed in~\cite{Carreira-Perpinan15}
reconstructs the input vector from the binary codes,
which is effectively solved using the auxiliary coordinates algorithm.
A simplified algorithm~\cite{BaluFJ14}
finds a binary hash code
that can be used to effectively reconstruct the vector
through a linear transformation.

\section{Other Topics}
\label{sec:others}
Most hashing learning algorithms
assume that the similarity information in the input space,
especially the semantic similarity information,
and the database items
have already been given.
There are some approaches to
learn hash functions without such assumptions:
\emph{active hashing}~\cite{ZhenY13}
that actively selects the labeled pairs which are most informative
for hash function learning,
\emph{online hashing}~\cite{HuangYZ13},
\emph{smart hashing}~\cite{YangHZL13},
\emph{online sketching hashing}~\cite{LengWC0L15},
and~\emph{online adaptive hashing}~\cite{CakirS15},
which learn the hash functions
when the similar/dissimilar pairs come
sequentially.

The manifold structure in the database
is exploited
for hashing,
which is helpful
for semantic similarity search,
such as
\emph{locally linear hashing}~\cite{IrieLWC14},
\emph{spline regression hashing}~\cite{LiuWYZH12},
and~\emph{inductive manifold hashing}~\cite{ShenSSHT13}.
Multi-table hashing,
aimed at improving locality sensitive hashing,
is also studied,
such as
\emph{complementary hashing}~\cite{XuWLZLY11}
and its multi-view extension~\cite{LiuHDLL15},
\emph{reciprocal hash tables}~\cite{LiuHL13}
and its query-adaptive extension~\cite{LiuDLTL16},
and so on.

There are some works extending the Hamming distance.
In contrast to multi-dimensional spectral hashing~\cite{WeissFT12}
in which the weights for the weighted Hamming distance
are the same for arbitrary queries,
the query-dependent distance approaches
learn a distance measure
whose weights or parameters depend on
a specific query.
\emph{Query adaptive hashing}~\cite{LiuYJHZ13},
a learning-to-hash version extended
from query adaptive locality sensitive hashing~\cite{JegouASG08},
aims to select the hash bits (thus hash functions forming the hash bits)
according to the query vector.
\emph{Query-adaptive class-specific bit weighting}
\cite{JiangWC11},~\cite{JiangWXC13} presents a weighted Hamming distance measure
by learning the class-specific bit weights
from the class information of the query.
\emph{Bits reconfiguration}~\cite{MuCLCY12}
is to learn a good distance measure
over the hash codes precomputed
from a pool of hash functions.

The following reviews three research topics:
joint feature and hash learning with deep learning,
fast search in the Hamming space
replacing the exhaustive search,
and the important application of the Cartesian quantization
to inverted index.

\subsection{Joint Feature and Hash Learning via Deep Learning}
%Recent deep learning developments also indicate
%an emerging topic,
%learning an end-to-end hashing system
%without a separate intermediary feature extraction step.
The great success in deep neural network for representation learning
has inspired a lot of deep compact coding algorithms~\cite{XiaPLLY14,LaiPLY15,GaoSZZS15, ZhaoHWT15}.
Typically,
these approaches except~\cite{LaiPLY15}
simultaneously learn the representation
using a deep neural network
and the hashing function under some loss functions,
rather than separately learn the features and then learn the hash functions.

The methodology is similar to other learning to hash algorithms
that do not adopt deep learning,
and the hash function is more general and could be a deep neural network.
We provide here a separate discussion because this area is relatively new.
However, we will not discuss semantic hashing~\cite{SalakhutdinovH07}
which is usually not thought as a feature learning approach
but just a hash function learning approach.
In general, almost all non-deep-learning hashing algorithms
if the similarity order (e.g., semantic similarity) is given,
can be extended to deep learning based hashing algorithms.
In the following,
we discuss the deep learning based algorithms
and also categorize them according to their similarity preserving manners.
\begin{itemize}
  \item Pairwise similarity preserving.
  The similarity-similarity difference minimization criterion
  is adopted in~\cite{XiaPLLY14}.
  It uses a two-step scheme:
  the hash codes are computed
  by minimizing the similarity-similarity difference
  without considering the visual information,
  and then the image representation and hash function
  are jointly learnt through deep learning.

  \item Multiwise similarity preserving. The triplet loss is used in~\cite{ZhaoHWT15,LaiPLY15},
which adopt the loss function defined in Equation~(\ref{eqn:tripletlosshashing})
($1$ is dropped in~\cite{LaiPLY15})
  \item Quantization. Following the scalar quantization approach,
deep hashing~\cite{LiongLWMZ15}
defines a loss to penalize the difference between the binary hash codes (see Equation~(\ref{eqn:scalarquantizationloss}))
and the real values from which a linear projection is used to generate the binary codes,
and introduces the bit balance and bit uncorrelation conditions.
\end{itemize}

\subsection{Fast Search in the Hamming Space}
The computation of the Hamming distance
is shown much faster
than the computation of the distance in the input space.
It is still expensive, however,
to handle a large scale data set
using linear scan.
Thus, some indexing algorithms
already shown effective and efficient for
general vectors
are borrowed for the search in the Hamming space.
For example,
min-hash, a kind of LSH,
is exploited to search over
high-dimensional binary data~\cite{ShrivastavaL12}.
In the following,
we discuss other representative algorithms.

\emph{Multi-index hashing}~\cite{NorouziPF12}
and its extension~\cite{SongSWHSW16} aim to
partition the binary codes into $M$ disjoint substrings
and build $M$ hash tables each corresponding to a substring,
indexing all the binary codes $M$ times.
Given a query,
the method outputs the NN candidates which are near to the query
at least in one hash table.
\emph{FLANN}~\cite{MujaL12} extends the FLANN algorithm~\cite{MujaL09}
that was initially designed for ANN search
over real-value vectors
to search over binary vectors.
The key idea is to build multiple hierarchical cluster trees to organize the binary vectors
and to search for the nearest neighbors
simultaneously over the multiple trees
by traversing each tree in a best-first manner.

PQTable~\cite{MatsuiYA15}
extends multi-index hashing
from the Hamming space to the product-quantization coding space,
for fast exact search.
Unlike multi-index hashing
flipping the bits in the binary codes
to find candidate tables,
PQTable adopts the multi-sequence algorithm~\cite{BabenkoL12}
for efficiently finding candidate tables.
The neighborhood graph-based search algorithm~\cite{WangL12}
for real-value vectors
is extended to the Hamming space~\cite{JiangXDXW16}.

\subsection{Inverted Multi-Index}
Hash table lookup with binary hash codes is
a form of inverted index.
Retrieving multiple hash buckets
for multiple hash tables
is computationally cheaper
compared with the subsequent reranking step
using the true distance computed in the input space.
It is also cheap
to visit more buckets in a single table
if the standard Hamming distance is used,
as the nearby hash codes of the hash code of the query
which can be obtained
by flipping the bits of the hash code of the query.
If there are a lot of empty buckets
which increases the retrieval cost,
the double-hash scheme or the fast search algorithm
in the Hamming space,
e.g.,~\cite{NorouziPF12, MujaL12}
can be used to fast retrieve the hash buckets.

Thanks to the multi-sequence algorithm,
the Cartesian quantization algorithms
are also applied to the inverted index~\cite{BabenkoL12},~\cite{ZhangQTW15},~\cite{GeHK013}
(called inverted multi-index),
in which each composed quantization center
corresponds to an inverted list.
Instead of comparing the query with all the composed quantization centers,
which is computationally expensive,
the multi-sequence algorithm~\cite{BabenkoL12} is able
to efficiently produce a sequence of ($T$) inverted lists
ordered by the increasing distances between the query
and the composed quantization centers,
whose cost is $O(T\log T)$.
The study (Figure 5 in~\cite{WangWZGLG13})
shows that the time cost of the multi-sequence algorithm
when retrieving $10K$ candidates over the two datasets: SIFT$1M$ and GIST$1M$
is the smallest compared with other non-hashing inverted index algorithms.

Though the cost of the multi-sequence algorithm is greater
than that with binary hash codes,
both are relatively small and negligible
compared with the subsequent reranking step
that is often conducted in real applications.
Thus the quantization-based inverted index (hash table)
is more widely used
compared with
the conventional hash tables with binary hash codes.

%\begin{table}[!t]
%% increase table row spacing, adjust to taste
%% \renewcommand{\arraystretch}{1.3}
%\caption{A summary of evaluation metrics}
%\label{table:evaluationmetrics}
%\centering
%% Some packages, such as MDW tools, offer better commands for making tables
%% than the plain LaTeX2e tabular which is used here.
%\begin{tabular}{|c|c||c|c|}
%\hline
%\multicolumn{2}{|c||}{Euclidean} & \multicolumn{2}{c|}{Semantic} \\
%\hline
% recall@$R$ & recall vs. query time & PR curve & mAP \\
%\hline
%\end{tabular}
%\end{table}

\section{Evaluation Protocols}
\label{sec:evaluationprotocols}
\subsection{Evaluation Metrics}
There are three main concerns
for an approximate nearest neighbor search algorithm:
space cost,
search efficiency,
and search quality.
The space cost for hashing algorithms
depends on the code length for hash code ranking,
and the code length and the table number for hash table lookup.
The search performance is usually measured
under the same space cost,
i.e.,
the code length (and the table number)
is chosen the same for different algorithms.

The search efficiency
is measured
as the time taken to
return the search result for a query,
which is usually computed as the average time
over a number of queries.
The time cost often does not include the cost
of the reranking step
(using the original feature representations)
as it is assumed that
such a cost given the same number of candidates
does not depend on the hashing algorithms
and can be viewed as a constant.
When comparing the performance
in the case the Hamming distance in hash code ranking
is used in the coding space,
it is not necessary
to report the search time costs
because they are the same.
It is necessary to report the search time cost
when a non-Hamming distance or the hash table lookup scheme
is used.

The search quality is measured using recall@$R$ (i.e., a recall-$R$ curve).
For each
query, we retrieve its $R$ nearest items
and compute the ratio of the true nearest items in
the retrieved $R$ items to $T$ ,
i.e., the fraction of $T$ ground-truth nearest
neighbors are found in the retrieved $R$ items.
The average
recall score over all the queries is used as the measure.
The ground-truth nearest neighbors are computed over the
original features using linear scan.
Note that the recall@$R$ is equivalent to the accuracy computed
after reordering the $R$ retrieved nearest items
using the original features
and returning the top $T$ items.
In the case where the linear scan cost in the hash coding space
is not the same
(e.g., binary code hashing, and quantization-based hashing),
the curve in terms of
search recall and search time cost is usually reported.

The semantic similarity search,
a variant of nearest neighbor search,
sometimes uses the precision,
the recall,
the precision-recall curve,
and
mean average precision (mAP).
The precision is computed
at the retrieved position $R$,
i.e., $R$ items are retrieved,
as the ratio of the number of retrieved true positive items to $R$.
The recall is computed,
also at position $R$,
as the ratio of the number of retrieved true positive items
to the number of all true positive items in the database.
The pairs of recall
and precision
in the precision-recall curve
are computed by varying the retrieved position $R$.
The mAP score is computed as follows:
the average precision for a query,
the area under the precision-recall curve is computed as
$\sum_{t=1}^N P(t)\Delta(t)$,
where $P(t)$ is the precision at cut-off $t$ in the ranked list
and $\Delta(t)$ is the change
in recall from items $t-1$ to $t$;
the mean of average precisions over all the queries
is computed as the final score.

\begin{table}[!t]
% \renewcommand{\arraystretch}{1.3}
\caption{A summary of evaluation datasets}
\vspace{-.3cm}
\label{table:evaluationdatasets}
\centering
% Some packages, such as MDW tools, offer better commands for making tables
% than the plain LaTeX2e tabular which is used here.
\begin{tabular}{|c||c||c|c|c|}
\hline
 & Dim & Reference set & Learning set & Query set \\
\hline
MNIST & $784$ & $60K$ & - & $10K$\\
SIFT$10K$ & $128$ & $10K$ & $25K$ & $100$  \\
SIFT$1M$ & $128$ & $1M$ & $100K$ & $10K$  \\
GIST$1M$ & $960$ & $1M$ & $50K$ & $1K$  \\
Tiny$1M$ & $384$ & $1M$ & - & $100K$  \\
SIFT$1B$ & $128$ & $1B$ & $100M$/$1M$ & $10K$  \\
GloVe$1.2M$ & $200$ & $\approx1.2M$ & - & $10K$  \\
\hline
\end{tabular}
\vspace{-.3cm}
\end{table}
\subsection{Evaluation Datasets}
The widely-used evaluation datasets
have different scales
from small, large,
to very large.
Various features have been used,
such as SIFT features~\cite{Lowe04}
extracted from Photo-tourism~\cite{SnavelySS06}
and Caltech $101$~\cite{FeiFP04},
GIST features~\cite{OlivaT01}
from LabelMe~\cite{RussellTMF08} and Peekaboom~\cite{AhnLB06},
as well as some features used in object retrieval:
Fisher vectors~\cite{PerronninLSP10} and VLAD vectors~\cite{JegouDSP10}.
The following presents a brief introduction
to several representative datasets,
which is summarized in Table~\ref{table:evaluationdatasets}.

MNIST~\cite{LeCunBBH01} includes $60K$ $784$-dimensional raw pixel features
describing grayscale images of handwritten digits
as a reference set,
and $10K$ features as the queries.

SIFT$10K$~\cite{JegouDS11}
consists of
$10K$ $128$-dimensional SIFT vectors as the reference set,
$25K$ vectors as the learning set,
and $100$ vectors as the query set.
SIFT$1M$~\cite{JegouDS11} is composed of $1M$
$128$-dimensional SIFT vectors
as the reference set,
$100K$ vectors as the learning set,
and $10K$ as the query set.
The learning sets in SIFT$10K$ and SIFT$1M$
are extracted from Flicker images
and the reference sets and the query sets
are from the INRIA holidays images~\cite{JegouDS08}.

GIST$1M$~\cite{JegouDS11}
consists of $1M$ $960$-dimensional GIST vectors as the reference set,
$50K$ vectors as the learning set,
$1K$ vectors as the query set.
The learning set is extracted from the first $100K$ images
from the tiny images~\cite{TorralbaFF08}.
The reference set is from the Holiday images
combined with Flickr$1M$~\cite{JegouDS08}.
The query set is from the Holiday image queries.
Tiny$1M$~\cite{WangWJLZZH13}\footnote{\url{http://research.microsoft.com/~jingdw/SimilarImageSearch/NNData/NNdatasets.html}}
consists of $1M$ $384$-dimensional GIST vectors as the reference set
and $100K$ vectors as the query set.
The two sets are extracted from the $1100K$ tiny images.

SIFT$1B$~\cite{JegouTDA11}
includes $1B$ $128$-dimensional BYTE-valued SIFT vectors as the reference set,
$100M$ vectors as the learning set
and $10K$ vectors as the query set.
The three sets are extracted from around $1M$ images.
This dataset, and SIFT$10K$, SIFT$1M$ and GIST$1M$
are publicly available\footnote{\url{http://corpus-texmex.irisa.fr/}}.

GloVe$1.2M$~\cite{pennington2014glove}\footnote{\url{http://nlp.stanford.edu/projects/glove/}}
contains $1,193,514$ $200$-dimensional word feature vectors extracted from Tweets.
We randomly sample $10K$ vectors as the query set
and use the remaining as the training set.

\subsection{Training Sets and Hyper-Parameters Selection}
There are three main choices of the training set
over which the hash functions are learnt
for learning-to-hash algorithms.
The first choice is a separate set
used for learning hash functions,
which is not contained in the reference set.
The second choice is to sample a small subset
from the reference set.
The third choice is to use all the reference set
to train hash functions.
The query set and the reference set
are then used to evaluate the learnt hash functions.

In the case where the query is transformed to a hash code,
e.g., when adopting the Hamming distance for most binary hash algorithms,
learning over the whole reference set might lead to over-fitting
and the performance might be worse than learning
with a subset of the reference set
or a separate set.
In the case where the raw query is used without any processing,
e.g., when adopting the asymmetric distance in Cartesian quantization,
learning over the whole reference set is better
as it results in better approximation
of the reference set.

There are some hyper-parameters in the objective functions,
e.g, the objective functions in minimal loss hashing~\cite{NorouziF11} and composite quantization~\cite{ZhangDW14}.
It is unfair and not suggested
to select the hyper-parameters
corresponding to the best performance over the query set.
It is suggested instead
to select the hyper-parameters by validation,
e.g., sampling a subset from the reference set as the validation set
which is reasonable
because the validation criterion is not the objective
function value but the search performance.

\begin{table}[!t]
% \renewcommand{\arraystretch}{1.3}
\caption{A summary of query performance comparison
for approximate nearest neighbor search
under Euclidean distance.}
\vspace{-.3cm}
\label{table:queryperformance}
\centering
% Some packages, such as MDW tools, offer better commands for making tables
% than the plain LaTeX2e tabular which is used here.
\begin{tabular}{|c||c|c|c|}
\hline
 & Accuracy & Efficiency & Overall  \\
\hline
pairwise & low & high & low \\
multiwise & fair & high & fair \\
quantization & high & fair & high \\
\hline
\end{tabular}
\vspace{-.3cm}
\end{table}

\section{Performance Analysis}
\label{sec:discussion}
\subsection{Query Performance}
We summarize empirical observations and the analysis
of the nearest neighbor search performance
using the compact coding approach,
most of which have already been mentioned
or discussed in the existing works.
We discuss about both hash table lookup
and hash code ranking,
with more focus on hash code ranking
because the major usage of
the learning to hash algorithms
lies in hash code ranking
for retrieving top candidates
from a set of candidates
obtained from the inverted index
or other hash table lookup algorithms.
The analysis is mainly focusing on
the major application of hashing:
nearest neighbor search with the Euclidean distance.
The conclusion for semantic similarity search
is similar in principle
and the performance also depends on the ability
of representing the semantic meaning
of the input features.
We also present empirical results
of the quantization algorithms
and the representative binary coding algorithms
for hash code ranking.

\subsubsection{Query Performance with Hash Table Lookup}
We give a performance summary
of the query scheme using hash table lookup
for the two main hash algorithms:
the binary hash codes
and the quantization-based hash codes.

In terms of space cost,
hash table lookup with binary hash codes
has a little but negligible advantage
over that with quantization-based hash codes
because the main space cost comes from the indices of the reference items
and the extra cost from the centers
corresponding to the buckets
using quantization
is relatively small.
Multi-assignment and multiple hash tables
increase space cost
as they require to store multiple copies of reference vector indices.
As an alternative choice,
single-assignment with a single table can be used
but more buckets are retrieved
for high recall.

When retrieving the same number of candidates,
hash table lookup using binary hash codes
is better in terms of the query time cost,
but inferior to the quantization approach
in terms of the recall,
which has probably been firstly discussed in~\cite{PauleveJA10}.
In terms of recall vs. time cost
the quantization approach is overall superior
as the cost from the multi-sequence algorithm is relatively small and negligible
compared with the subsequent reranking step,
which is observed from our experience,
and can be derived from~\cite{MujaL09} and~\cite{BabenkoL12}.
In general,
the performance for other algorithms based on the weighted Hamming distance and the learnt distance
is in between.
The observation holds
for a single table
with single assignment and
multiple assignment,
or multiple tables.

\begin{figure}[t]
\centering
\small{(a)}~{\includegraphics[width=.4\linewidth, clip]{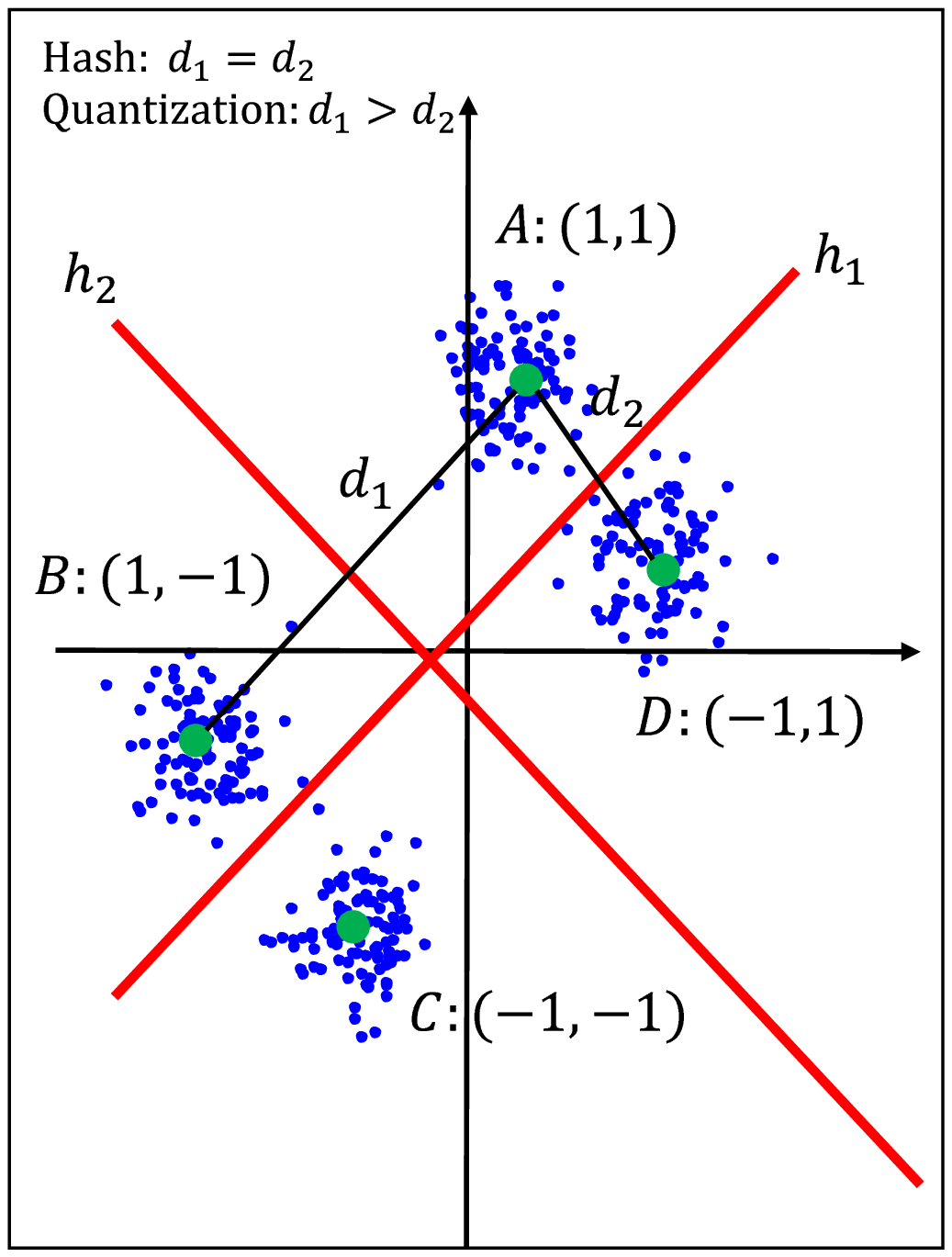}}~~~~
\small{(b)}~{\includegraphics[width=.4\linewidth, clip]{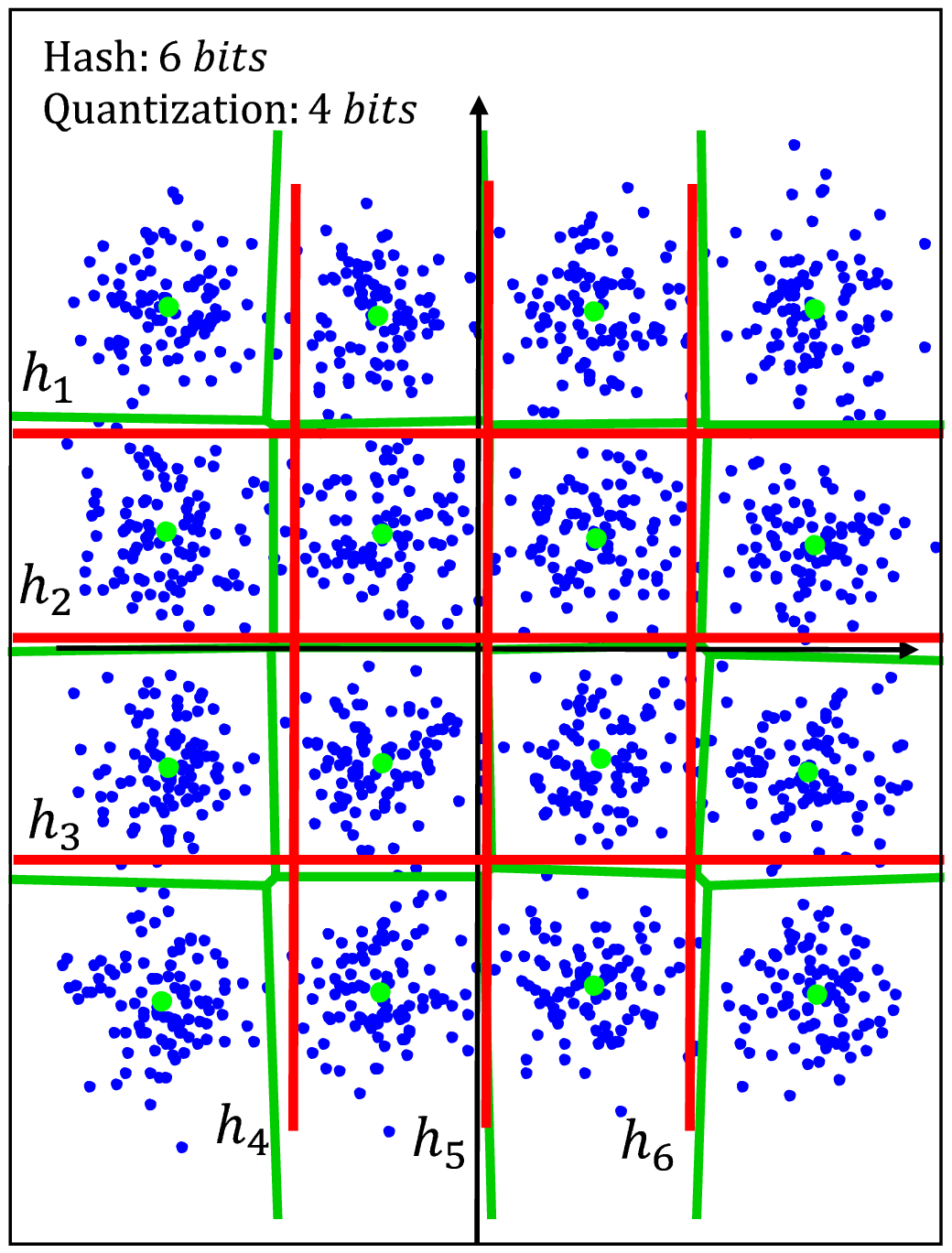}}
\caption{$2$D toy examples illustrating the comparison
between binary code hashing and quantization.
(a) shows the Hamming distances from clusters $B$ and $D$ to cluster $A$,
usually adopted in the binary code hashing algorithms,
are the same
while the Euclidean distances,
used in the quantization algorithms,
are different.
(b) the binary code hashing algorithms need $6$ hash bits (red lines show the corresponding hash functions)
to
differentiate the $16$ uniformly-distributed clusters
while the quantization algorithms only require $4$ ($= \log 16 $) bits
(green lines show the partition line).
}
\label{fig:QuantizationOverHash}
\vspace{-.5cm}
\end{figure}

\subsubsection{Query Performance with Hash Code Ranking}
\label{sec:discussion:queryperformance:HCR}
The following provides a short summary
of the overall performance
for three main categories:
pairwise similarity preserving,
multiwise similarity preserving,
and quantization
in terms of search cost and search accuracy
under the same space cost,
guaranteed by coding the items
using the same number of bits,
ignoring the small space cost of the dictionary
in Cartesian quantization
and the distance lookup tables.

\textbf{Search accuracy:}
Multiwise similarity preserving is better
than pairwise similarity preserving
as it considers more information
for hash function learning.
There is no observation/conclusion
on
which algorithm, pairwise or multiwise similarity preserving algorithm,
performs consistently the best.
Nevertheless, there is a large amount of pairwise and multiwise similarity preserving algorithms
because different algorithms may be suitable to different data distributions
and optimization also affects the performance.

It has been shown in Section~\ref{sec:LTH2:HD}
that
the cost function of hypercubic quantization
is an approximation of the distance-distance difference.
But it outperforms pairwise and multiwise
similarity preserving algorithms.
This is because it is infeasible to
consider all pairs (triples) of items for the distance-distance difference
in pairwise (multiwise) similarity preserving algorithms,
and thus only a small subset of the pairs (triples),
by sampling a subset of items or pairs (triples),
is considered
for almost all the pairwise (multiwise) similarity preserving hashing algorithms,
while the cost function for quantization
is an approximation for all pairs of items.
This point is also discussed in~\cite{WangZQTW16}.
%In general, most binary code hashing algorithms
%can benefit from the kernel hash function,
%and weighted Hamming distances as well as learnt distances
%for binary codes,
%which increase the search accuracy
%and also the search cost.

Compared with binary code hashing including
hypercubic quantization,
another reason for the superiority of Cartesian quantization,
as discussed in~\cite{ZhangDW14},
is that there are only a small number ($L+1$) of distinct Hamming distances
in the coding space for binary code hashing
with the code length being $L$,
while the number of distinct distances for Cartesian quantization
is much larger.
It is shown
that
the performance from learning a distance measure using a way like the quantization approach~\cite{GordoPGL14}
or directly learning a distance lookup table~\cite{WangSYYLW14}\footnote{A similar idea is concurrently proposed
in~\cite{QinCGG14a,QinCGG14b} to learn a better similarity for a bag-of-words representation
and quantized kernels.}
from precomputed hash codes
is comparable to the performance of the Cartesian quantization approach
if the codes from the quantization approach are given as the input.

\textbf{Search cost:}
The evaluation of the Hamming distance
using the CPU instruction $\operatorname{\_\_popcnt}$
is faster than the distance-table lookup.
For example,
it is around twice faster for the same code length $L$
than distance table lookup
if a sub-table corresponds to a byte
and there are totally $\frac{L}{8}$ sub-tables.
It is worth pointing
(also observed in~\cite{ZhangDW14})
that the Cartesian quantization approaches relying on the distance table lookup
still achieve better search accuracy
even with a code of the half length,
which indicates that
the overall performance of the quantization approaches
in terms of space cost, query time cost, and search accuracy
is superior.

In summary,
if the online performance in terms of
space cost, query time cost,
and search accuracy
is cared about,
the quantization algorithms are suggested
for hash code ranking, hash table lookup,
as well as
the scheme of combining inverted index (hash table lookup) and hash code ranking.
The comparison of the query performances
of pairwise and multiwise similarity preserving algorithms, as well as quantization
is summarized in Table~\ref{table:queryperformance}.

Figure~\ref{fig:QuantizationOverHash} presents
$2$D toy examples.
Figure~\ref{fig:QuantizationOverHash} (a) shows
that the quantization algorithm is able to
discriminate the non-uniformly distributed clusters
with different between-cluster distances
while the binary code hashing algorithm is lacking such a capability
due to the Hamming distance.
Figure~\ref{fig:QuantizationOverHash} (b)
shows that the binary hash coding algorithms require
more ($6$) hash bits to differentiate the $16$ uniformly-distributed clusters
while the quantization algorithms only require $4$ ($= \log 16 $) bits.

\subsubsection{Empirical Results}
We present the empirical results of the several representative hashing and quantization algorithms
over SIFT$1M$~\cite{JegouDS11}.
We show the results for searching the nearest neighbor ($T=1$)
with $128$ bits
and the conclusion holds for searching more nearest neighbors ($T>1$)
and with other numbers of bits.
More results, such as the search time cost,
and results using inverted multi-index with different quantization algorithms
can be found in~\cite{ZhangQTW15}.
We also conduct experiments over
word feature vectors GloVe$1.2M$.
We present the results using
recall$@R$ for searching the nearest neighbor ($T=1$)
with $128$ bits.

We also report the results over deep learning features
extracted from the ILSVRC $2012$ dataset.
The ILSVRC $2012$ dataset is a subset of ImageNet~\cite{deng2009imagenet} and contains over $1.2$ million images.
We use the provided training set, $1,281,167$ images, as the retrieval database
and {use the provided validation set, $50,000$ images, as the test set}.
Similar to~\cite{ShenSLS15}, the $4096$-dimensional feature
extracted from the convolution neural networks (CNN)
in~\cite{krizhevsky2012imagenet} is used to represent each image.
We evaluate the search performance under the Euclidean distance
in terms of recall@$R$, where $R$ is the number of the returned top candidates,
and under the semantic similarity in terms of MAP vs. $\#$bits.

\begin{figure*}
\centering
\small{(a)}\includegraphics[width=.22\linewidth, clip]{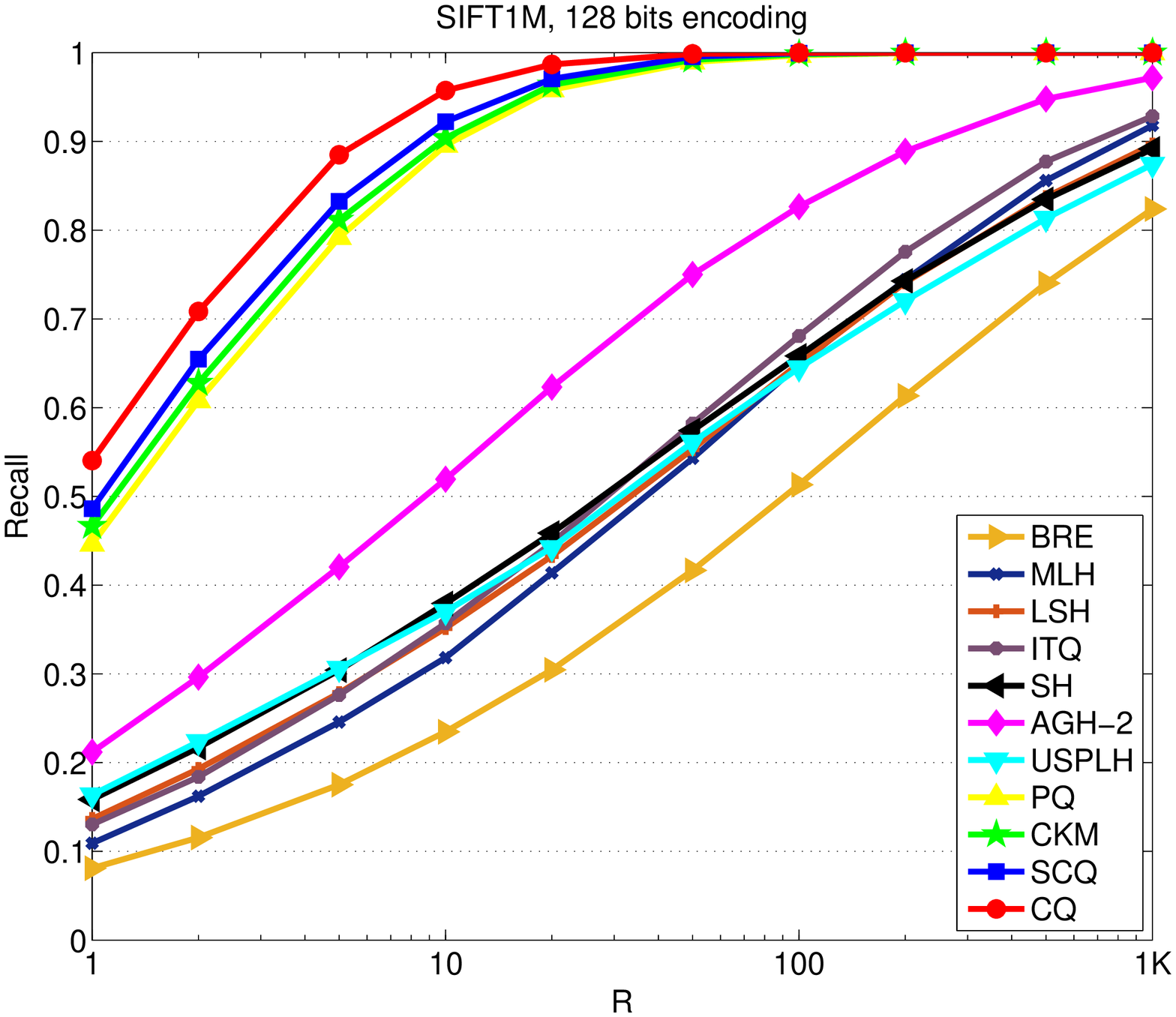}~
\small{(b)}\includegraphics[width=.22\linewidth, clip]{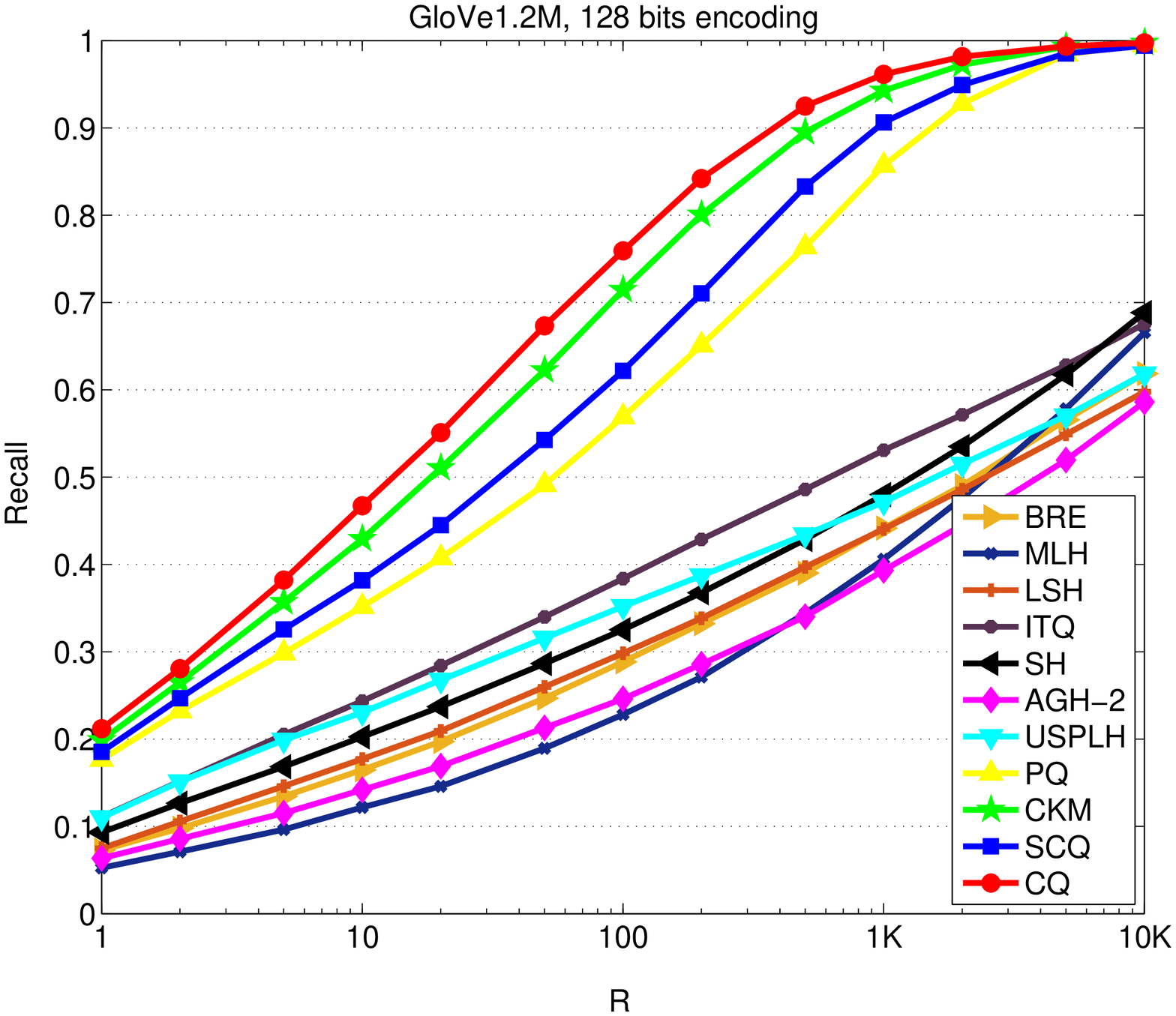}~
\small{(c)}\includegraphics[width=.22\linewidth, clip]{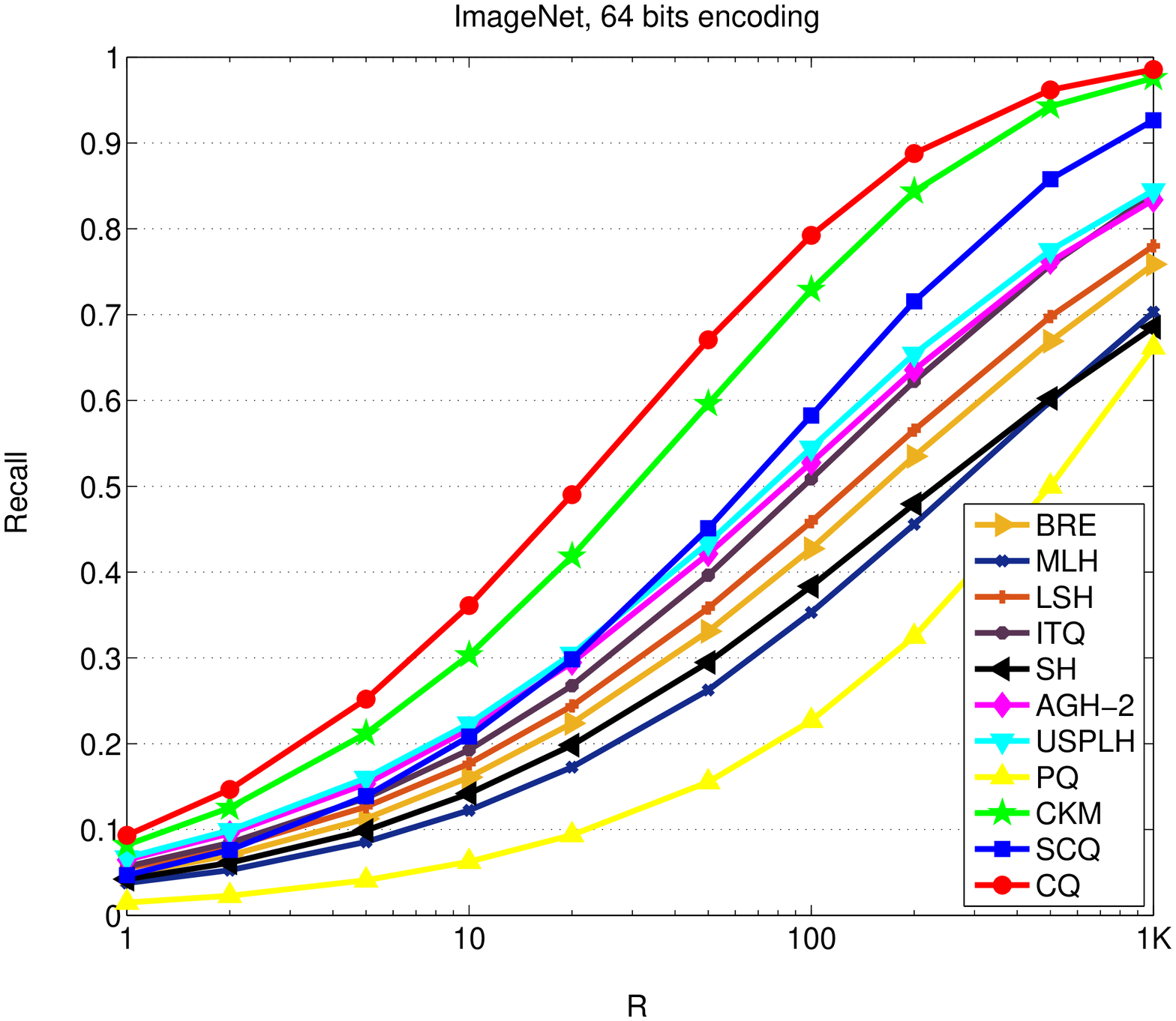}~
\small{(d)}\includegraphics[width=.22\linewidth, clip]{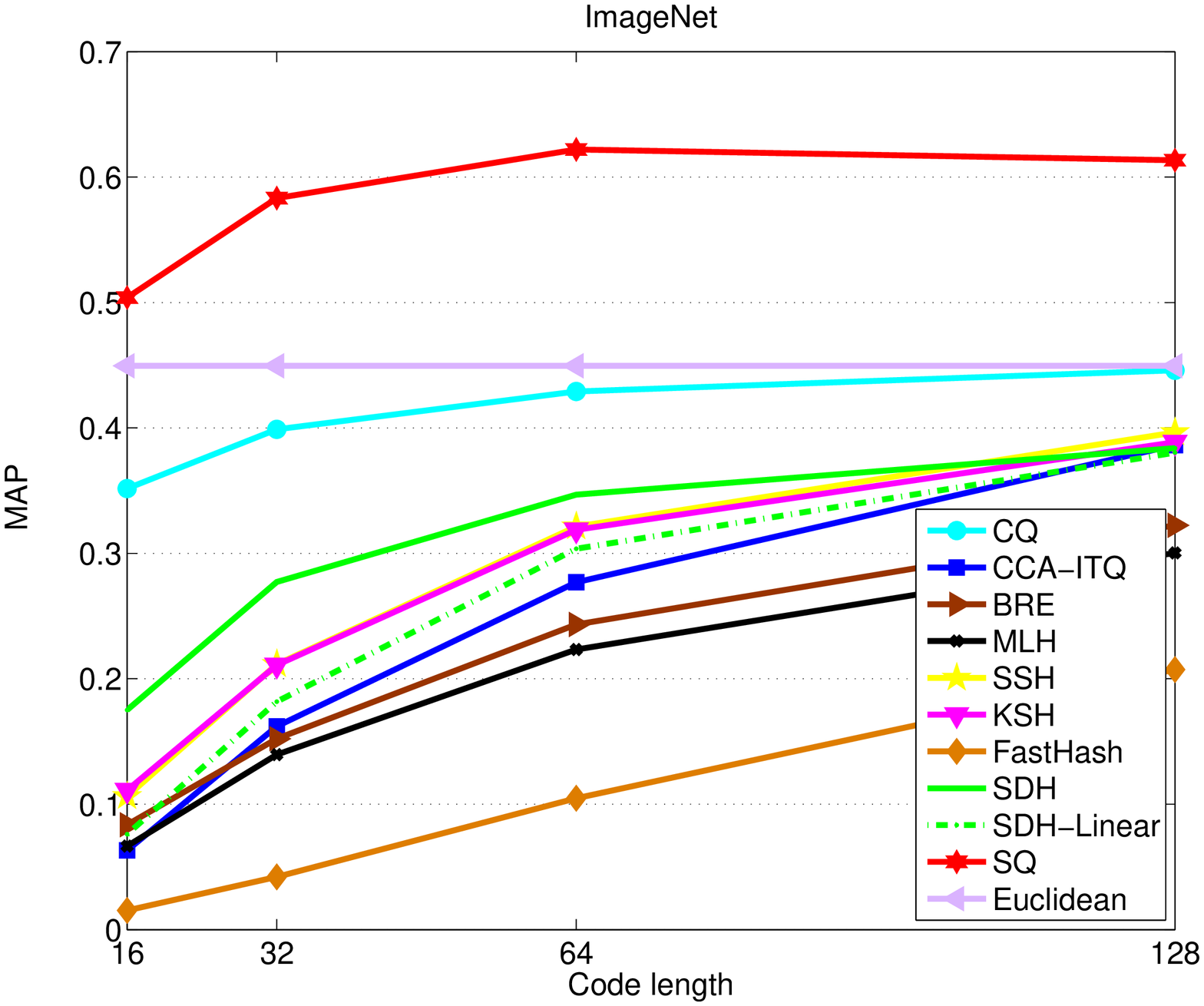}
\caption{(a) and (b) show
the performance in terms of recall@$R$
over SIFT$1M$ and GloVe$1.2M$
for the representative quantization algorithms.
(c) and (d) show the performance
over the ILSVRC 2012 ImageNet dataset
under the Euclidean distance
in terms of recall@$R$
and under the semantic similarity
in terms of mAP vs. $\#$ bits.
BRE = binary reconstructive embedding~\cite{KulisD09},
MLH = minimal loss hashing~\cite{NorouziF11},
LSH = locality sensitive hashing~\cite{Charikar02},
ITQ = iterative quantization~\cite{GongL11,GongLGP13},
SH = spectral hashing~\cite{WeissTF08},
AGH-2 = two-layer hashing with graphs~\cite{LiuWKC11},
USPLH = unsupervised sequential projection learning hashing~\cite{WangKC12},
PQ = product quantization~\cite{JegouDS11},
CKM = Cartesian $k$-means~\cite{NorouziF13},
CQ = composite quantization~\cite{ZhangDW14},
SCQ = sparse composite quantization~\cite{ZhangQTW15} whose dictionary is
the same sparse with PQ.
CCA-ITQ = iterative quantization with canonical correlation analysis~\cite{GongLGP13},
SSH = semi-supervised hashing~\cite{WangKC12},
KSH = supervised hashing with kernels~\cite{LiuWJJC12},
FastHash = fash supervised hashing~\cite{LinSSHS14},
SDH = supervised discrete hashing with kernels~\cite{ShenSLS15},
SDH-linear = supervised discrete hashing without using kernel representations~\cite{ShenSLS15},
SQ = supervised quantization~\cite{WangZQTW16},
Euclidean = linear scan with the Euclidean distance.}
\label{fig:ResultsRecallR}
\end{figure*}

Figure~\ref{fig:ResultsRecallR} shows the recall@$R$ curves and the MAP results.
We have several observations.
(1) The performance of the quantization method is better than the hashing method in most cases
for both Euclidean distance-based and semantic search.
(2) LSH, a data-independent algorithm is generally worse than other learning to hash approaches.
(3) For Euclidean distance-based search
the performance of CQ
is the best among quantization methods,
which is
consistent with the analysis
and the $2D$ illustration shown in Figure~\ref{fig:quantization}.

\subsection{Training Time Cost}
We present the analysis of the training time cost
for the case of using the linear hash function.
The pairwise similarity preserving category
considers the similarities
of all pairs of items,
and thus in general the training process
takes quadratic time
with respect to the number $N$ of the training samples
($O(N^2M + N^2d)$).
To reduce the computational cost,
sampling schemes are adopted:
sample a small number (e.g., $O(N)$) of pairs,
whose time complexity becomes linear with respect to $N$, resulting in
($O(NM + Nd)$),
or sample a subset of the training items
(e.g., containing $\bar{N}$ items),
whose time complexity becomes smaller
($O(\bar{N}^2M + \bar{N}^2d)$).
The multiwise similarity preserving category considers the similarities
of all triples of items,
and in general the training cost is greater
and the sampling scheme is also used for acceleration.
The analysis for kernel hash functions and other complex functions
is similar,
and the time complexity for both training hash functions
and encoding database items is higher.

Iterative quantization consists of a PCA preprocessing step
whose time complexity is $O(Nd^2)$,
and the hash code and hash function optimization step,
whose time complexity is $O(NM^2 + M^3)$ ($M$ is the number of hash bits).
The whole complexity is $O(Nd^2 + NM^2 + M^3)$.
Product quantization includes the $k$-means process for each partition,
and the complexity is $TNKP$,
where $K$ is usually $256$, $P = \frac{M}{8}$,
and $T$ is the number of iterations for the $k$-means algorithm.
The complexity of Cartesian $k$-means
is $O(Nd^2 + d^3)$.
The time complexity of composite quantization
is $O(NKPd + NP^2 + P^2K^2d)$.
In summary,
the time complexity of iterative quantization is the lowest
and that of composite quantization is the highest.
This indicates that it takes larger offline computation cost
to get a higher (online) search performance.
%The comparison of the query performances
%and the training cost
%of various quantization algorithms
%is summarized in Table~\ref{table:quantizationperformance}.
%In comparison to binary code hashing,
%the quantization category is in theory cheaper
%and both the categories can benefit
%from sampling a subset of items.

\section{Emerging Topics}
\label{sec:trends}
The main goal of the hashing algorithm
is to accelerate the online search
as the distance can be efficiently computed
through fast Hamming distance computation
or fast distance table lookup.
The offline hash function learning
and hash code computation
are shown to be still expensive,
and have become attractive in research.
The computation cost of the distance table used for looking up
is thought ignorable
and in reality could be higher
when handling high-dimensional databases.
There is also increasing interest
in topics
such as multi-modality and cross-modality hashing~\cite{IrieAT15}
and semantic quantization.

\subsection{Speed up the Learning and Query Processes}
\emph{Scalable Hash Function Learning.}
The algorithms depending on the pairwise similarity,
such as binary reconstructive embedding,
usually sample a small subset of pairs
to reduce the cost of learning hash functions.
It has been shown that
the search accuracy is increased
with a high sampling rate,
but the training cost is greatly increased.
The algorithms even without relying on the pairwise similarity,
e.g., quantization,
were also shown to be slow and even infeasible
when handling very large data, e.g., $1B$ data items,
and usually have to learn hash functions over a small subset, e.g., $1M$ data items.
This poses a challenging request
to learn the hash function over larger datasets.

\emph{Hash Code Computation Speedup.}
Existing hashing algorithms rarely take
into consideration the cost of encoding a data item.
Such a cost during the query stage becomes significant
in the case
that only a small number of database items
or a small database
are compared to the query.
The search combined with the
inverted index and compact codes
is such a case.
When kernel hash functions are used,
encoding the database items to binary codes
is also much more expensive
than that with linear hash functions.
The composite quantization-like approach also takes
much time
to compute the hash codes.

A recent work, circulant binary embedding~\cite{YuKGC14},
accelerates the encoding process for the linear hash functions,
and tree-quantization~\cite{BabenkoK15}
sparsifies the dictionary items
into a tree structure,
to speeding up the assignment process.
However, more research is needed
to speed up the hash code computation for other hashing algorithms,
such as composite quantization.

\emph{Distance Table Computation Speedup.}
Product quantization and its variants
need to precompute the distance table
between the query and the elements of the dictionaries.
Most existing algorithms claim that
the cost of distance table computation is negligible.
However in practice, the cost becomes bigger
when using the codes computed from quantization
to rank the candidates retrieved from the inverted index.
This is a research direction that will attract research interest in the near future,
such as a recent study, sparse composite quantization~\cite{ZhangQTW15}.

\subsection{Promising Extensions}
\emph{Semantic Quantization.}
Existing quantization algorithms focus on
the search under the Euclidean distances.
Like binary code hashing algorithms
where many studies on semantic similarity have been conducted,
learning quantization-based hash codes with semantic similarity
is attracting interest.
There are already a few studies.
For example, we have proposed an supervised quantization approach~\cite{WangZQTW16}
and some comparisons are provided in Figure~\ref{fig:ResultsRecallR}.

\emph{Multiple and Cross Modality Hashing.}
One important characteristic of big data is the variety of data types and data sources. This
is particularly true for multimedia data, where various media types (e.g., video, image, audio
and hypertext) can be described by many different low- and high-level features, and relevant
multimedia objects may come from different data sources contributed by different users and
organizations.
This raises a research direction,
performing joint-modality hashing learning
by exploiting the relation among multiple modalities,
for supporting some special applications, such as
cross-modal search.
This topic is attracting a lot of research efforts nowadays,
such as
collaborative hashing~\cite{LiuHDL14, ZhangSLHLC16},
collaborative quantization~\cite{ZhangW16},
and cross-media hashing~\cite{SongYHSL13},~\cite{SongYYHS13},~\cite{ZhuHSZ13},~\cite{XuSYSL17},~\cite{LiuLSSDH17}.

\section{Conclusion}
\label{sec:con}
In this paper,
we categorize the learning-to-hash algorithms
into four main groups:
pairwise similarity preserving,
multiwise similarity preserving,
implicit similarity preserving,
and quantization,
present a comprehensive survey
with a discussion
about their relations.
We point out the empirical observation
that quantization is superior
in terms of search accuracy,
search efficiency
and space cost.
In addition, we introduce a few emerging topics and the promising extensions.

\section*{Acknowledgements}
This work was partially supported by
the National Nature Science Foundation of China No. 61632007.

\bibliographystyle{ieee}
\bibliography{hash}

\begin{thebibliography}{100}\itemsep=-1pt

\bibitem{AndoniI06}
A.~Andoni and P.~Indyk.
\newblock Near-optimal hashing algorithms for approximate nearest neighbor in
  high dimensions.
\newblock In {\em FOCS}, pages 459--468, 2006.

\bibitem{BabenkoK14}
A.~Babenko and V.~Lempitsky.
\newblock Additive quantization for extreme vector compression.
\newblock In {\em CVPR}, pages 931--939, 2014.

\bibitem{BabenkoK15}
A.~Babenko and V.~Lempitsky.
\newblock Tree quantization for large-scale similarity search and
  classification.
\newblock In {\em CVPR}, 2015.

\bibitem{BabenkoL12}
A.~Babenko and V.~S. Lempitsky.
\newblock The inverted multi-index.
\newblock In {\em CVPR}, pages 3069--3076, 2012.

\bibitem{BaluFJ14}
R.~Balu, T.~Furon, and H.~Jegou.
\newblock Beyond "project and sign" for cosine estimation with binary codes.
\newblock In {\em ICASSP}, pages 6884--6888, 2014.

\bibitem{BentleySW77}
J.~L. Bentley, D.~F. Stanat, and E.~H.~W. Jr.
\newblock The complexity of finding fixed-radius near neighbors.
\newblock {\em Inf. Process. Lett.}, 6(6):209--212, 1977.

\bibitem{BergamoTF11}
A.~Bergamo, L.~Torresani, and A.~W. Fitzgibbon.
\newblock Picodes: Learning a compact code for novel-category recognition.
\newblock In {\em NIPS.}, pages 2088--2096, 2011.

\bibitem{BoufounosB08}
P.~Boufounos and R.~G. Baraniuk.
\newblock 1-bit compressive sensing.
\newblock In {\em {CISS}}, pages 16--21, 2008.

\bibitem{Brandt10}
J.~Brandt.
\newblock Transform coding for fast approximate nearest neighbor search in high
  dimensions.
\newblock In {\em CVPR}, pages 1815--1822, 2010.

\bibitem{Broder97}
A.~Z. Broder.
\newblock On the resemblance and containment of documents.
\newblock In {\em Proceedings of the Compression and Complexity of Sequences
  1997}, SEQUENCES '97, pages 21--29, Washington, DC, USA, 1997. IEEE Computer
  Society.

\bibitem{BroderGMZ97}
A.~Z. Broder, S.~C. Glassman, M.~S. Manasse, and G.~Zweig.
\newblock Syntactic clustering of the web.
\newblock {\em Computer Networks}, 29(8-13):1157--1166, 1997.

\bibitem{CakirS15}
F.~{\c{C}}akir and S.~Sclaroff.
\newblock Adaptive hashing for fast similarity search.
\newblock In {\em ICCV}, pages 1044--1052, 2015.

\bibitem{Carreira-Perpinan15}
M.~{\'{A}}. Carreira{-}Perpi{\~{n}}{\'{a}}n and R.~Raziperchikolaei.
\newblock Hashing with binary autoencoders.
\newblock In {\em CVPR}, pages 557--566, 2015.

\bibitem{Charikar02}
M.~Charikar.
\newblock Similarity estimation techniques from rounding algorithms.
\newblock In {\em STOC}, pages 380--388, 2002.

\bibitem{Cherian14}
A.~Cherian.
\newblock Nearest neighbors using compact sparse codes.
\newblock In {\em ICML (2)}, pages 1053--1061, 2014.

\bibitem{CherianMP12}
A.~Cherian, V.~Morellas, and N.~Papanikolopoulos.
\newblock Robust sparse hashing.
\newblock In {\em ICIP}, pages 2417--2420, 2012.

\bibitem{ChumM10}
O.~Chum and J.~Matas.
\newblock Large-scale discovery of spatially related images.
\newblock {\em {IEEE} Trans. Pattern Anal. Mach. Intell.}, 32(2):371--377,
  2010.

\bibitem{DaiLWJ16}
Q.~Dai, J.~Li, J.~Wang, and Y.~Jiang.
\newblock Binary optimized hashing.
\newblock In {\em {ACM} Multimedia}, pages 1247--1256, 2016.

\bibitem{DasguptaKS11}
A.~Dasgupta, R.~Kumar, and T.~Sarl{\'o}s.
\newblock Fast locality-sensitive hashing.
\newblock In {\em KDD}, pages 1073--1081, 2011.

\bibitem{DatarIIM04}
M.~Datar, N.~Immorlica, P.~Indyk, and V.~S. Mirrokni.
\newblock Locality-sensitive hashing scheme based on p-stable distributions.
\newblock In {\em Symposium on Computational Geometry}, pages 253--262, 2004.

\bibitem{DeanRSSVY13}
T.~L. Dean, M.~A. Ruzon, M.~Segal, J.~Shlens, S.~Vijayanarasimhan, and
  J.~Yagnik.
\newblock Fast, accurate detection of 100, 000 object classes on a single
  machine.
\newblock In {\em CVPR}, pages 1814--1821, 2013.

\bibitem{deng2009imagenet}
J.~Deng, W.~Dong, R.~Socher, L.-J. Li, K.~Li, and L.~Fei-Fei.
\newblock Imagenet: A large-scale hierarchical image database.
\newblock In {\em CVPR}, pages 248--255, 2009.

\bibitem{DingHFP15}
K.~Ding, C.~Huo, B.~Fan, and C.~Pan.
\newblock knn hashing with factorized neighborhood representation.
\newblock In {\em ICCV}, pages 1098--1106, 2015.

\bibitem{DoDC16}
T.~Do, A.~Doan, and N.~Cheung.
\newblock Learning to hash with binary deep neural network.
\newblock In {\em {ECCV}}, pages 219--234, 2016.

\bibitem{DoDNC16}
T.~Do, A.~Doan, D.~T. Nguyen, and N.~Cheung.
\newblock Binary hashing with semidefinite relaxation and augmented lagrangian.
\newblock In {\em {ECCV}}, pages 802--817, 2016.

\bibitem{DuW14}
C.~Du and J.~Wang.
\newblock Inner product similarity search using compositional codes.
\newblock {\em CoRR}, abs/1406.4966, 2014.

\bibitem{Fan13}
L.~Fan.
\newblock Supervised binary hash code learning with jensen shannon divergence.
\newblock In {\em ICCV}, pages 2616--2623, 2013.

\bibitem{FeiFP04}
L.~Fei-Fei, R.~Fergus, and P.~Perona.
\newblock Learning generative visual models from few training examples: an
  incremental bayesian approach tested on 101 object categories.
\newblock In {\em CVPR 2004 Workshop on Generative-Model Based Vision}, 2004.

\bibitem{GanFFN12}
J.~Gan, J.~Feng, Q.~Fang, and W.~Ng.
\newblock Locality-sensitive hashing scheme based on dynamic collision
  counting.
\newblock In {\em SIGMOD Conference}, pages 541--552, 2012.

\bibitem{GaoSZZS15}
L.~Gao, J.~Song, F.~Zou, D.~Zhang, and J.~Shao.
\newblock Scalable multimedia retrieval by deep learning hashing with relative
  similarity learning.
\newblock In {\em ACM Multimedia}, pages 903--906, 2015.

\bibitem{GeHK013}
T.~Ge, K.~He, Q.~Ke, and J.~Sun.
\newblock Optimized product quantization for approximate nearest neighbor
  search.
\newblock In {\em CVPR}, pages 2946--2953, 2013.

\bibitem{GeH014}
T.~Ge, K.~He, and J.~Sun.
\newblock Graph cuts for supervised binary coding.
\newblock In {\em {ECCV}}, pages 250--264, 2014.

\bibitem{GongKRL13}
Y.~Gong, S.~Kumar, H.~A. Rowley, and S.~Lazebnik.
\newblock Learning binary codes for high-dimensional data using bilinear
  projections.
\newblock In {\em CVPR}, pages 484--491, 2013.

\bibitem{GongKVL12}
Y.~Gong, S.~Kumar, V.~Verma, and S.~Lazebnik.
\newblock Angular quantization-based binary codes for fast similarity search.
\newblock In {\em NIPS}, pages 1205--1213, 2012.

\bibitem{GongL11}
Y.~Gong and S.~Lazebnik.
\newblock Iterative quantization: A procrustean approach to learning binary
  codes.
\newblock In {\em CVPR}, pages 817--824, 2011.

\bibitem{GongLGP13}
Y.~Gong, S.~Lazebnik, A.~Gordo, and F.~Perronnin.
\newblock Iterative quantization: A procrustean approach to learning binary
  codes for large-scale image retrieval.
\newblock {\em IEEE Trans. Pattern Anal. Mach. Intell.}, 35(12):2916--2929,
  2013.

\bibitem{GordoPGL14}
A.~Gordo, F.~Perronnin, Y.~Gong, and S.~Lazebnik.
\newblock Asymmetric distances for binary embeddings.
\newblock {\em IEEE Trans. Pattern Anal. Mach. Intell.}, 36(1):33--47, 2014.

\bibitem{GrayN98}
R.~M. Gray and D.~L. Neuhoff.
\newblock Quantization.
\newblock {\em IEEE Transactions on Information Theory}, 44(6):2325--2383,
  1998.

\bibitem{HeCRB11}
J.~He, S.-F. Chang, R.~Radhakrishnan, and C.~Bauer.
\newblock Compact hashing with joint optimization of search accuracy and time.
\newblock In {\em CVPR}, pages 753--760, 2011.

\bibitem{HeLC10}
J.~He, W.~Liu, and S.-F. Chang.
\newblock Scalable similarity search with optimized kernel hashing.
\newblock In {\em KDD}, pages 1129--1138, 2010.

\bibitem{HeoLHCY12}
J.-P. Heo, Y.~Lee, J.~He, S.-F. Chang, and S.-E. Yoon.
\newblock Spherical hashing.
\newblock In {\em CVPR}, pages 2957--2964, 2012.

\bibitem{HeoLY14}
J.-P. Heo, Z.~Lin, and S.-E. Yoon.
\newblock Distance encoded product quantization.
\newblock In {\em CVPR}, pages 2139--2146, 2014.

\bibitem{HuangYZ13}
L.-K. Huang, Q.~Yang, and W.-S. Zheng.
\newblock Online hashing.
\newblock In {\em IJCAI}, 2013.

\bibitem{IndykM98}
P.~Indyk and R.~Motwani.
\newblock Approximate nearest neighbors: Towards removing the curse of
  dimensionality.
\newblock In {\em STOC}, pages 604--613, 1998.

\bibitem{IrieAT15}
G.~Irie, H.~Arai, and Y.~Taniguchi.
\newblock Alternating co-quantization for cross-modal hashing.
\newblock In {\em ICCV}, pages 1886--1894, 2015.

\bibitem{IrieLWC14}
G.~Irie, Z.~Li, X.-M. Wu, and S.-F. Chang.
\newblock Locally linear hashing for extracting non-linear manifolds.
\newblock In {\em CVPR}, pages 2123--2130, 2014.

\bibitem{JainPGZJ16}
H.~Jain, P.~P{\'{e}}rez, R.~Gribonval, J.~Zepeda, and H.~J{\'{e}}gou.
\newblock Approximate search with quantized sparse representations.
\newblock In {\em {ECCV}}, pages 681--696, 2016.

\bibitem{JegouASG08}
H.~Jegou, L.~Amsaleg, C.~Schmid, and P.~Gros.
\newblock Query adaptative locality sensitive hashing.
\newblock In {\em {ICASSP}}, pages 825--828, 2008.

\bibitem{JegouDS08}
H.~Jegou, M.~Douze, and C.~Schmid.
\newblock Hamming embedding and weak geometric consistency for large scale
  image search.
\newblock In {\em ECCV}, pages 304--317, 2008.

\bibitem{JegouDS11}
H.~J{\'e}gou, M.~Douze, and C.~Schmid.
\newblock Product quantization for nearest neighbor search.
\newblock {\em IEEE Trans. Pattern Anal. Mach. Intell.}, 33(1):117--128, 2011.

\bibitem{JegouDSP10}
H.~J{\'e}gou, M.~Douze, C.~Schmid, and P.~P{\'e}rez.
\newblock Aggregating local descriptors into a compact image representation.
\newblock In {\em CVPR}, pages 3304--3311, 2010.

\bibitem{JegouFF12}
H.~J{\'e}gou, T.~Furon, and J.-J. Fuchs.
\newblock Anti-sparse coding for approximate nearest neighbor search.
\newblock In {\em ICASSP}, pages 2029--2032, 2012.

\bibitem{JegouTDA11}
H.~J{\'e}gou, R.~Tavenard, M.~Douze, and L.~Amsaleg.
\newblock Searching in one billion vectors: Re-rank with source coding.
\newblock In {\em ICASSP}, pages 861--864, 2011.

\bibitem{JiLYTZ13}
J.~Ji, J.~Li, S.~Yan, Q.~Tian, and B.~Zhang.
\newblock Min-max hash for jaccard similarity.
\newblock In {\em ICDM}, pages 301--309, 2013.

\bibitem{JiLYZT12}
J.~Ji, J.~Li, S.~Yan, B.~Zhang, and Q.~Tian.
\newblock Super-bit locality-sensitive hashing.
\newblock In {\em NIPS}, pages 108--116, 2012.

\bibitem{JiangL15}
Q.~Jiang and W.~Li.
\newblock Scalable graph hashing with feature transformation.
\newblock In {\em IJCAI}, pages 2248--2254, 2015.

\bibitem{JiangWC11}
Y.-G. Jiang, J.~Wang, and S.-F. Chang.
\newblock Lost in binarization: query-adaptive ranking for similar image search
  with compact codes.
\newblock In {\em ICMR}, page~16, 2011.

\bibitem{JiangWXC13}
Y.-G. Jiang, J.~Wang, X.~Xue, and S.-F. Chang.
\newblock Query-adaptive image search with hash codes.
\newblock {\em IEEE Transactions on Multimedia}, 15(2):442--453, 2013.

\bibitem{JiangXDXW16}
Z.~Jiang, L.~Xie, X.~Deng, W.~Xu, and J.~Wang.
\newblock Fast nearest neighbor search in the hamming space.
\newblock In {\em {MMM}}, pages 325--336, 2016.

\bibitem{JinHLZLCL13}
Z.~Jin, Y.~Hu, Y.~Lin, D.~Zhang, S.~Lin, D.~Cai, and X.~Li.
\newblock Complementary projection hashing.
\newblock In {\em ICCV}, pages 257--264, 2013.

\bibitem{JolyB11}
A.~Joly and O.~Buisson.
\newblock Random maximum margin hashing.
\newblock In {\em CVPR}, pages 873--880, 2011.

\bibitem{KalantidisA14}
Y.~Kalantidis and Y.~Avrithis.
\newblock Locally optimized product quantization for approximate nearest
  neighbor search.
\newblock In {\em CVPR}, pages 2329--2336, 2014.

\bibitem{KongL12a}
W.~Kong and W.-J. Li.
\newblock Isotropic hashing.
\newblock In {\em NIPS}, pages 1655--1663, 2012.

\bibitem{KoudasOST04}
N.~Koudas, B.~C. Ooi, H.~T. Shen, and A.~K.~H. Tung.
\newblock Ldc: Enabling search by partial distance in a hyper-dimensional
  space.
\newblock In {\em ICDE}, pages 6--17, 2004.

\bibitem{krizhevsky2012imagenet}
A.~Krizhevsky, I.~Sutskever, and G.~E. Hinton.
\newblock Imagenet classification with deep convolutional neural networks.
\newblock In {\em NIPS}, pages 1097--1105, 2012.

\bibitem{KulisD09}
B.~Kulis and T.~Darrell.
\newblock Learning to hash with binary reconstructive embeddings.
\newblock In {\em NIPS}, pages 1042--1050, 2009.

\bibitem{LaiPLY15}
H.~Lai, Y.~Pan, Y.~Liu, and S.~Yan.
\newblock Simultaneous feature learning and hash coding with deep neural
  networks.
\newblock In {\em CVPR}, pages 3270--3278, 2015.

\bibitem{LeCunBBH01}
Y.~LeCun, L.~Bottou, Y.~Bengio, and P.~Haffner.
\newblock Gradient-based learning applied to document recognition.
\newblock In {\em Intelligent Signal Processing}, pages 306--351. IEEE Press,
  2001.

\bibitem{LengWC0L15}
C.~Leng, J.~Wu, J.~Cheng, X.~Bai, and H.~Lu.
\newblock Online sketching hashing.
\newblock In {\em CVPR}, pages 2503--2511, 2015.

\bibitem{LiCH06}
P.~Li, K.~W. Church, and T.~Hastie.
\newblock Conditional random sampling: A sketch-based sampling technique for
  sparse data.
\newblock In {\em NIPS}, pages 873--880, 2006.

\bibitem{LiHC06}
P.~Li, T.~Hastie, and K.~W. Church.
\newblock Very sparse random projections.
\newblock In {\em KDD}, pages 287--296, 2006.

\bibitem{LiK10b}
P.~Li and A.~C. K{\"o}nig.
\newblock b-bit minwise hashing.
\newblock In {\em WWW}, pages 671--680, 2010.

\bibitem{LiKG10a}
P.~Li, A.~C. K{\"o}nig, and W.~Gui.
\newblock b-bit minwise hashing for estimating three-way similarities.
\newblock In {\em NIPS}, pages 1387--1395, 2010.

\bibitem{LiOZ12}
P.~Li, A.~B. Owen, and C.-H. Zhang.
\newblock One permutation hashing.
\newblock In {\em NIPS}, pages 3122--3130, 2012.

\bibitem{LiWCXL13}
P.~Li, M.~Wang, J.~Cheng, C.~Xu, and H.~Lu.
\newblock Spectral hashing with semantically consistent graph for image
  indexing.
\newblock {\em IEEE Transactions on Multimedia}, 15(1):141--152, 2013.

\bibitem{LinSSHS14}
G.~Lin, C.~Shen, Q.~Shi, A.~van\_den\_Hengel, and D.~Suter.
\newblock Fast supervised hashing with decision trees for high-dimensional
  data.
\newblock In {\em CVPR}, pages 1971--1978, 2014.

\bibitem{LinSSH13}
G.~Lin, C.~Shen, D.~Suter, and A.~van~den Hengel.
\newblock A general two-step approach to learning-based hashing.
\newblock In {\em ICCV}, pages 2552--2559, 2013.

\bibitem{LinRY10}
R.-S. Lin, D.~A. Ross, and J.~Yagnik.
\newblock Spec hashing: Similarity preserving algorithm for entropy-based
  coding.
\newblock In {\em CVPR}, pages 848--854, 2010.

\bibitem{LinJCYL13}
Y.~Lin, R.~Jin, D.~Cai, S.~Yan, and X.~Li.
\newblock Compressed hashing.
\newblock In {\em CVPR}, pages 446--451, 2013.

\bibitem{LiongLWMZ15}
V.~E. Liong, J.~Lu, G.~Wang, P.~Moulin, and J.~Zhou.
\newblock Deep hashing for compact binary codes learning.
\newblock In {\em CVPR}, pages 2475--2483, 2015.

\bibitem{LiuYJHZ13}
D.~Liu, S.~Yan, R.-R. Ji, X.-S. Hua, and H.-J. Zhang.
\newblock Image retrieval with query-adaptive hashing.
\newblock {\em TOMCCAP}, 9(1):2, 2013.

\bibitem{Liu0SC16}
H.~Liu, R.~Wang, S.~Shan, and X.~Chen.
\newblock Deep supervised hashing for fast image retrieval.
\newblock In {\em {CVPR}}, pages 2064--2072, 2016.

\bibitem{LiuLSSDH17}
L.~Liu, Z.~Lin, L.~Shao, F.~Shen, G.~Ding, and J.~Han.
\newblock Sequential discrete hashing for scalable cross-modality similarity
  retrieval.
\newblock {\em {IEEE} Trans. Image Processing}, 26(1):107--118, 2017.

\bibitem{LiuMKC14}
W.~Liu, C.~Mu, S.~Kumar, and S.~Chang.
\newblock Discrete graph hashing.
\newblock In {\em NIPS}, pages 3419--3427, 2014.

\bibitem{LiuWJJC12}
W.~Liu, J.~Wang, R.~Ji, Y.-G. Jiang, and S.-F. Chang.
\newblock Supervised hashing with kernels.
\newblock In {\em CVPR}, pages 2074--2081, 2012.

\bibitem{LiuWKC11}
W.~Liu, J.~Wang, S.~Kumar, and S.-F. Chang.
\newblock Hashing with graphs.
\newblock In {\em ICML}, pages 1--8, 2011.

\bibitem{LiuWMKC12}
W.~Liu, J.~Wang, Y.~Mu, S.~Kumar, and S.-F. Chang.
\newblock Compact hyperplane hashing with bilinear functions.
\newblock In {\em ICML}, 2012.

\bibitem{LiuDLTL16}
X.~Liu, C.~Deng, B.~Lang, D.~Tao, and X.~Li.
\newblock Query-adaptive reciprocal hash tables for nearest neighbor search.
\newblock {\em {IEEE} Transactions on Image Processing}, 25(2):907--919, 2016.

\bibitem{LiuHDL14}
X.~Liu, J.~He, C.~Deng, and B.~Lang.
\newblock Collaborative hashing.
\newblock In {\em CVPR}, pages 2147--2154, 2014.

\bibitem{LiuHL13}
X.~Liu, J.~He, and B.~Lang.
\newblock Reciprocal hash tables for nearest neighbor search.
\newblock In {\em AAAI}, 2013.

\bibitem{LiuHDLL15}
X.~Liu, L.~Huang, C.~Deng, J.~Lu, and B.~Lang.
\newblock Multi-view complementary hash tables for nearest neighbor search.
\newblock In {\em ICCV}, pages 1107--1115, 2015.

\bibitem{LiuSXWZ13}
Y.~Liu, J.~Shao, J.~Xiao, F.~Wu, and Y.~Zhuang.
\newblock Hypergraph spectral hashing for image retrieval with heterogeneous
  social contexts.
\newblock {\em Neurocomputing}, 119:49--58, 2013.

\bibitem{LiuWYZH12}
Y.~Liu, F.~Wu, Y.~Yang, Y.~Zhuang, and A.~G. Hauptmann.
\newblock Spline regression hashing for fast image search.
\newblock {\em IEEE Transactions on Image Processing}, 21(10):4480--4491, 2012.

\bibitem{Lowe04}
D.~G. Lowe.
\newblock Distinctive image features from scale-invariant keypoints.
\newblock {\em International Journal of Computer Vision}, 60(2):91--110, 2004.

\bibitem{LvJWCL07}
Q.~Lv, W.~Josephson, Z.~Wang, M.~Charikar, and K.~Li.
\newblock Multi-probe lsh: Efficient indexing for high-dimensional similarity
  search.
\newblock In {\em VLDB}, pages 950--961, 2007.

\bibitem{MartinezCHL16}
J.~Martinez, J.~Clement, H.~H. Hoos, and J.~J. Little.
\newblock Revisiting additive quantization.
\newblock In {\em {ECCV}}, pages 137--153, 2016.

\bibitem{MatsuiYA15}
Y.~Matsui, T.~Yamasaki, and K.~Aizawa.
\newblock Pqtable: Fast exact asymmetric distance neighbor search for product
  quantization using hash tables.
\newblock In {\em ICCV}, pages 1940--1948, 2015.

\bibitem{MatsushitaW09}
Y.~Matsushita and T.~Wada.
\newblock Principal component hashing: An accelerated approximate nearest
  neighbor search.
\newblock In {\em PSIVT}, pages 374--385, 2009.

\bibitem{MoonNPLSHP16}
Y.~Moon, S.~Noh, D.~Park, C.~Luo, A.~Shrivastava, S.~Hong, and K.~Palem.
\newblock Capsule: A camera-based positioning system using learning.
\newblock In {\em {SOCC}}, 2015.

\bibitem{MotwaniNP07}
R.~Motwani, A.~Naor, and R.~Panigrahy.
\newblock Lower bounds on locality sensitive hashing.
\newblock {\em SIAM J. Discrete Math.}, 21(4):930--935, 2007.

\bibitem{MuCLCY12}
Y.~Mu, X.~Chen, X.~Liu, T.-S. Chua, and S.~Yan.
\newblock Multimedia semantics-aware query-adaptive hashing with bits
  reconfigurability.
\newblock {\em IJMIR}, 1(1):59--70, 2012.

\bibitem{MuSY10}
Y.~Mu, J.~Shen, and S.~Yan.
\newblock Weakly-supervised hashing in kernel space.
\newblock In {\em CVPR}, pages 3344--3351, 2010.

\bibitem{MujaL09}
M.~Muja and D.~G. Lowe.
\newblock Fast approximate nearest neighbors with automatic algorithm
  configuration.
\newblock In {\em VISSAPP (1)}, pages 331--340, 2009.

\bibitem{MujaL12}
M.~Muja and D.~G. Lowe.
\newblock Fast matching of binary features.
\newblock In {\em CRV}, pages 404--410, 2012.

\bibitem{MujaL14}
M.~Muja and D.~G. Lowe.
\newblock Scalable nearest neighbor algorithms for high dimensional data.
\newblock {\em {IEEE} Trans. Pattern Anal. Mach. Intell.}, 36(11):2227--2240,
  2014.

\bibitem{MukherjeeRIHS15}
L.~Mukherjee, S.~N. Ravi, V.~K. Ithapu, T.~Holmes, and V.~Singh.
\newblock An {NMF} perspective on binary hashing.
\newblock In {\em ICCV}, pages 4184--4192, 2015.

\bibitem{NorouziF11}
M.~Norouzi and D.~J. Fleet.
\newblock Minimal loss hashing for compact binary codes.
\newblock In {\em ICML}, pages 353--360, 2011.

\bibitem{NorouziF13}
M.~Norouzi and D.~J. Fleet.
\newblock Cartesian k-means.
\newblock In {\em CVPR}, pages 3017--3024, 2013.

\bibitem{NorouziFS12}
M.~Norouzi, D.~J. Fleet, and R.~Salakhutdinov.
\newblock Hamming distance metric learning.
\newblock In {\em NIPS}, pages 1070--1078, 2012.

\bibitem{NorouziPF12}
M.~Norouzi, A.~Punjani, and D.~J. Fleet.
\newblock Fast search in hamming space with multi-index hashing.
\newblock In {\em CVPR}, pages 3108--3115, 2012.

\bibitem{ODonnellWZ11}
R.~O'Donnell, Y.~Wu, and Y.~Zhou.
\newblock Optimal lower bounds for locality sensitive hashing (except when q is
  tiny).
\newblock In {\em ICS}, pages 275--283, 2011.

\bibitem{OlivaT01}
A.~Oliva and A.~Torralba.
\newblock Modeling the shape of the scene: A holistic representation of the
  spatial envelope.
\newblock {\em International Journal of Computer Vision}, 42(3):145--175, 2001.

\bibitem{Panigrahy06}
R.~Panigrahy.
\newblock Entropy based nearest neighbor search in high dimensions.
\newblock In {\em SODA}, pages 1186--1195, 2006.

\bibitem{PauleveJA10}
L.~Paulev{\'e}, H.~J{\'e}gou, and L.~Amsaleg.
\newblock Locality sensitive hashing: A comparison of hash function types and
  querying mechanisms.
\newblock {\em Pattern Recognition Letters}, 31(11):1348--1358, 2010.

\bibitem{pennington2014glove}
J.~Pennington, R.~Socher, and C.~D. Manning.
\newblock Glove: Global vectors for word representation.
\newblock In {\em Empirical Methods in Natural Language Processing (EMNLP)},
  pages 1532--1543, 2014.

\bibitem{PerronninLSP10}
F.~Perronnin, Y.~Liu, J.~S{\'{a}}nchez, and H.~Poirier.
\newblock Large-scale image retrieval with compressed fisher vectors.
\newblock In {\em {CVPR}}, pages 3384--3391, 2010.

\bibitem{QinCGG14b}
D.~Qin, X.~Chen, M.~Guillaumin, and L.~J.~V. Gool.
\newblock Quantized kernel learning for feature matching.
\newblock In {\em {NIPS}}, pages 172--180, 2014.

\bibitem{QinCGG14a}
D.~Qin, Y.~Chen, M.~Guillaumin, and L.~J.~V. Gool.
\newblock Learning to rank histograms for object retrieval.
\newblock In {\em {BMVC}}, 2014.

\bibitem{RussellTMF08}
B.~C. Russell, A.~Torralba, K.~P. Murphy, and W.~T. Freeman.
\newblock Labelme: {A} database and web-based tool for image annotation.
\newblock {\em International Journal of Computer Vision}, 77(1-3):157--173,
  2008.

\bibitem{SalakhutdinovH07}
R.~Salakhutdinov and G.~E. Hinton.
\newblock Semantic hashing.
\newblock In {\em SIGIR workshop on Information Retrieval and applications of
  Graphical Models}, 2007.

\bibitem{SalakhutdinovH09}
R.~Salakhutdinov and G.~E. Hinton.
\newblock Semantic hashing.
\newblock {\em Int. J. Approx. Reasoning}, 50(7):969--978, 2009.

\bibitem{SanchezP11}
J.~S{\'{a}}nchez and F.~Perronnin.
\newblock High-dimensional signature compression for large-scale image
  classification.
\newblock In {\em CVPR}, pages 1665--1672, 2011.

\bibitem{SandhawaliaJ10}
H.~Sandhawalia and H.~Jegou.
\newblock Searching with expectations.
\newblock In {\em {ICASSP}}, pages 1242--1245, 2010.

\bibitem{ShaoWOZ12}
J.~Shao, F.~Wu, C.~Ouyang, and X.~Zhang.
\newblock Sparse spectral hashing.
\newblock {\em Pattern Recognition Letters}, 33(3):271--277, 2012.

\bibitem{ShenSLS15}
F.~Shen, C.~Shen, W.~Liu, and H.~T. Shen.
\newblock Supervised discrete hashing.
\newblock In {\em {CVPR}}, pages 37--45, 2015.

\bibitem{ShenSSHT13}
F.~Shen, C.~Shen, Q.~Shi, A.~van~den Hengel, and Z.~Tang.
\newblock Inductive hashing on manifolds.
\newblock In {\em CVPR}, pages 1562--1569, 2013.

\bibitem{ShenZ0SST16}
F.~Shen, X.~Zhou, Y.~Yang, J.~Song, H.~T. Shen, and D.~Tao.
\newblock A fast optimization method for general binary code learning.
\newblock {\em {IEEE} Trans. Image Processing}, 25(12):5610--5621, 2016.

\bibitem{ShiXCZXY16}
X.~Shi, F.~Xing, J.~Cai, Z.~Zhang, Y.~Xie, and L.~Yang.
\newblock Kernel-based supervised discrete hashing for image retrieval.
\newblock In {\em {ECCV}}, pages 419--433, 2016.

\bibitem{ShrivastavaL12}
A.~Shrivastava and P.~Li.
\newblock Fast near neighbor search in high-dimensional binary data.
\newblock In {\em {ECML} {PKDD}}, pages 474--489, 2012.

\bibitem{ShrivastavaL14}
A.~Shrivastava and P.~Li.
\newblock Densifying one permutation hashing via rotation for fast near
  neighbor.
\newblock In {\em ICML (1)}, pages 557--65, 2014.

\bibitem{SnavelySS06}
N.~Snavely, S.~M. Seitz, and R.~Szeliski.
\newblock Photo tourism: exploring photo collections in 3d.
\newblock {\em {ACM} Trans. Graph.}, 25(3):835--846, 2006.

\bibitem{SongLJMS15}
D.~Song, W.~Liu, R.~Ji, D.~A. Meyer, and J.~R. Smith.
\newblock Top rank supervised binary coding for visual search.
\newblock In {\em ICCV}, pages 1922--1930, 2015.

\bibitem{SongSWHSW16}
J.~Song, H.~T. Shen, J.~Wang, Z.~Huang, N.~Sebe, and J.~Wang.
\newblock A distance-computation-free search scheme for binary code databases.
\newblock {\em {IEEE} Trans. Multimedia}, 18(3):484--495, 2016.

\bibitem{SongYHSL13}
J.~Song, Y.~Yang, Z.~Huang, H.~T. Shen, and J.~Luo.
\newblock Effective multiple feature hashing for large-scale near-duplicate
  video retrieval.
\newblock {\em IEEE Transactions on Multimedia}, 15(8):1997--2008, 2013.

\bibitem{SongYYHS13}
J.~Song, Y.~Yang, Y.~Yang, Z.~Huang, and H.~T. Shen.
\newblock Inter-media hashing for large-scale retrieval from heterogeneous data
  sources.
\newblock In {\em SIGMOD Conference}, pages 785--796, 2013.

\bibitem{StrechaBBF12}
C.~Strecha, A.~M. Bronstein, M.~M. Bronstein, and P.~Fua.
\newblock Ldahash: Improved matching with smaller descriptors.
\newblock {\em IEEE Trans. Pattern Anal. Mach. Intell.}, 34(1):66--78, 2012.

\bibitem{TorralbaFF08}
A.~B. Torralba, R.~Fergus, and W.~T. Freeman.
\newblock 80 million tiny images: A large data set for nonparametric object and
  scene recognition.
\newblock {\em IEEE Trans. Pattern Anal. Mach. Intell.}, 30(11):1958--1970,
  2008.

\bibitem{VedaldiZisserman12}
A.~Vedaldi and A.~Zisserman.
\newblock Efficient additive kernels via explicit feature maps.
\newblock {\em {IEEE} Trans. Pattern Anal. Mach. Intell.}, 34(3):480--492,
  2012.

\bibitem{VedaldiZ12}
A.~Vedaldi and A.~Zisserman.
\newblock Sparse kernel approximations for efficient classification and
  detection.
\newblock In {\em CVPR}, pages 2320--2327, 2012.

\bibitem{AhnLB06}
L.~von Ahn, R.~Liu, and M.~Blum.
\newblock Peekaboom: a game for locating objects in images.
\newblock In {\em CHI}, pages 55--64, 2006.

\bibitem{WangKC10a}
J.~Wang, O.~Kumar, and S.-F. Chang.
\newblock Semi-supervised hashing for scalable image retrieval.
\newblock In {\em CVPR}, pages 3424--3431, 2010.

\bibitem{WangKC10b}
J.~Wang, S.~Kumar, and S.-F. Chang.
\newblock Sequential projection learning for hashing with compact codes.
\newblock In {\em ICML}, pages 1127--1134, 2010.

\bibitem{WangKC12}
J.~Wang, S.~Kumar, and S.-F. Chang.
\newblock Semi-supervised hashing for large-scale search.
\newblock {\em IEEE Trans. Pattern Anal. Mach. Intell.}, 34(12):2393--2406,
  2012.

\bibitem{WangL12}
J.~Wang and S.~Li.
\newblock Query-driven iterated neighborhood graph search for large scale
  indexing.
\newblock In {\em {ACM} Multimedia}, pages 179--188, 2012.

\bibitem{WangLKC16}
J.~Wang, W.~Liu, S.~Kumar, and S.~Chang.
\newblock Learning to hash for indexing big data - {A} survey.
\newblock {\em Proceedings of the {IEEE}}, 104(1):34--57, 2016.

\bibitem{WangLSJ13}
J.~Wang, W.~Liu, A.~X. Sun, and Y.-G. Jiang.
\newblock Learning hash codes with listwise supervision.
\newblock In {\em ICCV}, pages 3032--3039, 2013.

\bibitem{WangSSJ14}
J.~Wang, H.~T. Shen, J.~Song, and J.~Ji.
\newblock Hashing for similarity search: {A} survey.
\newblock {\em CoRR}, abs/1408.2927, 2014.

\bibitem{WangSYYLW14}
J.~Wang, H.~T. Shen, S.~Yan, N.~Yu, S.~Li, and J.~Wang.
\newblock Optimized distances for binary code ranking.
\newblock In {\em ACM Multimedia}, pages 517--526, 2014.

\bibitem{WangWSXSL14}
J.~Wang, J.~Wang, J.~Song, X.-S. Xu, H.~T. Shen, and S.~Li.
\newblock Optimized cartesian $k$-means.
\newblock {\em CoRR}, abs/1405.4054, 2014.

\bibitem{WangWYL13}
J.~Wang, J.~Wang, N.~Yu, and S.~Li.
\newblock Order preserving hashing for approximate nearest neighbor search.
\newblock In {\em ACM Multimedia}, pages 133--142, 2013.

\bibitem{WangWZGLG13}
J.~Wang, J.~Wang, G.~Zeng, R.~Gan, S.~Li, and B.~Guo.
\newblock Fast neighborhood graph search using cartesian concatenation.
\newblock In {\em ICCV}, pages 2128--2135, 2013.

\bibitem{WangWJLZZH13}
J.~Wang, N.~Wang, Y.~Jia, J.~Li, G.~Zeng, H.~Zha, and X.-S. Hua.
\newblock Trinary-projection trees for approximate nearest neighbor search.
\newblock {\em IEEE Trans. Pattern Anal. Mach. Intell.}, 2013.

\bibitem{WangZS13}
Q.~Wang, D.~Zhang, and L.~Si.
\newblock Weighted hashing for fast large scale similarity search.
\newblock In {\em CIKM}, pages 1185--1188, 2013.

\bibitem{WangZQTW16}
X.~Wang, T.~Zhang, G.-J. Qi, J.~Tang, and J.~Wang.
\newblock Supervised quantization for similarity search.
\newblock In {\em CVPR}, 2016.

\bibitem{WeissFT12}
Y.~Weiss, R.~Fergus, and A.~Torralba.
\newblock Multidimensional spectral hashing.
\newblock In {\em ECCV (5)}, pages 340--353, 2012.

\bibitem{WeissTF08}
Y.~Weiss, A.~Torralba, and R.~Fergus.
\newblock Spectral hashing.
\newblock In {\em NIPS}, pages 1753--1760, 2008.

\bibitem{WuZCCB13}
C.~Wu, J.~Zhu, D.~Cai, C.~Chen, and J.~Bu.
\newblock Semi-supervised nonlinear hashing using bootstrap sequential
  projection learning.
\newblock {\em IEEE Trans. Knowl. Data Eng.}, 25(6):1380--1393, 2013.

\bibitem{XiaPLLY14}
R.~Xia, Y.~Pan, H.~Lai, C.~Liu, and S.~Yan.
\newblock Supervised hashing for image retrieval via image representation
  learning.
\newblock In {\em AAAI}, pages 2156--2162, 2014.

\bibitem{XuBLCHC13}
B.~Xu, J.~Bu, Y.~Lin, C.~Chen, X.~He, and D.~Cai.
\newblock Harmonious hashing.
\newblock In {\em IJCAI}, 2013.

\bibitem{XuWLZLY11}
H.~Xu, J.~Wang, Z.~Li, G.~Zeng, S.~Li, and N.~Yu.
\newblock Complementary hashing for approximate nearest neighbor search.
\newblock In {\em ICCV}, pages 1631--1638, 2011.

\bibitem{XuSYSL17}
X.~Xu, F.~Shen, Y.~Yang, H.~T. Shen, and X.~Li.
\newblock Learning discriminative binary codes for large-scale cross-modal
  retrieval.
\newblock {\em IEEE Transactions on Image Processing}, 26(5):2494--2507, May
  2017.

\bibitem{YangBZRZC14}
H.~Yang, X.~Bai, J.~Zhou, P.~Ren, Z.~Zhang, and J.~Cheng.
\newblock Adaptive object retrieval with kernel reconstructive hashing.
\newblock In {\em CVPR}, pages 1955--1962, 2014.

\bibitem{YangHZL13}
Q.~Yang, L.-K. Huang, W.-S. Zheng, and Y.~Ling.
\newblock Smart hashing update for fast response.
\newblock In {\em IJCAI}, 2013.

\bibitem{YangSSLL15}
Y.~Yang, F.~Shen, H.~T. Shen, H.~Li, and X.~Li.
\newblock Robust discrete spectral hashing for large-scale image semantic
  indexing.
\newblock {\em {IEEE} Trans. Big Data}, 1(4):162--171, 2015.

\bibitem{YuKGC14}
F.~Yu, S.~Kumar, Y.~Gong, and S.-F. Chang.
\newblock Circulant binary embedding.
\newblock In {\em ICML (2)}, pages 946--954, 2014.

\bibitem{ZhangWCL10b}
D.~Zhang, J.~Wang, D.~Cai, and J.~Lu.
\newblock Self-taught hashing for fast similarity search.
\newblock In {\em SIGIR}, pages 18--25, 2010.

\bibitem{ZhangSLHLC16}
H.~Zhang, F.~Shen, W.~Liu, X.~He, H.~Luan, and T.~Chua.
\newblock Discrete collaborative filtering.
\newblock In {\em {SIGIR}}, pages 325--334, 2016.

\bibitem{ZhangZSLC16}
H.~Zhang, N.~Zhao, X.~Shang, H.~Luan, and T.~Chua.
\newblock Discrete image hashing using large weakly annotated photo
  collections.
\newblock In {\em {AAAI}}, pages 3669--3675, 2016.

\bibitem{ZhangZTGLT13}
L.~Zhang, Y.~Zhang, J.~Tang, X.~Gu, J.~Li, and Q.~Tian.
\newblock Topology preserving hashing for similarity search.
\newblock In {\em ACM Multimedia}, pages 123--132, 2013.

\bibitem{ZhangLGZ16}
S.~Zhang, J.~Li, J.~Guo, and B.~Zhang.
\newblock Scalable discrete supervised hash learning with asymmetric matrix
  factorization.
\newblock {\em CoRR}, abs/1609.08740, 2016.

\bibitem{ZhangDW14}
T.~Zhang, C.~Du, and J.~Wang.
\newblock Composite quantization for approximate nearest neighbor search.
\newblock In {\em ICML (2)}, pages 838--846, 2014.

\bibitem{ZhangQTW15}
T.~Zhang, G.-J. Qi, J.~Tang, and J.~Wang.
\newblock Sparse composite quantization.
\newblock In {\em CVPR}, 2015.

\bibitem{ZhangW16}
T.~Zhang and J.~Wang.
\newblock Collaborative quantization for cross-modal similarity search.
\newblock In {\em {CVPR}}, pages 2036--2045, 2016.

\bibitem{ZhaoHWT15}
F.~Zhao, Y.~Huang, L.~Wang, and T.~Tan.
\newblock Deep semantic ranking based hashing for multi-label image retrieval.
\newblock In {\em CVPR}, pages 1556--1564, 2015.

\bibitem{ZhaoLM14}
K.~Zhao, H.~Lu, and J.~Mei.
\newblock Locality preserving hashing.
\newblock In {\em AAAI}, pages 2874--2881, 2014.

\bibitem{ZhenY13}
Y.~Zhen and D.-Y. Yeung.
\newblock Active hashing and its application to image and text retrieval.
\newblock {\em Data Min. Knowl. Discov.}, 26(2):255--274, 2013.

\bibitem{ZhuHCCS13}
X.~Zhu, Z.~Huang, H.~Cheng, J.~Cui, and H.~T. Shen.
\newblock Sparse hashing for fast multimedia search.
\newblock {\em ACM Trans. Inf. Syst.}, 31(2):9, 2013.

\bibitem{ZhuHSZ13}
X.~Zhu, Z.~Huang, H.~T. Shen, and X.~Zhao.
\newblock Linear cross-modal hashing for efficient multimedia search.
\newblock In {\em ACM Multimedia}, pages 143--152, 2013.

\bibitem{ZhuangLWZS11}
Y.~Zhuang, Y.~Liu, F.~Wu, Y.~Zhang, and J.~Shao.
\newblock Hypergraph spectral hashing for similarity search of social image.
\newblock In {\em ACM Multimedia}, pages 1457--1460, 2011.

\end{thebibliography}


\begin{IEEEbiography}[{\includegraphics[width=1in,height=1.25in,clip,keepaspectratio]{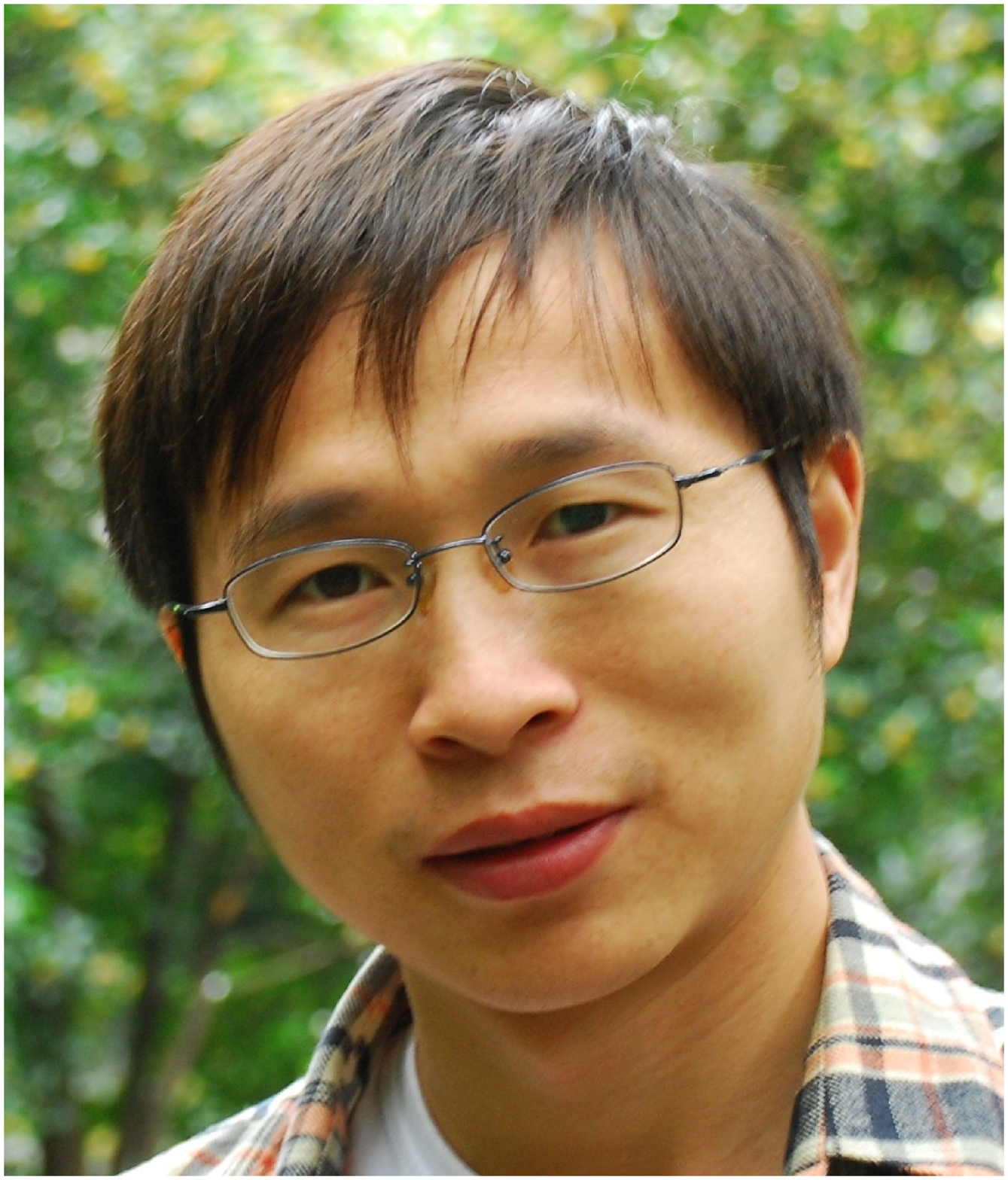}}]{Jingdong Wang}
is a Lead Researcher at the Visual Computing Group,
Microsoft Research Asia.
He received the B.Eng. and M.Eng. degrees
from the  Department of Automation, Tsinghua University,
Beijing, China, in 2001 and 2004, respectively,
and the PhD degree
from the Department of Computer Science and Engineering,
the Hong Kong University of Science and Technology,
Hong Kong, in 2007.
His areas of interest include deep learning, large-scale indexing, human understanding, and person re-identification.
He has been served/serving as an Associate Editor of IEEE TMM,
and an area chair of ICCV 2017, CVPR 2017, ECCV 2016 and ACM Multimedia 2015.
\end{IEEEbiography}
\begin{IEEEbiography}[{\includegraphics[width=1in,height=1.25in,clip,keepaspectratio]{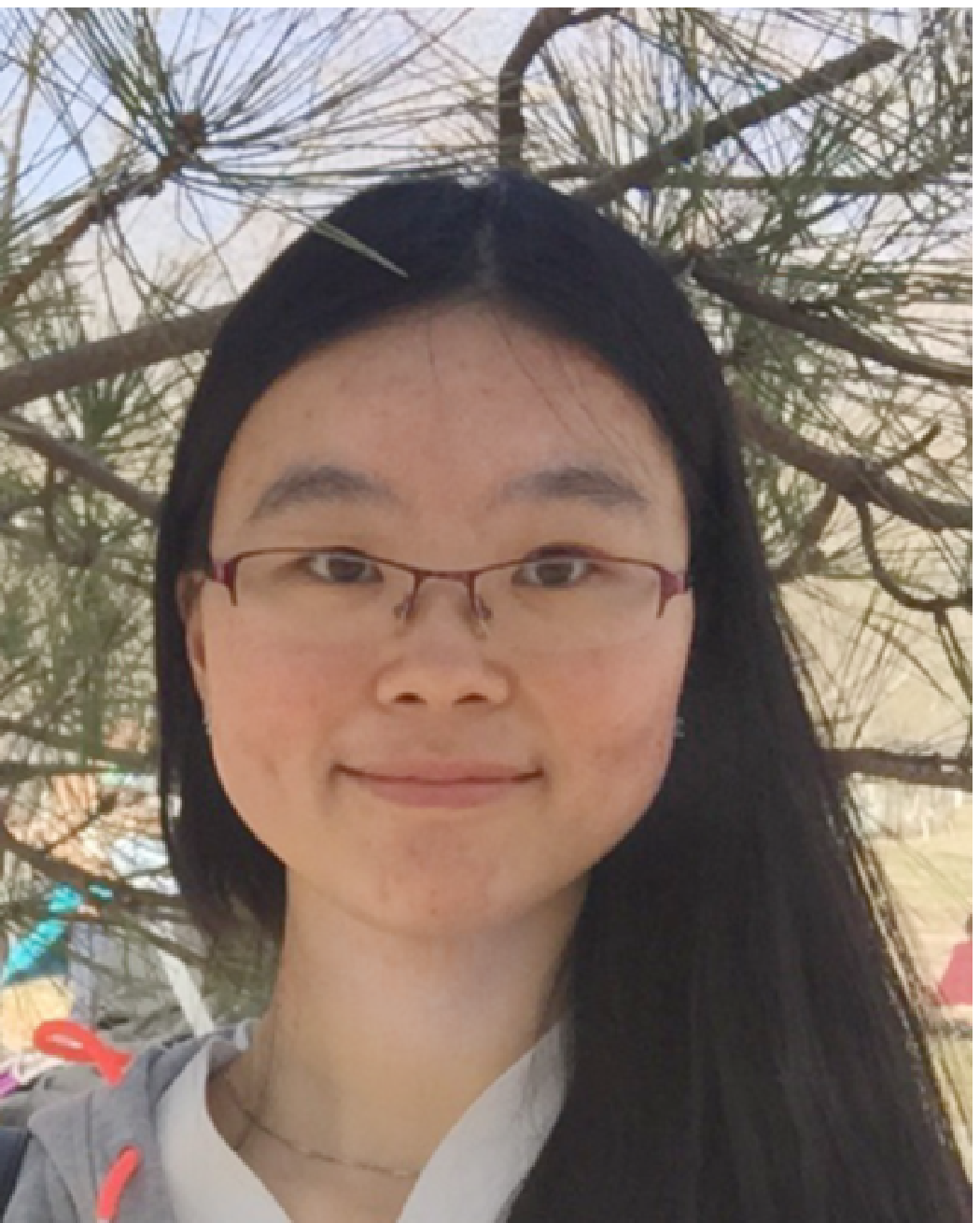}}]{Ting Zhang}
is a PhD candidate
in the department of Automation at University of Science and Technology of China.
She received the Bachelor degree in mathematical science from the school of the gifted young in 2012.
Her main research interests include machine learning, computer vision and pattern recognition.
She is currently a research intern at Microsoft Research, Beijing.
\end{IEEEbiography}
\begin{IEEEbiography}[{\includegraphics[width=1in,height=1.25in,clip,keepaspectratio]{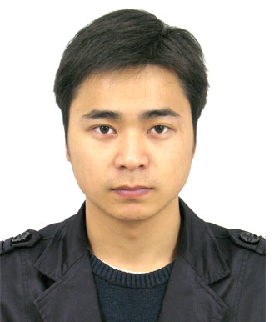}}]{Jingkuan Song}
received his PhD degree in
Information Technology from The University
of Queensland, Australia. He received his BS
degree in Software Engineering from University
of Electronic Science and Technology of China.
Currently, he a postdoctoral researcher in the
Dept. of Information Engineering and Computer
Science, University of Trento, Italy. His research
interest includes large-scale multimedia search
and machine learning.
\end{IEEEbiography}
\begin{IEEEbiography}[{\includegraphics[width=1in,height=1.25in,clip,keepaspectratio]{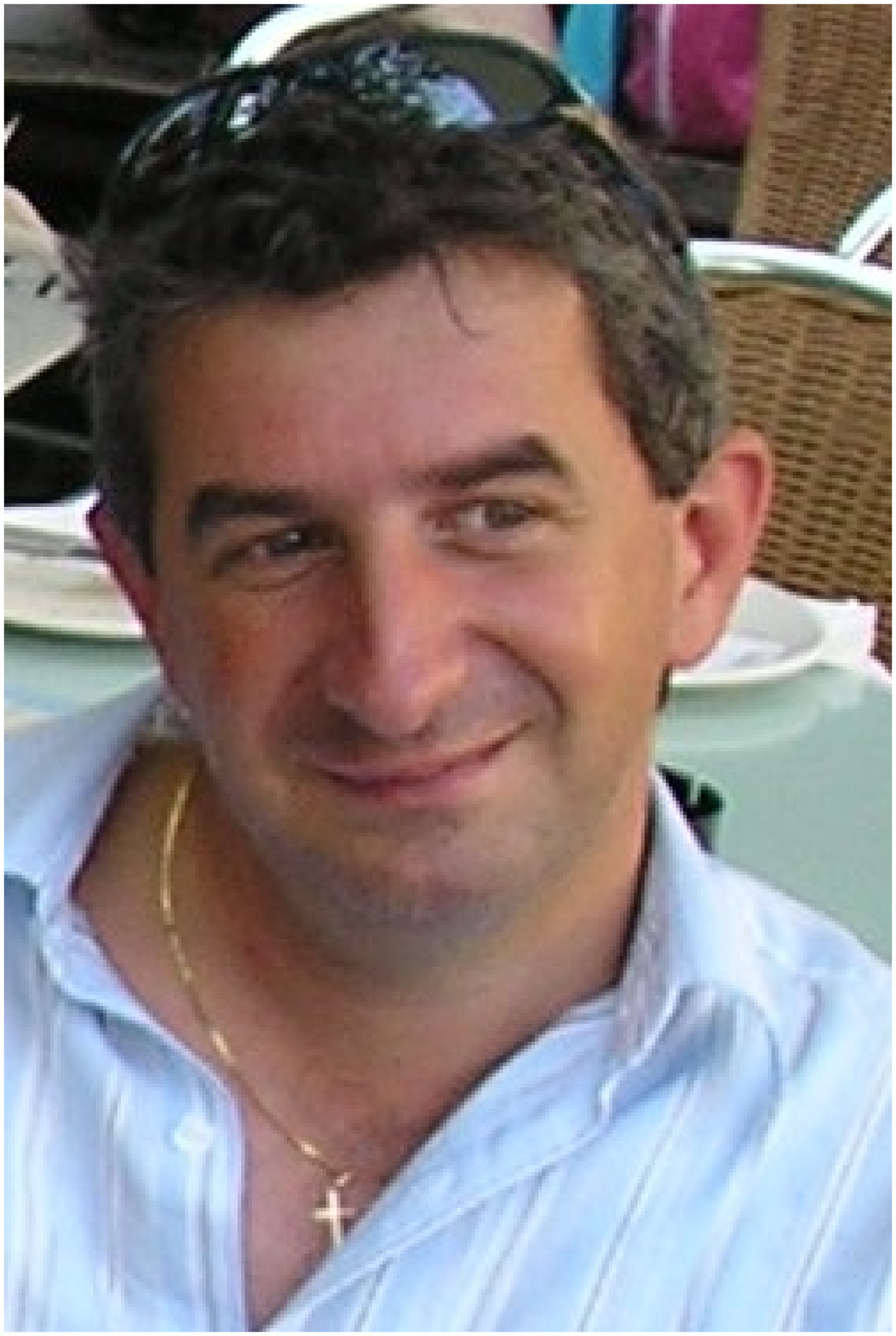}}]{Nicu Sebe}
is currently a Professor with the
University of Trento, Italy, leading the research
in the areas of multimedia information retrieval
and human behavior understanding. He was the
General Co-Chair of the IEEE FG Conference
2008 and ACM Multimedia 2013, and the Program
Chair of the International Conference on
Image and Video Retrieval in 2007 and 2010, and
ACM Multimedia 2007 and 2011.
He is/will be the
Program Chair of ECCV 2016 and ICCV 2017.
%He is a co-chair of the IEEE Computer Society Task Force on Human-centered Computing and is an associate editor Computer Vision and Image Understanding, Machine Vision and Applications, Image and Vision Computing, International Journal of Human-computer Studies and of Journal of Multimedia.
 He is a fellow of the International Association
for Pattern Recognition.
\end{IEEEbiography}
\begin{IEEEbiography}[{\includegraphics[width=1in,height=1.25in,clip,keepaspectratio]
{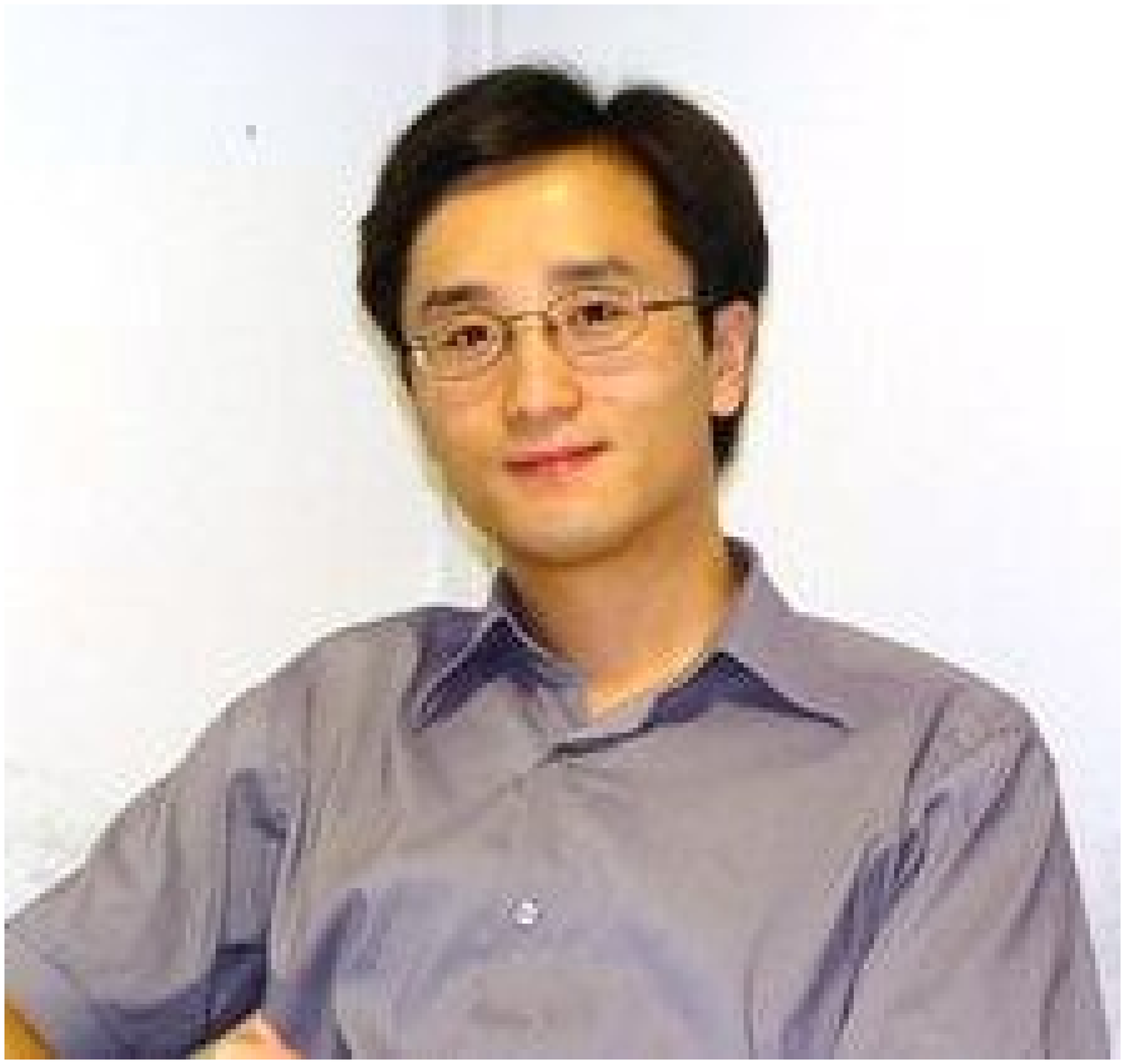}}]{Heng Tao Shen}
obtained his BSc with 1st class Honours and PhD
from Department of Computer Science, National University of Singapore
in 2000 and 2004 respectively.
He then joined University of Queensland as a Lecturer, Senior Lecturer, Reader,
and became a Professor in late 2011.
He is currently a Professor of National "Thousand Talents Plan"
and the Director of Future Media Research Center
at University of Electronic Science and Technology of China.
His research interests mainly include Multimedia Search,
Computer Vision, and Big Data Management on spatial, temporal,
multimedia and social media databases. Heng Tao has extensively published
and served on program committees
in most prestigious international publication venues of interests.
He received the Chris Wallace Award for outstanding Research Contribution
in 2010 conferred by Computing Research and Education Association, Australasia. He has served as a PC Co-Chair for ACM Multimedia 2015 and currently is an Associate Editor of IEEE Transactions on Knowledge and Data Engineering.
\end{IEEEbiography}

\end{document}